\documentclass{article}
\usepackage[x11names,svgnames,table]{xcolor}

\usepackage{PRIMEarxiv}

\usepackage[utf8]{inputenc} 
\usepackage[T1]{fontenc}    
\usepackage{hyperref}       
\usepackage{url}            
\usepackage{booktabs}       
\usepackage{amsfonts}       
\usepackage{nicefrac}       
\usepackage{listings}
\usepackage{microtype}      
\usepackage{lipsum}
\usepackage{tcolorbox}
\usepackage{fancyhdr}       
\usepackage{graphicx}       
\graphicspath{{media/}}     
\usepackage{algorithm,algpseudocode}
\usepackage{multirow}
\usepackage{makecell}
\usepackage{amsmath}
\usepackage{amssymb}
\usepackage{float}
\usepackage{color}

\title{Phi-Ground Tech Report:  Advancing Perception in GUI Grounding
}

\author{
  Miaosen Zhang \ \ 
  Ziqiang Xu \ \
  Jialiang Zhu \ \
    Qi Dai \ \ 
    Kai Qiu \ \
    Yifan Yang \\
    \textbf{Chong Luo \ \ 
    Tianyi Chen \ \ 
    Justin Wagle \ \
    Tim Franklin \ \
    Baining Guo} 
    \vspace{0.3cm}\\
    {\textbf{Microsoft}} \\
}

\begin{document}
\maketitle

\vspace{-20px}
\begin{abstract}
With the development of multimodal reasoning models, Computer Use Agents (CUAs), akin to Jarvis from \textit{"Iron Man"}, are becoming a reality. 
GUI grounding is a core component for CUAs to execute actual actions, similar to mechanical control in robotics, and it directly leads to the success or failure of the system.
It determines actions such as clicking and typing, as well as related parameters like the coordinates for clicks.
Current end-to-end grounding models still achieve less than 65\% accuracy on challenging benchmarks like ScreenSpot-pro and UI-Vision, indicating they are far from being ready for deployment.
In this work, we conduct an empirical study on the training of grounding models, examining details from data collection to model training. Ultimately, we developed the \textbf{Phi-Ground} model family, which achieves state-of-the-art performance across all five grounding benchmarks for models under $10B$ parameters in agent settings. In the end-to-end model setting, our model still achieves SOTA results with scores of \textit{\textbf{43.2}} on ScreenSpot-pro and \textit{\textbf{27.2}} on UI-Vision. We believe that the various details discussed in this paper, along with our successes and failures, not only clarify the construction of grounding models but also benefit other perception tasks.
Project homepage: 
\href{https://zhangmiaosen2000.github.io/Phi-Ground/}{https://zhangmiaosen2000.github.io/Phi-Ground/}

\end{abstract}

\vspace{-10px}

\begin{figure}[H]
    \centering
    \includegraphics[width=\linewidth, trim=0 30 0 50, clip]{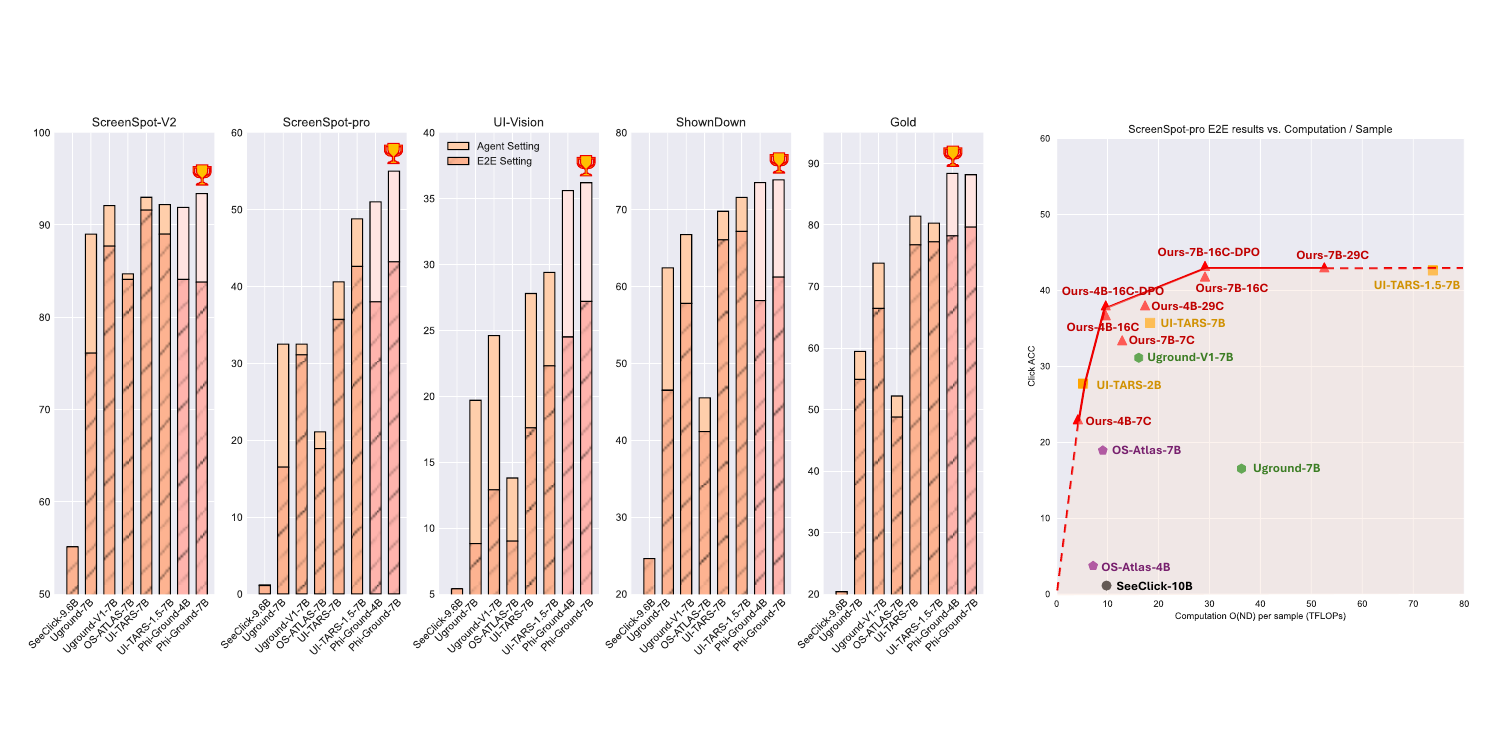}
    \vspace{-20px}
    \caption{
    \textbf{Left}: The comparison chart of our grounding model results across five GUI grounding benchmarks. Our model, trained specifically for the agent setting, achieved SOTA results on all benchmarks under this focus. Even in the general end-to-end model setting, our model attained SOTA results on three of the benchmarks. \textbf{Right}: The relationship between model performance and computational cost on ScreenSpot-pro demonstrates that our model supports the Pareto frontier, indicating its efficiency. Most GUI research traditionally considers only the parameter count $N$ for comparison, but our experiments highlight that computational cost during testing, such as the number of image tokens, also significantly impacts performance. The X-axis in the right figure represents $ND$, where $D$ is the number of image tokens. Training and inference latency are more linearly correlated with $ND$ than with $N$. A graph using latency as the X-axis closely resembles the right figure, but latency is often influenced by hardware and acceleration libraries such as vllm, so we did not use latency as X-axis.
}
    \label{fig:abs}
\end{figure}

\begin{figure}[htb]
    \centering
    \includegraphics[width=\linewidth, trim=0 0 0 20, clip]{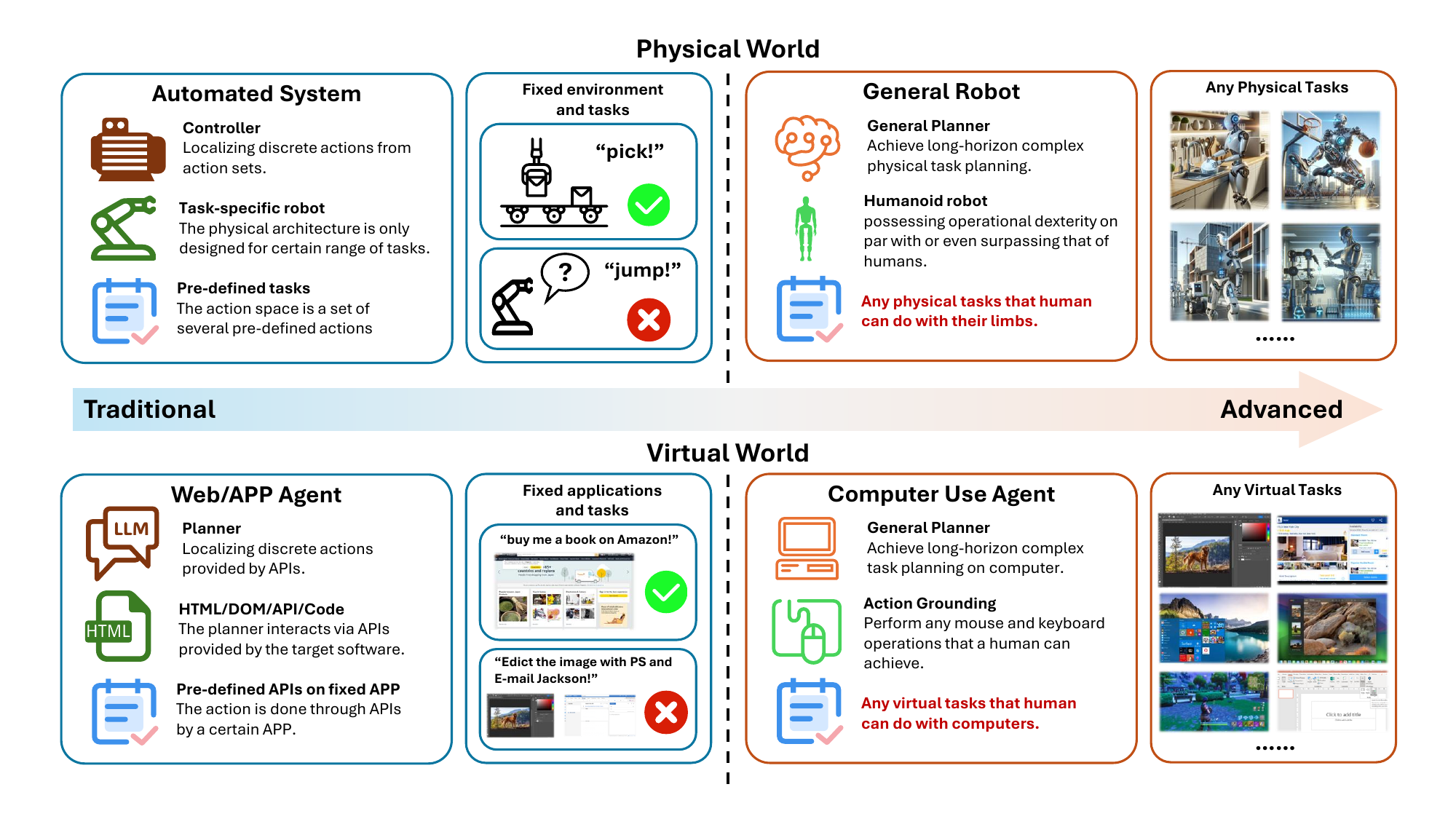}
    \vspace{-20px}
    \caption{Agent evolution across physical and virtual worlds. Traditional systems rely on fixed controllers and pre-defined workflows to execute domain-specific tasks, either in physical environments (e.g., task-specific robots) or virtual environments (e.g., API-based Web/APP agents). In the modern era, intelligent automation has emerged. In the physical world, general-purpose robots perform versatile limb-based operations. In the virtual world, Computer Use Agents (CUAs) achieve human-level behaviors through general purpose planner, GUI grounding, enabling them to complete any virtual task achievable via mouse and keyboard interactions.}
    \label{fig:intro-motivation}
\end{figure}

\section{Introduction}

Large model based autonomous agents \cite{agent1, agent3}, by leveraging their robust reasoning capabilities and interactions with real-world environments \cite{agent2}, enable humans to tackle more complex tasks and significantly enhance productivity. Given that a substantial portion of modern human work is conducted via computers, Computer Use Agents (CUAs) hold immense potential and commercial value \cite{cua1, cua3}. While CUAs streamline virtual work, robots \cite{robot1, robot2} simplify physical tasks, as shown in Figure \ref{fig:intro-motivation}, and together, they are poised to drive a revolution in productivity. With the development of reasoning models like O3 \cite{o3o4, claude4, deepseekr1, qwen3}, we are beginning to create prototypes of CUAs, such as OpenAI Operator \cite{operator} and Claude Computer Use \cite{claudeCUA}. However, there remains a considerable gap before CUAs can be fully commercialized. This is due to the irreversible effects of many computer operations \cite{cua2}, such as closing unsaved files, as well as significant concerns regarding user privacy protection \cite{privacy1, privacy2, privacy3}.

Specifically, CUAs can execute actions through the following methods: APIs (e.g., HTML/DOM, Office scripts), scripting tools (such as AutoHotkey, Power Automate, CLI), and simulated input device interactions (such as mouse clicks and keyboard inputs). Traditional approaches based on APIs \cite{html1, html2} and scripts are often constrained by the platform in use, environmental dependencies, and the specific APIs provided by applications \cite{aguvis}. In contrast, solutions based on GUI and simulated input device interactions \cite{guiact} are aligned with human operations. This alignment not only facilitates user supervision, enhancing privacy and security, but also theoretically allows for any interaction that a human can perform, without being limited by the application. As a result, GUI-based CUAs are becoming a focal point of research in this field, as illustrated in Figure \ref{fig:intro-motivation}.

When accomplishing a task, GUI-based CUA can be divided into two steps: temporal planning and grounding \cite{pg, tpsp}. 
Planning involves analyzing the task description and the current state to determine the actions that should be taken in the future, while grounding refers to the above mentioned simulation of input devices. Among all interactions, mouse click operations are the most important and common actions for GUI grounding. Since keyboard commands, such as pressing the "A" key, are discrete, MLLMs can effectively handle this type of grounding. Hence, our focus is primarily on mouse commands, where the main challenge lies in the fact that mouse command parameters are screen coordinates, and most MLLMs struggle to accurately identify these coordinates \cite{uivision, sspro, seeclick}. Therefore, specialized training is required for determining the precise click coordinates.

\begin{figure}[tb]
    \centering
    \includegraphics[width=\linewidth, trim=0 30 0 0, clip]{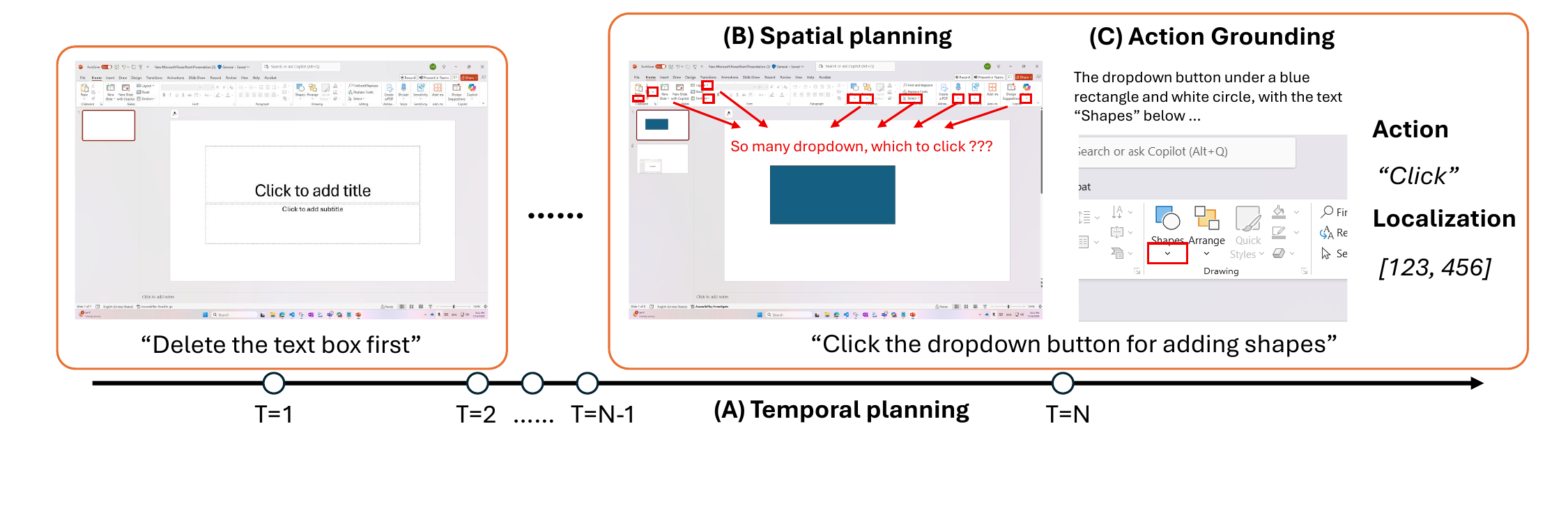}
    \vspace{-30px}
    \caption{Three levels of task of CUAs. Each coral block represent an action step. We focus on the click action, as it is the most important and common operation. Others such as keyboard input can be effectively handled by MLLM.}
    \label{fig:planning-grounding}
\end{figure}

Motivated by the above considerations, this paper conducts a detailed empirical study on the training of GUI grounding models. We divide GUI grounding into two components: the first component is spatial planning \cite{sp, tpsp}, which involves identifying which specific element in the image needs to be manipulated to execute a given instruction. The second component is localization, where, after determining the target location, the model needs to output the correct coordinates, as illustrated in Figure \ref{fig:planning-grounding}. 
Many existing CUA models \cite{uitars, uitars15, Uground} attempt to accomplish the grounding task in an end-to-end manner. 
However, due to the spatial planning component, which also demands a strong level of expertise, common sense, and spatial reasoning abilities, there is a need for larger or even deep reasoning models. To achieve better results at this stage, we have adopted a two-phase approach \cite{osatlas}. Initially, an advanced MLLM provides a detailed description of the location, followed by our trained grounding model outputting the specific coordinates.
In this technical report, we delve into numerous often-overlooked details spanning data, algorithms, and training methodologies. Counterintuitively, we found that many techniques that appear sound and frequently appear in previous work, such as tokenized coordinates \cite{tokenized1, tokenized2, tokenized3}
, coordinate label smoothing \cite{smoothing1, smoothing2}, and loss reweighting, become trivial when dealing with large-scale training. We have retained only the following techniques that remain meaningful under extensive training in the main text.

Firstly, at the model input level, we discovered that the order of modality inputs can significantly impact the underlying mechanism of feature modeling, leading to notable differences in results. Secondly, data augmentation \cite{dataaug} is very common in traditional object detection training, yet is rarely mentioned in the era of large models. Our re-experiments indicate that certain data augmentations can greatly enhance results in high-resolution scenarios (such as ScreenSpot-pro).
Thirdly, the academic community predominantly focuses on the out-of-distribution generalization ability of models, with insufficient research on how models can appropriately perform in-domain continual learning in specific scenarios (such as considering only Photoshop software). This area, however, holds significant practical value. We discuss data strategies and training algorithms for in-domain post-training. Lastly, existing GUI grounding work typically considers only the size of the parameters when making comparisons. We incorporate the computational load during model inference (primarily influenced by the number of image tokens) into the study of scaling laws\cite{sl1, sl2, sl3} and evaluations.

Ultimately, we devised an efficient and rational training recipe, collecting over 40M data samples from multiple sources \cite{guiact, osatlas, e2isynth}. By scaling up the training volume, we successfully developed the Phi-Ground model family.

In our evaluation process, to avoid the limitations of relying on a single benchmark and to prevent systemic overfitting \cite{systemoverfit1, systemoverfit2} due to optimization for a specific benchmark, we consider multiple benchmarks in both ablation studies and final evaluations. We conducted a survey and collected four publicly available test datasets \cite{sspro, uivision, showdown, seeclick}, complemented by our internally constructed dataset focusing on commonly used software on Windows, which we refer to as the Gold dataset. This brings our total to five evaluation datasets, whereas most CUA research works \cite{Uground, osatlas, uitars, uitars15} on grounding typically utilize only one or two of these. The evaluation results demonstrate that, within the Agent setting we are focused on, our Phi-Ground model achieves state-of-the-art results across all benchmarks, with particularly high scores of 55.0 and 36.2 on ScreenSpot-pro \cite{sspro} and UI-Vision \cite{uivision}, respectively. Furthermore, in the end-to-end model setting, we also achieved the best results in three of the benchmarks, scoring 43.2 and 27.2 on ScreenSpot-pro and UI-Vision, respectively. These findings indicate that our model possesses strong generalization capabilities.

We also present a detailed erroneous case study in Section \ref{sec:error-analyze} and the appendix. We consider GUI grounding to be a classic scenario for multimodal model perception \cite{lmmpercept1, lmmpercept2, lmmpercept3}, and believe that our experiences can be effectively generalized to other fields involving multimodal perception. We hope our research will benefit related domains.

\section{Methodology}

\begin{figure}
    \centering
    \includegraphics[width=\linewidth, trim=120 210 100 50, clip]{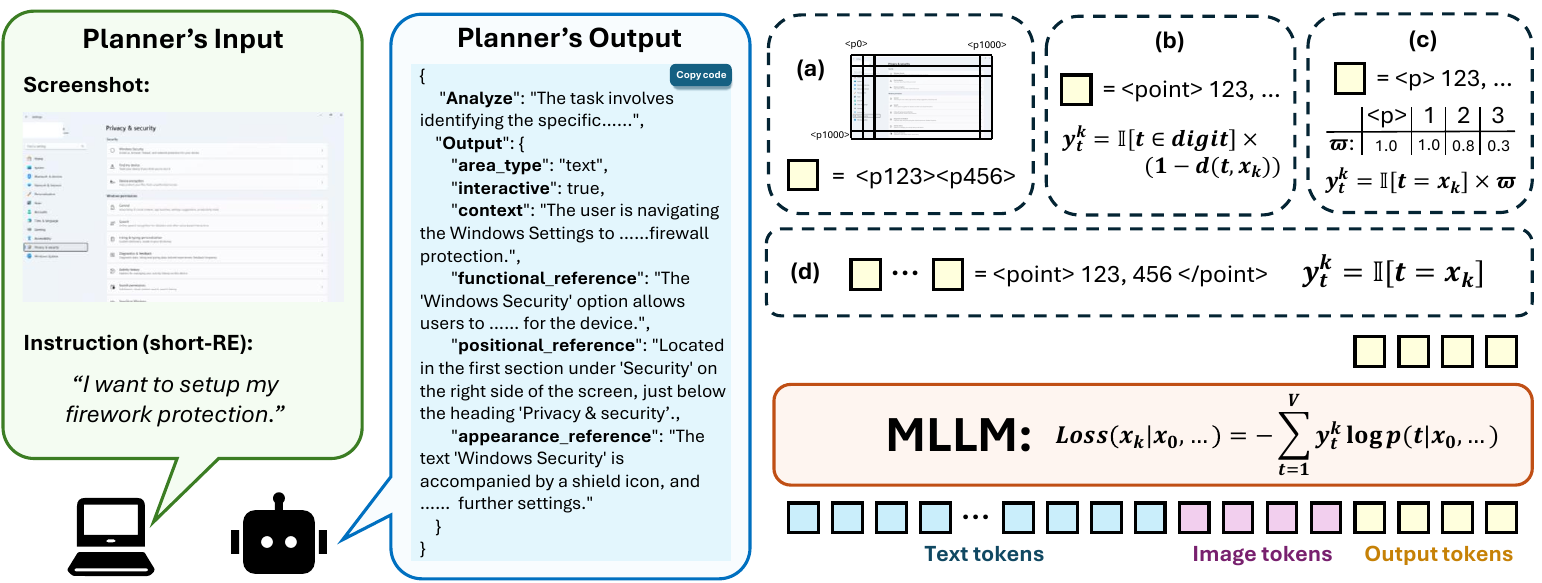}
    \caption{In terms of coordinate representation and loss strategies, we experimented with the following approaches: (a) tokenized coordinate representation, (b) label smoothing for coordinates, (c) loss reweighting, and (d) direct textual representation of coordinates.}
    \label{fig:methodology}
\end{figure}

At the agent level, we adopt a two-stage implementation approach: a large multimodal model (such as GPT-4O) is utilized to generate detailed and specific Reference Expressions (RE), while a smaller multimodal model that we have trained is responsible for generating coordinates based on the RE. At the model level, we fine-tune a MLLM to directly output coordinates, which are generated in text form. The coordinates are represented as relative values scaled by 1000 and then rounded (e.g., a value of 500 corresponds to a coordinate value of 0.5). We present the following discussion and experiments on the aforementioned implementation.

\paragraph{Coordinates Representations and Loss} During the development, we explored several potential implementation approaches as follows:

\begin{itemize}
    \item \textbf{Tokenized coordinates. } Some existing works represent regions as tokens \cite{tokenized1, tokenized2, tokenized3}, using learnable special tokens to denote a specific position. However, in GUI grounding, the large image resolution makes using 2D regions as tokens impractical. We attempted to divide both the horizontal and vertical coordinates into 1000 intervals (which is necessary for large screens like 4K displays), with each interval represented by a special token, similar to \cite{tokenizetext1, tokenizetext2}. Despite experimenting with various initialization and training strategies, our results indicated that introducing a large number of unlearned tokens during finetuning pretrained MLLM can lead to model collapse and poor performance. For further details, see the Appendix \ref{sec:special-token}.
    \item \textbf{Label smoothing. } To approximate the classic regression loss, we applied label smoothing to the loss of digit tokens, making the magnitude of the loss proportional to the distance between the prediction and the target. For instance, if the target digit is "5", we assign smaller labels to "4" and "6" to indicate their proximity to the target. The detailed implementation can be found in the Appendix \ref{sec:label-smoothing}. However, the experimental results indicate that this technique may offer improvements only when the training dataset is small and the batch size is minimal. When we eventually increased the batch size to 2048 and the total training samples reached million level, the application of this technique showed no significant impact or improvement.
    \item \textbf{Loss re-weighting. } Following a similar idea as Label smoothing, we assigned different loss weights to the digits representing the hundreds, tens, and units places of the coordinates. The results were consistent with the previous findings: not only were the corresponding parameters extremely sensitive, but the technique also lost its advantages as the batch size and training volume increased. See Appendix. \ref{sec:loss-reweright}.
\end{itemize}

From the above experiments, we found that the most straightforward text output combined with GPT loss (NTP) can effectively facilitate training. We chose to use relative coordinates for expressing positions because their value range is fixed and relatively dense (as the resolution width and height of GUIs are generally greater than 1000 pixels), which facilitates better model learning. Additionally, since relative coordinates are floating-point numbers that are $\leq 1$, the leading "0." is insignificant. Therefore, following \cite{osatlas}, we scaled these values by multiplying them by 1000.

\section{Data Preparation}

\subsection{Processing Open Source Data}

We utilized a portion of the open-source data from OS-Atlas \cite{osatlas}, which includes several subsets such as Windows, macOS, Linux, and Fineweb. Additionally, we incorporated data from SeeClick \cite{seeclick}, E2ISynth \cite{e2isynth} and GUIAct \cite{guiact}. The total data volume amounts to approximately 10M entries. By employing the prompt outlined in Appendix \ref{sec:prompt-long-gold}, we re-annotated all the data using GPT-4O to generate ``Long-gold'' type reference expressions (as described in Sec \ref{sec:evaluation-setting}). Furthermore, we constructed training data by randomly combining three types of reference expressions.

\subsection{CommonCrawl Data}
\label{sec:ccdata}
\begin{figure}
    \centering
    \includegraphics[width=\linewidth, trim=40 90 30 10, clip]{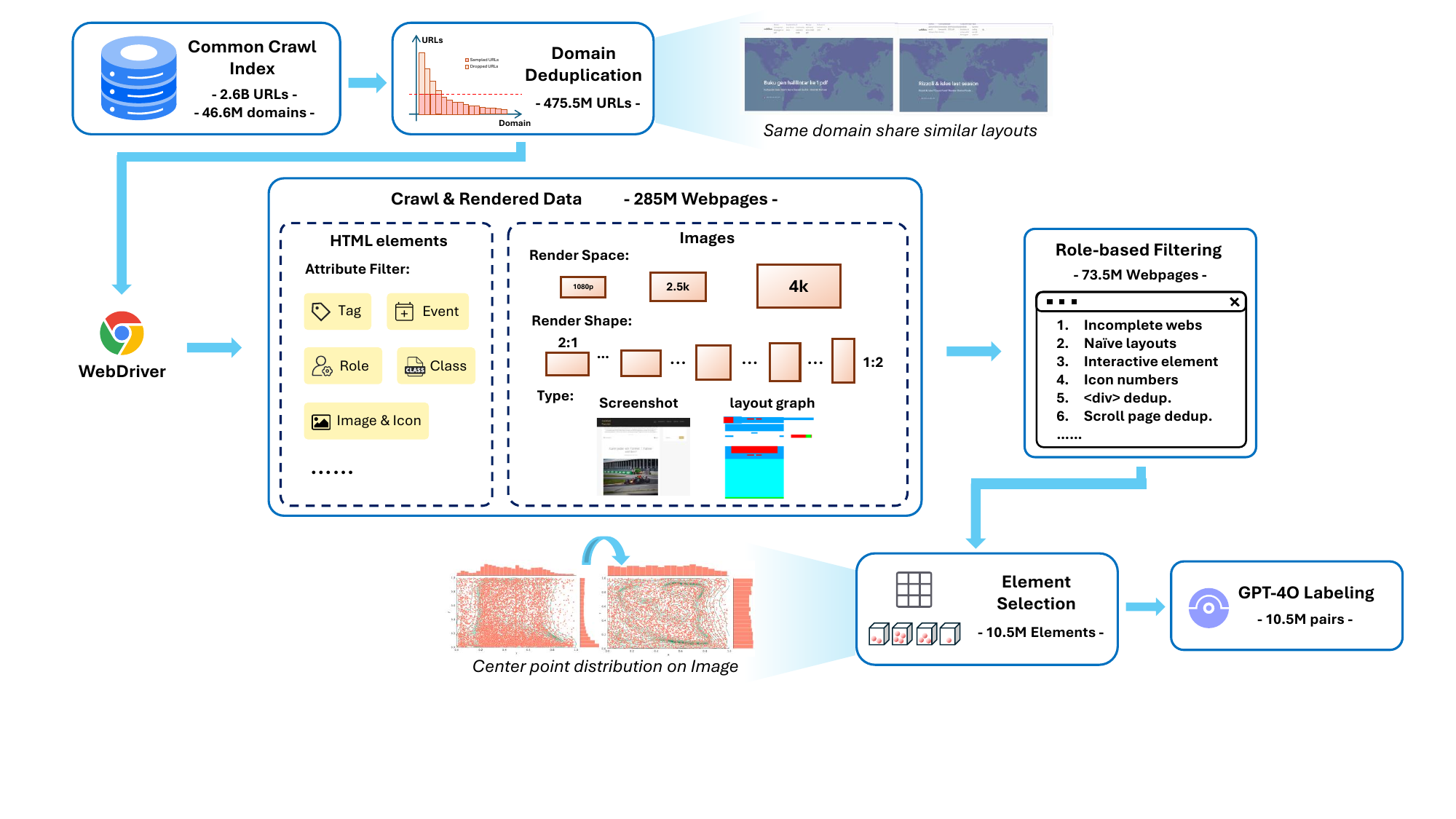}
    \caption{CommonCrawl data processing pipeline.}
    \label{fig:CC-cleaning}
\end{figure}

To acquire larger-scale data for better scaling up of training, we also obtained web pages from CommonCrawl \cite{commoncrawl} and rendered screenshots to generate training data. However, the web data contained a significant amount of noisy data that caused training failures. To address this, we constructed a highly specific data cleaning pipeline, as illustrated in Figure \ref{fig:CC-cleaning}. Below are the detailed steps of each stage:

\paragraph{Index and domain deduplication} We utilized the \textit{CC-MAIN-2024-46} crawl from CommonCrawl. After a basic deduplication of URLs (exact match), filtering by language (retaining only English), and webpage status (retaining only 2xx, 301, and 302), we were left with 2.6 billion URLs. These 2.6B URLs originate from 45.6 million unique domains, with the number of pages from the same domain displaying a long-tail distribution. For instance, the largest domain contains 204K different pages. We observed that pages from the same domain exhibit strong consistency in layout. Therefore, to ensure the generalizability of our model, we performed random sampling so that no more than 50 pages were selected from each domain. After this round of sampling, we were left with 475.45M URLs.

\paragraph{Rendering} We utilized the Selenium library and Google Chrome Driver to render webpage screenshots. During the rendering process, we randomly selected from three different pixel areas corresponding to 1080p, 2K, and 4K screen resolutions. The aspect ratio of the images was randomly chosen between 2:1 and 1:2. For the elements within the webpage HTML, we designed several rules for filtering and retaining them, as detailed in Appendix \ref{sec:CC-render-rule}. This process allowed us to preserve elements that are likely to be interactive components. At this stage, we save webpage screenshots, element information, and layout graphs (with different types such as interactive text buttons, interactive icon buttons, and images corresponding to specific colors). After this stage, there retained 285M webpages.

\paragraph{Rule-based filtering} Subsequently, we designed more fine-grained filters and deduplication techniques at the webpage and element levels based on the preserved webpages. The specific details are provided in Appendix \ref{sec:CC-filter-rule}. These filters eliminated many erroneous and overly simplistic webpages. After this phase, 73.5M webpages remained.

\paragraph{Element selection and labeling} Finally, when selecting elements, we consider the distribution of element centroids and their types, as we found it necessary to do so in Sec. \ref{sec:data-dist}. Specifically, we discretize and uniformly sample across various regions of the canvas (see Appendix \ref{sec:re-sample-alg}). During sampling in a discrete area, we prioritize sampling icon elements, as they are less frequent. We sample only one element per webpage screenshot. Consequently, after this stage, 10.5M elements and screenshots remain. Finally, we use GPT-4O to annotate all the data with prompt in Appendix \ref{sec:prompt-long-gold}.

\begin{figure}[h!]
  \begin{minipage}[t]{0.4\textwidth}
\begin{figure}[H]
    \centering
    \includegraphics[width=0.8\linewidth, trim=0 0 0 0, clip]{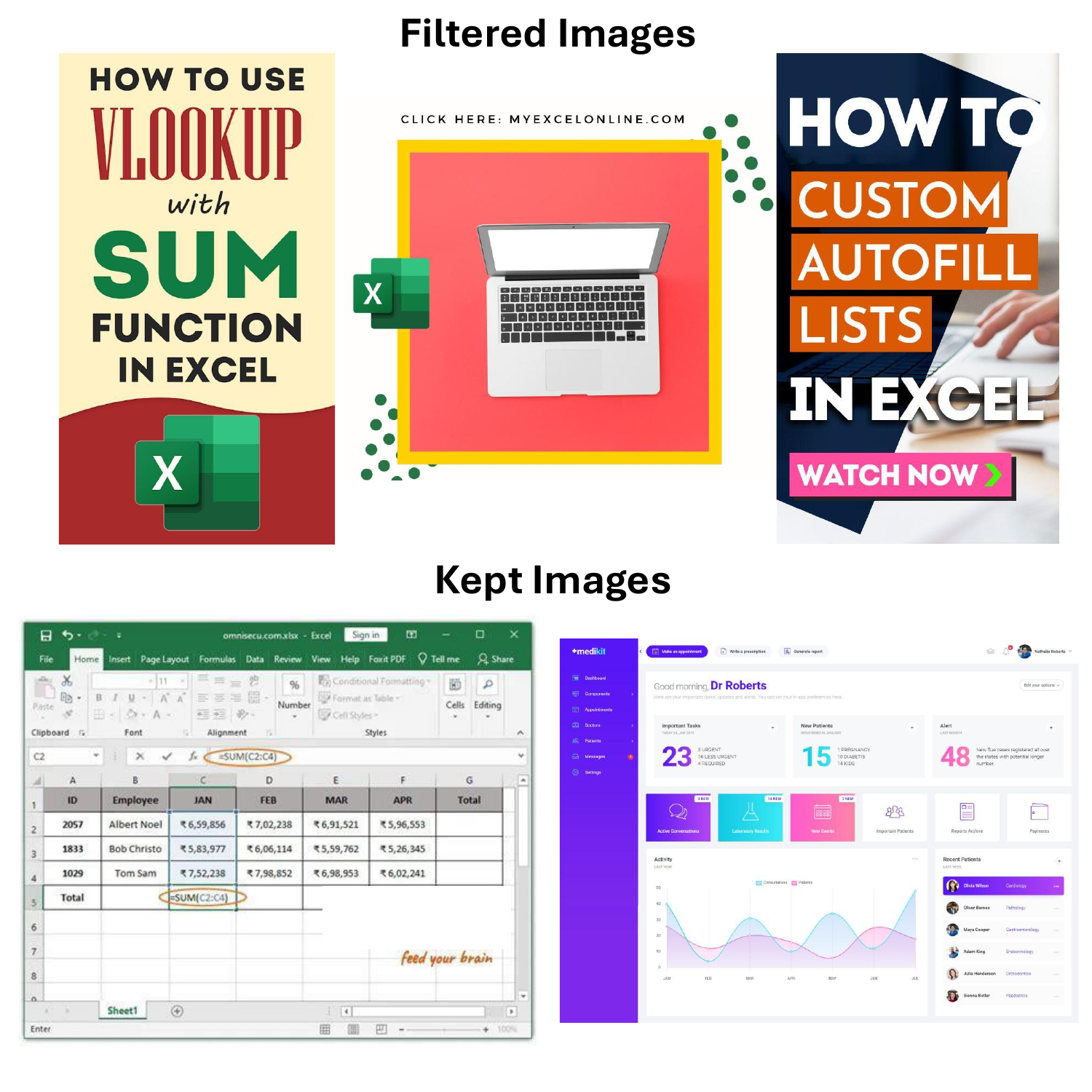}
    \caption{Filtered and kept image of the Web Search Data with the classifier.}
    \label{fig:dwebsearch-image}
\end{figure}

  \end{minipage}
  \hfill
  \begin{minipage}[t]{0.6\textwidth}
  \vspace{0.5cm}
    \begin{table}[H]
  \centering
  \caption{The applications distribution of the search queries. These applications are derived from an internal statistical list of the most frequently used applications on Windows.}
   \resizebox{\textwidth}{!}{
    \begin{tabular}{lcrc}
    \toprule
    \textbf{Domain} & \textbf{\# Apps} & \multicolumn{1}{l}{\textbf{Example apps}} & \textbf{\# Queries} \\
    \midrule
    Office \& Productivity & 10    & \multicolumn{1}{l}{Word, Excel} & 50 \\
    Media Playback \& Creation  & 11    & \multicolumn{1}{l}{VLC, Photoshop, Clipchamp } & 55 \\
    Gaming Platforms & 10    & \multicolumn{1}{l}{Steam, League of Legends} & 50 \\
    Social \& Communication & 10    & \multicolumn{1}{l}{Discord, Zoom } & 50 \\
    Security \& System Tools & 10    & \multicolumn{1}{l}{CCleaner, Malwarebytes} & 50 \\
    Core Windows UI  & 15    & \multicolumn{1}{l}{Settings, File Explorer} & 75 \\
    \midrule
    \textbf{Total} & \textbf{66} &  \multicolumn{1}{l}{\textbf{-}}     & \textbf{330} \\
    \bottomrule
    \end{tabular}}
  \label{tab:websearch-dist}%
\end{table}
  \end{minipage}
\end{figure}


\subsection{Web Search Data}
We construct a complementary, high-resolution screenshot corpus with the Bing Image Search v7 API. We first generate queries from 6 UI domains which covering a total of 66 desktop applications, as shown in Table \ref{tab:websearch-dist}.  For each application we manually construct file search phrases such as \textit{``Word ribbon interface''} or \textit{``screen shot of VLC playback controls''}, producing 330 queries altogether.  Every query retrieves 2048 candidate images and enforces a minimum resolution of $200 \times 200$ px. \textbf{The request also specifies Bing’s \textit{license} flag to ensure that all the images we use comply with copyright requirements}. Each downloaded image is then scored by a CLIP-based classifier to remove non-screenshot images from the search result. For the final filtered screenshots, we employed OmniParserV2 \cite{omniparser} to annotate all bounding boxes. We then used the same method described in \ref{sec:ccdata} to label reference expressions and perform rule-based filtering at the bounding box level. This process resulted in approximately 158K data samples.

\subsection{Human Labeled Data}
\label{sec:human-label-data}
To address our focus on specific scenarios (Windows and common applications) and to explore in-domain training techniques, we have developed a pipeline for constructing human-labeled data. The data construction process involves three steps:
\begin{itemize}
    \item In the first step, labelers use custom screen recording software to interact with the target scenarios, which include specific software or Windows system settings pages. They are required to navigate and operate various pages and functions of the software. The screen recording software automatically captures different pages and retrieves the UIATree of the page. If the software does not implement UIA, we utilize OmniParser-V2 \cite{omniparser} to obtain the bounding boxes of elements.
    \item In the second step, recognizing that the acquired bounding boxes contain numerous errors, redundancies, and non-interactive content (such as text portions in Word documents), we developed another software tool to enable labelers to remove erroneous bounding boxes.
    \item Finally, we employ the Long-gold method, as described in Sec. \ref{sec:eval-proto}, to have GPT-4O annotate references for all bounding boxes.
\end{itemize}
Ultimately, we generated 80K training samples derived from a comprehensive suite including Microsoft Office, Windows settings, and over a dozen common software applications such as 7zip, PhotoShop, ClipChamp and audio tools.

\section{Evaluation Settings}
\label{sec:evaluation-setting}

\subsection{Benchmarks} 

To ensure the model's generalization capability and to avoid systematic overfitting to well-known benchmarks such as ScreenSpot, we have gathered several recent open-source and internally developed evaluation datasets. This approach aims to ensure the comprehensiveness of our testing.

\begin{figure}
    \centering
    \includegraphics[width=\linewidth, trim=0 0 0 0, clip]{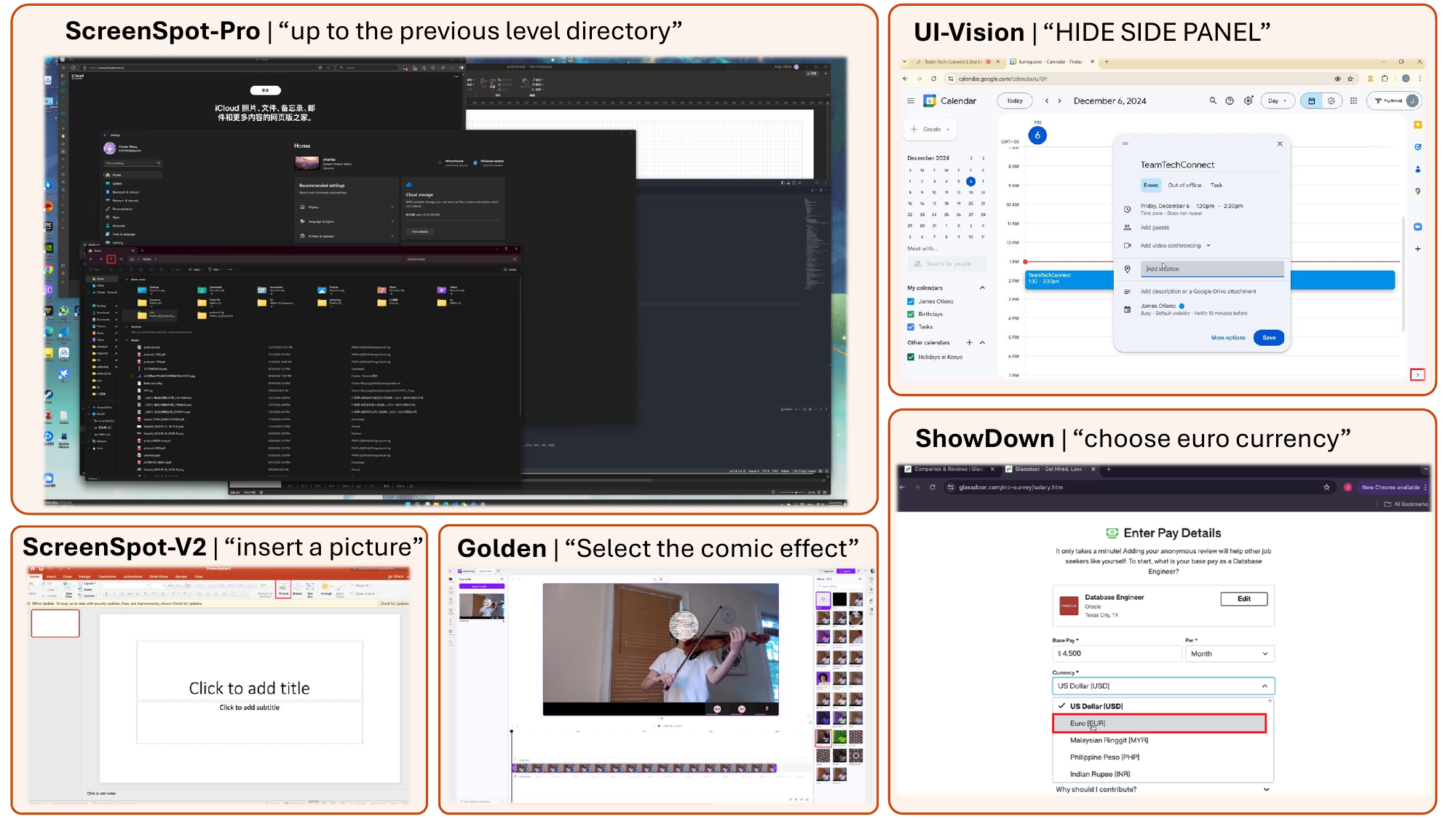}
    \caption{Examples of benchmarks used in evaluation.}
    \label{fig:benchmark-samples}
\end{figure}

\paragraph{ScreenSpot V1 \& V2}
ScreenSpot \cite{seeclick} is the first realistic GUI grounding benchmark that includes over 600 interface screenshots from mobile, desktop, and web environments. It encompasses two different types of elements, namely, texts and icons. 
However, due to several incorrect or ambiguous annotations in the original benchmark, OS-Atlas \cite{osatlas} introduces an updated version called ScreenSpot-V2, which corrects various mislabeled instances in the original ScreenSpot dataset.

\paragraph{ScreenSpot Pro}
ScreenSpot-Pro \cite{sspro} focuses on high-resolution GUI tasks in professional applications. It contains 1,581 samples encompassing 23 applications across 3 operating systems, and the micro-elements only cover average 0.07\% of the screen, which leads to a highly challenging assessment.

\paragraph{UI-Vision}
UI-Vision \cite{uivision} provides three fine-to-coarse grained tasks—Element Grounding, Layout Grounding, and Action Prediction. Similar to ScreenSpot-pro, this benchmark encompasses another set of practical applications such as education and entertainment. It also categorizes three different types of references for testing purposes. Due to the selection of smaller buttons or those requiring more expert knowledge, this benchmark is equally challenging as ScreenSpot-pro. Moreover, it features a larger sample size, enhancing the stability and reliability of the evaluations.


\paragraph{Showdown-click-dev}
Showdown \cite{showdown} is a collection of 5,679 left clicks of humans performing various tasks in a macOS desktop environment. A subset containing 557 clicks was released on GitHub and huggingface, namely Showdown-click-dev. 
It reported grounding results and latency of several advanced CUAs like OpenAI Operator \cite{operator} and Claude computer use \cite{claudeCUA}.

\paragraph{Gold Dataset (proprietary)}
We have also internally constructed an evaluation dataset tailored for application scenarios within Windows. This dataset includes six scenarios: Photoshop (213 samples), ClipChamp (179 samples), PowerPoint (82 samples), Excel (107 samples), Word (87 samples), and Windows settings (1266 samples). The bounding boxes and instructions (Short RE) for each scenario were annotated by professional engineers familiar with Windows. While these share the same domain as the human-labeled data in Sec \ref{sec:human-label-data}, the labelers come from different teams and use different computer settings, such as theme colors and screen sizes. In contrast, the selection of elements and the generation of REs in our training set's human-labeled data are automated or AI-generated, lacking expert knowledge. We employ this Gold dataset to evaluate in-domain training research. Additionally, it serves as an effective validation of practicality within Windows scenarios. In subsequent experiments, to facilitate faster testing, we manually extracted a smaller test set of 211 samples from this dataset, referred to as \textbf{Gold-S}.

We present examples of these benchmarks in Figure \ref{fig:benchmark-samples}. In addition, there are benchmarks like WinClick \cite{winclick} that are still in the process of being open-sourced. As a result, we are unable to present their results in this paper. However, we plan to showcase these results on the project homepage once they are released.

\subsection{Evaluation Protocols and metrics}
\label{sec:eval-proto}
\paragraph{Reference expression type} A Reference Expression (RE) refers to a segment of referring text provided to a grounding model or GUI agent model concerning a specific interactive area, such as a button. This expression can be an indirect instruction like "close the webpage", or a direct and specific description such as "the blue settings icon in the upper left corner". The use of different REs during training and testing can significantly impact the model's performance. Generally, we desire the model to generalize well across various REs. However, when using indirect REs, the model also needs some planning capability. To decouple model capabilities of planning and perception for research and enable efficient application in smaller models, previous works \cite{osatlas} introduced a simplified agent setting: employing more powerful MLLMs like GPT-4O \cite{gpt4o} and O4-mini \cite{o3o4} for planning and generating more detailed REs, while a smaller grounding model is used to produce coordinates. In this paper, to minimize misunderstanding, we define the following three types of REs.
\begin{itemize}
    \item \textbf{Short / instruction.} "Short RE" refers to a more concise form, potentially containing some instructions that require planning. During testing, we use "Short RE" to denote the reference provided by the benchmarks. This type of RE will be represented as the \textit{\textbf{"End-to-end model setting"}} in the subsequent results tables.
    \item \textbf{Long / agent.} Following \cite{Uground, osatlas}, for short RE, we can utilize advanced MLLM to expand it into three more explicit and detailed types of RE: functional, positional, and appearance. We allow the large model to generate these three types of references and concatenate the references together, forming what we call a Long RE. During testing, the input to the MLLM consists of only the screenshot and the short RE, and the output is the Long RE, which is then passed to a smaller model to generate coordinates. In subsequent tables, this approach is referred to as the \textit{\textbf{"Agent setting"}}. The specific system prompt for generating Long RE can be found in Appendix \ref{sec:prompt-long}.
    \item \textbf{Long-gold.} In the process of constructing the training dataset, on one hand, there is no short RE provided, while on the other, we can utilize ground truth (GT) bounding boxes. We also instruct the MLLM to generate the aforementioned three types of long REs. The difference lies in the input we provide to the MLLM, which includes a screenshot annotated with the GT bbox and a cropped image of the target region. This allows the model to better observe and generate high-quality REs. However, since the generation process relies on the GT, this Long-gold RE is used solely for generating training data and not for evaluation. The specific system prompt can be found in Appendix \ref{sec:prompt-long-gold}.
\end{itemize}

Since this paper primarily focuses on model perception rather than planning, and all training utilizes Long-gold RE, unless otherwise specified, all experimental results in Section \ref{sec:experiments} are based on Long RE generated by GPT-4O.

\paragraph{Metrics} In all tables within this paper, unless otherwise specified, the reported metric is click accuracy. The benchmark provides a GT bounding box, and the model being tested generates a click coordinate. If the model's output format is a bounding box, the center of the box is taken as the click coordinate. A click is considered correct if it falls within the area of the GT bounding box; otherwise, it is deemed incorrect. This method is used to calculate the accuracy.

\section{Experiments}
\label{sec:experiments}

\subsection{Input / Output Format}

\begin{figure}
    \centering
    \includegraphics[width=0.9\linewidth, trim=100 290 220 115, clip]{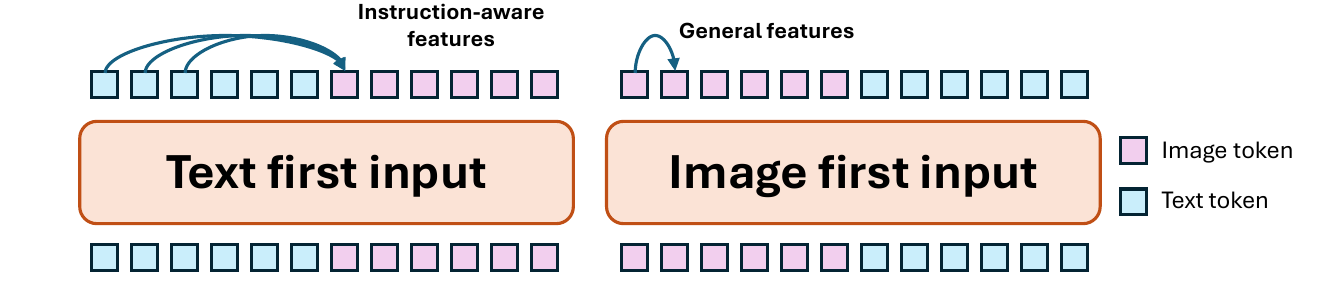}
    \caption{Illustration of the impact of modal input order on model training. }
    \label{fig:input-format}
\end{figure}

\paragraph{The order of input modality.} We investigated and experimented with the order in which text (or RE) and images are input into the model. The results displayed in Table \ref{tab:main-input-format} indicate that inputting text before images yields significantly better outcomes than the reverse order, aligning well with our expectations. Figure \ref{fig:input-format} illustrates the underlying mechanism: due to the use of causal masks in transformer decoder's attention, earlier tokens cannot be updated using later tokens. When images are input into the model first, the image tokens in the later layers, adapted through training, can be understood as being tailored for grounding tasks but unrelated to the RE. Conversely, when the order is reversed, the modeling of image tokens becomes instruction-aware. For perceptual tasks, the effectiveness of image modeling directly influences the final results. This observation bears significant resemblance to existing research related to instruction-aware models \cite{af1, af2, af3}. A simple modification of the modality order can effectively achieve this outcome.

\begin{table}[htbp]
  \centering
  \caption{Comparison of input order of modalities.}
    \begin{tabular}{c|c|ccccc}
    \toprule
    Input format & max\_crops & SSv2-Mobile & SSv2-Desktop & SSv2-Web & SS-pro & Gold-S \\
    \midrule
    \multirow{2}[2]{*}{Image first} & 6     &  85.1     & 82.7     & 81.1     &   14.6  & 84.3 \\
          & 15    &  85.4     & 83.2     & 82.6     & 27.7  & 86.1 \\
    \midrule
    \multirow{2}[2]{*}{Text first} & 6     & 87.5 (\textcolor{red}{+2.4})     & 83.2 (\textcolor{red}{+0.5})     & 83.1 (\textcolor{red}{+2.0})     & 18.7 (\textcolor{red}{+4.1}) &  86.7 (\textcolor{red}{+2.4})  \\
          & 15    & 87.1 (\textcolor{red}{+1.7})     & 85.3 (\textcolor{red}{+2.1})     & 84.2 (\textcolor{red}{+1.6})     &  30.6 (\textcolor{red}{+2.9})  &  89.2 (\textcolor{red}{+3.1}) \\
    \bottomrule
    \end{tabular}%
  \label{tab:main-input-format}%
\end{table}%

\paragraph{Output format.} We investigated the impact of output formats on results, where output formats refer to the representation of coordinates, such as bounding boxes and points. Depending on the scenario, developers might need to employ different output formats. For instance, in scenarios involving the implementation of an OS agent, grounding only requires providing the coordinates for a click, which means delivering a point. However, some developers may wish the model to generate a bounding box. For the output format of boxes, using the center point of a box can serve to determine the click position, thereby accommodating the previous scenario. We discussed the following formats and implementation strategies:
\begin{itemize}
    \item \textbf{Point+OmniParser.} The model will output text in the following format: $<point>mid_x, mid_y</point>$, where $mid_x$ and $mid_y$ represent the coordinates of the central point. In scenarios involving box output, we first utilize OmniParser-V2 \cite{omniparser} to annotate all elements within the image. Subsequently, we select the target box using the points provided by the grounding model.
    \item \textbf{XYXY.} The model will output a bounding box with format: $<box>x_1, y_1, x_2, y_2</box>$, where the $(x_1, y_1)$ and the $(x_2, y_2)$ are the top-left and bottom-right point.
    \item \textbf{XYWH.} The model will output a bounding box with format: $<box>x_1, y_1, w, h</box>$, where the $(x_1, y_1)$ is the top-left corner and $w,h$ are the width and height of the box.
    \item \textbf{MidWH.} The model will output a bounding box with format: $<box>mix_x, mid_y, w, h</box>$, where the $(mid_x, mid_y)$ is the center point of the bounding box and $w,h$ are the width and height of the box.
\end{itemize}

\begin{table}[htbp]
  \centering
  \small
  \caption{Results of training using different output formats. Note that for the point+OP format, the Click ACC reported in the table is derived directly from using the model's output point as the click target. In contrast, IoU-related metrics are calculated by selecting a box annotated by OmniParserV2 based on the model's output. The results in parentheses represent the average of the SSv2-desktop and SSv2-web subsets.}
    \begin{tabular}{ccccccccc}
    \toprule [0.1em]
    \multirow{2}[4]{*}{\textbf{Output format}} & \multicolumn{5}{c}{\textbf{ScreenSpot-V2}} & \multicolumn{3}{c}{\textbf{ScreenSpot-pro}} \\
\cmidrule(lr){2-6} \cmidrule(lr){7-9}           & Click ACC & IoU@0.3 & IoU@0.5 &  IoU@0.8 & IoU   & Click ACC & IoU@0.5 & IoU \\
    \midrule [0.1em]
    point+OP &  \textbf{85.5}     &  47.3 (61.8)  &  39.3 (53.8) &   22.0 (\textbf{30.9})  & 36.7 (48.1)   & \textbf{30.6}     &  22.3    & 18.8 \\
    \midrule
    XYXY  &   84.7     &  72.4 (71.2)   & \textbf{59.6 (59.0)} &   \textbf{26.6} (28.3)  &   \textbf{53.7 (53.7)}     & 30.0     & \textbf{23.8}     & \textbf{20.1} \\
    XYWH  &   84.2     &  73.6 (70.9)   &  58.9 (57.7)  &  22.5 (25.2)   &  52.6 (52.1)   & 29.4     & 21.8     & 18.6 \\
    MidWH &    85.0     &   \textbf{73.6 (71.5)}   &  57.1 (58.1)    &  21.7 (25.0)   &   51.6 (52.2)     &   30.4     & 22.4     & 19.2 \\
    \bottomrule [0.1em]
    \end{tabular}%
  \label{tab:main-output-format}%
\end{table}%

Results are shown in the table \ref{tab:main-output-format}. We found that directly outputting the point format achieved the best Click ACC, while the MidWH format better balanced Click ACC and the precision of the detection box. These conclusions align well with intuition. In terms of output precision for detection boxes, the XYXY format demonstrated the best performance. We observed that integrating OmniParser-V2 did not yield better results. On the one hand, the performance on the ScreenSpot-mobile subset was exceptionally poor. On other subsets, the results were still slightly lower compared to models that output box coordinates end-to-end. However, for high box precision required situations ($\text{IoU}>0.8$) and high-resolution scenarios like ScreenSpot-Pro, the gap was reduced, indicating that OmniParser's implementation may have certain advantages in high-resolution contexts and generate more accurate boxes.

\subsection{Data Augmentation}


\begin{figure}[h!]
  \begin{minipage}[t]{0.48\textwidth}
  \vspace{0.3cm}
    \begin{algorithm}[H]
      \caption{random crop}
      \begin{algorithmic}[1]
        \Require $img, box$,Croping probability: random\_crop, Minimal cropping ratio: min\_crop
        \Ensure Cropped image $img$,Updated box $box$
        \If{$\text{rand}() < \text{random\_crop}$}
            \State $(w, h) \gets img.size$
            \State $x_{left} \gets box[0]$
            \State $x_{right} \gets 1.0 - box[2]$
            \State $x_{crop\_factor} \gets \text{min\_crop} + \text{rand}() \times (1 - \text{min\_crop})$
            \State $crop_{x1} \gets w \times x_{left} \times (1 - x_{crop\_factor})$
            \State $crop_{x2} \gets w \times (box[2] + x_{right} \times x_{crop\_factor})$
            \\
            \State \text{... Same for Y axis ...}
            \\
            \State $img \gets \text{CropImage}(img, $
            \State
            $\qquad (crop_{x1}, crop_{y1}, crop_{x2}, crop_{y2}))$
            \State $box \gets \text{UpdateBox}(box)$
        \EndIf
      \end{algorithmic}

    \label{alg:random-crop}
    \end{algorithm}
  \end{minipage}
  \hfill
  \begin{minipage}[t]{0.48\textwidth}
    \begin{algorithm}[H]
  \caption{random resize and padding}
  \begin{algorithmic}[1]
    \Require $img$, Target dimensions $(W, H)$, Random resize probability: random\_resize, maximin screen size: max\_screen\_size 
    \Ensure Output image $canvas$, Updated box $box$
    \State $(w, h) \gets img.size$
    \State $canvas \gets \text{WhiteImage}(W, H)$
    \If{$\text{rand}() < \text{random\_resize}$}
        \State $S_{max} \gets \min(1.0, \frac{W}{w}, \frac{H}{h})$
        \State $S_{min} \gets \frac{W}{\text{max\_screen\_size}} \times S_{max}$
        \State $scale \gets S_{min} + \text{rand}() \times (S_{max} - S_{min})$
        \State $(\hat{w}, \hat{h}) \gets (w \times scale, h \times scale)$
        \State $img \gets \text{Resize}(img, (\hat{w}, \hat{h}))$
        \State $pos \gets (\text{randint}(0, W - \hat{w}), \text{randint}(0, H - \hat{h}))$
    \Else
        \State \text{Scale the image to fit in canvas...}
        \State $pos \gets (0,0)$ \EndIf    
    \State $canvas.\text{paste}(img, pos)$
    \State $box \gets \text{UpdateBox}(box)$
  \end{algorithmic}
\label{alg:random-resize}

\end{algorithm}
  \end{minipage}
\end{figure}

We investigated the effects of two data augmentation techniques on the training of UI grounding models:

\paragraph{Random Crop} When a user's software interface is larger than their screen, or under certain interface scaling conditions, it may result in the display of an incomplete page. Additionally, the ScreenSpot contains a small number of incomplete UI. Motivated by these observations, we introduce random crop as a data augmentation technique. We define the right edge of the bounding box and the right edge of the image as coordinates 0 and 1, respectively. We then select a random number within the range [min\_crop, 1], which determines the new right boundary of the image. The same process is applied to other directions. Notably, we use the same random number for both the left and right sides (as well as the top and bottom), ensuring that the cropping is proportional. This approach maintains the positional integrity of objects, such that if a box is located on the left side of the original image, it will remain on the left side of the cropped image. This helps avoid potential errors or changes in positional references. The detailed implementation is shown in Algorithm \ref{alg:random-crop}.

\paragraph{Random Resize} When a user reduces the software interface size or when the screen resolution is very high, items may become very small. To address this issue, we introduce a random resize data augmentation technique. The core of this data augmentation method is to shrink the image and place it onto a fixed-size white canvas. The canvas size is generally related to the design of the LMM model, particularly how it partitions the image into patches. We set a maximum screen size (e.g., 4K), and the canvas size is randomly selected between the size of the images in the training set and the maximum screen size. The purpose of this approach is to leverage the inherent size of the training set images (for instance, if the training set images are already large, excessive resizing should be avoided). For more details, please refer to Algorithm \ref{alg:random-resize}.

\begin{table}[htbp]
  \centering
  \caption{Data augmentation ablation experiments.}
    \begin{tabular}{lccccc}
    \toprule
    Data Augmentation & SSv2-desktop & SSv2-web & SSv2-mobile & SS-pro & Gold-S \\
    \midrule
    base  & 85.9     & 84.7     & 88.1     & 24.8  & 90.0 \\
    base + R-crop & 86.2     & 84.9     & 87.9     & 23.6  & 89.8 \\
    base + R-resize & 85.9     & \textbf{85.1}     & \textbf{88.3}     & \textbf{32.8} (\textcolor{red}{+8.0})  & 89.1\\
    base + R-crop + R-resize & \textbf{86.5}     & \textbf{85.1}     &   \textbf{88.3}    &  32.7 (\textcolor{red}{+7.9})  & \textbf{91.0} \\
    \bottomrule
    \end{tabular}%
  \label{tab:data-aug}%
\end{table}%

We conducted several sets of hyperparameter experiments, and the optimal combination is as follows: The probability for Random Resize is set to 100\%, with a maximum screen size of 4096, while the probability for random cropping is 0.3 with a $\text{min\_crop}=0.7$. We utilized these hyperparameters in all subsequent experiments. Table \ref{tab:data-abl} presents the results of the ablation study on data augmentation. The results indicate that in high-resolution testing environments such as ScreenSpot-pro, employing random resize significantly improves performance. On the other hand, random cropping does not have a substantial impact in various scenarios.

\subsection{Data Distribution}
\label{sec:data-dist}

\begin{figure}[h!]
  \begin{minipage}[t]{0.3\textwidth}
\begin{figure}[H]
    \centering
    \includegraphics[width=0.8\linewidth, trim=0 0 30 0, clip]{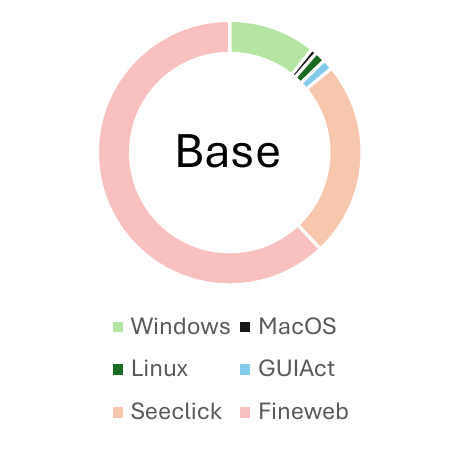}
    \caption{Data distribution.}
    \label{fig:data-ratio}
\end{figure}

  \end{minipage}
  \hfill
  \begin{minipage}[t]{0.7\textwidth}
  \vspace{0.5cm}
    \begin{table}[H]
  \centering
  \caption{Data ablation experiments. "Bing" refers to our BingSearch dataset, "CC" denotes the CommonCrawl dataset, and "CC" and "CC-re" represent the datasets before and after applying our proposed re-sampling technique, respectively.}
    \begin{tabular}{clccc}
    \toprule
    ID & Data & SSv2 & SS-pro & Gold-S \\
    \midrule
    1 & base  &   85.3   &  32.3   &  89.1  \\
    2 & $97\% $base + $3\% $Bing &   85.5   &  32.7    &   89.6  \\
    3 & $70\% $base + $30\% $CC &   \underline{85.8}   &  31.5    &   \textbf{90.0}  \\
    4 & $70\% $base + $30\% $CC-Re &   85.5   &  \underline{33.1}    &   89.6  \\
    5 & $67\% $base + $3\% $Bing + $30\% $CC-Re &   \textbf{86.1}   &  \textbf{33.3}    &   \underline{89.7}  \\
    \bottomrule
    \end{tabular}%
  \label{tab:data-abl}%
\end{table}%
  \end{minipage}
\end{figure}

\begin{figure}
    \centering
    \includegraphics[width=\linewidth, trim=10 0 0 0, clip]{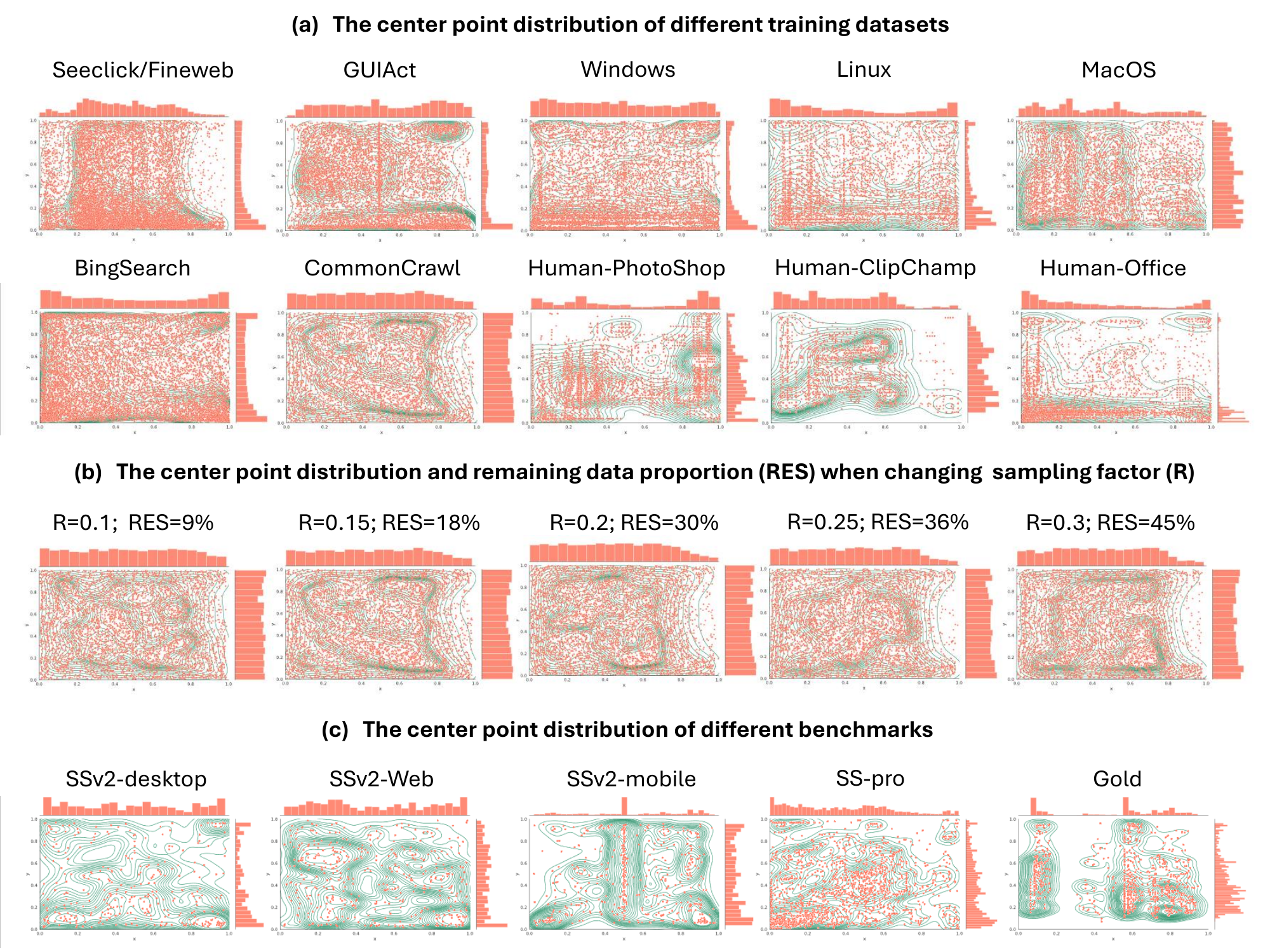}
    \caption{Center point distribution of training and evaluation data.}
    \label{fig:center-dist}
\end{figure}

We analyzed the distribution of the relative positions of the center points of all detection boxes within each dataset, as illustrated in Figure \ref{fig:center-dist}. Our findings reveal that data from different sources exhibit distinct distribution patterns. For instance, we observed that web-rendering data such as FineWeb, SeeClick, and our CommonCrawl data (when not resampled) exhibit an identical distribution pattern, as shown in the first graph of Figure \ref{fig:center-dist}(a). These datasets almost completely lack buttons on the right side, which can be attributed to the common web design practice of placing sidebars on the left and reserving the right side for scroll bars. However, such a distribution is not universal. For instance, in other datasets, the distribution tends to be more uniform. Our BingSearch demonstrates the best uniformity and diversity. Different software exhibits its unique distribution, as evidenced by our human-labeled data.

Therefore, we proposed an algorithm to resample our own CommonCrawl data. The basic idea of the algorithm is to divide the image into a $50\times 50$ grid and sample a fixed number of points from each grid cell. This approach ensures that the central points are uniformly distributed by area, as detailed in Appendix \ref{sec:re-sample-alg}. During the resampling process, we introduced a sampling factor to balance the trade-off between sampling rate and uniformity, as illustrated in Figure \ref{fig:center-dist}(b).

To validate the generated data and the effects of the aforementioned data resampling, we first constructed a base setting using open-source data, as shown in Figure \ref{fig:data-ratio}. The data proportions are roughly proportional to the size of the subset. We sampled 5M data points from this distribution to train the model. Based on this, we gradually replaced with our data, such as replacing 3\% of the BingSearch data (3\% of 5M is approximately equivalent to 1 epoch of BingSearch) or 30\% of the CommonCrawl data.

The results are shown in Table \ref{tab:data-abl}. Comparing experiments ID-1, ID-2, and ID-3, the model's performance on specific benchmarks improved after incorporating our data. Since we primarily replaced web data (Fineweb and SeeClick), this indicates that our data surpasses the web data from open-source datasets, mainly due to the greater variety of our data. Comparing experiments ID-3 and ID-4, there is a significant improvement on the ScreenSpot-pro dataset. This suggests that a uniform distribution of boxes is more beneficial for generalization in high-resolution scenarios. Finally, combining all our data achieved the best results (ID-5).

\subsection{In-domain Post-training}

\label{sec:post-training}

In this section, we will utilize our human-labeled data and explore fine-tuning strategies for target scenarios. In practical applications, developers might have a small set of target software that they wish to cover with their agent. We will use Adobe Photoshop (PS) as an example. Our human-labeled data includes screen recordings processed using the method described in Sec. \ref{sec:human-label-data}. The bounding boxes are derived from UIA-tree, while the references are sourced from GPT-4o. For our evaluation set, we focus on the PS subset of our Gold dataset, as well as the PS subset from ScreenSpot-pro. It is important to note that our Gold dataset was provided by another team, and both the bounding box and reference annotations were done by humans. Therefore, there may be discrepancies with our training set.

\paragraph{Domain Finetuning} When performing domain fine-tuning, we consider the following strategies:
\begin{itemize}
    \item Strategy A involves directly incorporating domain-specific data into the pre-trained model. 
    \item Strategy B entails performing SFT of a pre-trained model (without using domain data) using the domain data.
    \item Strategy C first introduces a small proportion of domain data during the pre-training phase, followed by using a larger proportion of domain data during the SFT phase.
\end{itemize}

We set the pre-training data size to 5M and the SFT data size to 200K. Training a subset for too many epochs (e.g., more than 10) can lead to overfitting and other issues. Thus, we limit the number of epochs of domain data for both the pre-training and SFT stages to approximately 3 epochs. Due to the difference in total data volume, the proportion of domain-specific data will vary accordingly.

\begin{table}[htbp]
  \centering
  \caption{Comparison of data selection strategies during in-domain post-training and pretraining.}
    \begin{tabular}{c|cc|cc|cc|cc}
    \toprule
    \multirow{2}[4]{*}{Strategy} & \multicolumn{2}{c|}{Pretrain} & \multicolumn{2}{c|}{Finetune} & \multicolumn{2}{c|}{General} & \multicolumn{2}{c}{Domain} \\
\cmidrule{2-9}          & base  & PS (r / ep)    & base  & PS (r / ep)    & SSv2  & SSpro & Gold-PS & SSpro-PS \\
    \midrule
    -      & 100\% & 0\% / 0   &    -   &    -   &   86.1    &    33.3   &    68.1   & 39.2  \\
    A-1     & 99.5\%    & 0.5\% / 2.2    &   -    &    -   &   \textbf{86.2}    &    32.7   &  71.8     &  41.2 \\
    A-2     & 99.0\%    & 1.0\%  / 4.5   &   -    &    -   &   86.0    &   \textbf{34.4}    &  72.7     &   41.2  \\
    \midrule
    B-1     & 100\% & 0\% / 0  & 88.0\%    & 12.0\% /  2.1   &    84.2   &   31.4    &   72.0    &  41.2 \\
    B-2     & 100\% & 0\% / 0   & 76.0\%    & 24.0\% / 4.3    &   84.4    &   30.7    &   74.2    & 41.2 \\
    C     & 99.5\%    & 0.5\% /  2.2  & 88.0\%    & 12.0\% / 2.1   &   \underline{\textit{85.8}}   &   \underline{\textit{34.2}}    &  \textbf{74.8}    &  \textbf{43.1} \\
    \bottomrule
    \end{tabular}%
  \label{tab:finetune-strategy}%
\end{table}%

The results, as shown in Table \ref{tab:finetune-strategy}, indicate that directly incorporating domain-specific data into the training set has a minimal impact on the model's general capabilities. However, the improvement in performance for the specific domain is quite limited. This may be due to the relatively small proportion of domain-specific data, which constitutes only a minor part of the optimization objective. In the case of Strategy B, where domain-specific data is introduced only during the fine-tuning phase, we observed a significant improvement in in-domain results. However, we also noted a marked decline in the model's general capabilities after fine-tuning.  This suggests that incorporating previously unseen domain data during fine-tuning may lead to overfitting and catastrophic forgetting of pre-trained knowledge.

Strategy C effectively balances general capabilities and in-domain performance. It maintains strong general abilities from the pre-training phase while also achieving the best results within the domain. During our fine-tuning process, we found that the gradient norm for Strategy C was smaller than that for Strategy B. This suggests that the decline in general capabilities observed in Strategy B may also stem from training instability caused by the introduction of new data. Since Strategy C is exposed to domain data during the pre-training phase, it experiences a more tempered impact during fine-tuning.

\paragraph{Post-training Algorithms} Beyond the impact at the data level, we also explored the effects of training algorithms during the post-training phase. We primarily considered the following categories of algorithms: Supervised Fine-Tuning, Curriculum Learning, and Reinforcement Learning. For reinforcement learning, we further examined several algorithms, including reject sampling finetuning (which is not strictly traditional reinforcement learning but has been shown by many studies \cite{rs1, rs2, rs3, rs4} to have similar characteristics and effects), DPO \cite{dpo}, REINFORCE \cite{reinforce}, PPO \cite{PPO}, REINFORCE++ \cite{reinforce++} and GRPO \cite{grpo}. However, we failed to make positive increase on results with PPO, REINFORCE++, and GRPO, primarily due to three reasons: (1) The pure perception tasks lacked textual exploration and reasoning and hence, in this situation, the essence of using RL loss will be more closely aligned with the original SFT loss.  (2) Although some previous work \cite{ui-r1, gui-g1, gui-r1} has shown that RL can provide benefits in purely perceptual tasks, these studies typically begin with relatively low baselines or apply RL directly from scratch without prior SFT. In contrast, our work involves pre-training on in-domain tasks to an optimal level, leaving little room for further optimization.  (3) The absence of exploration and reasoning led to low diversity in the answers among rollouts for the same sample, frequently resulting in all rollouts being either entirely correct or entirely incorrect (about 70\% probability). This made the training of algorithms like GRPO trivial and further hindered the effective training of the critic model in PPO. This prompted us to consider using simpler RL algorithms such as reject sampling finetuning and DPO, as well as manual intervention in the selection of rollout samples.

\begin{figure}[t!]
  \begin{minipage}[t]{0.4\textwidth}
\begin{figure}[H]
    \centering
    \includegraphics[width=\linewidth, trim=20 110 10 30, clip]{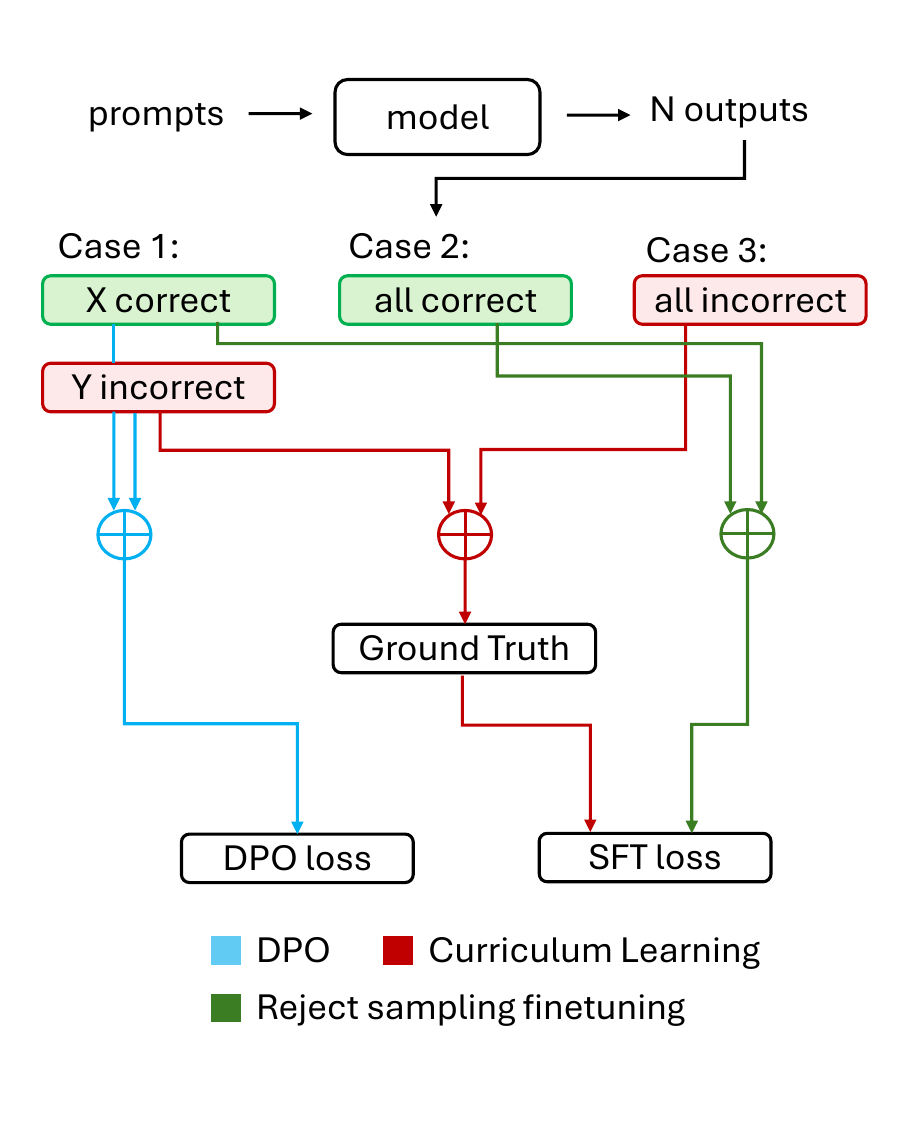}
    \caption{\textcolor{SkyBlue}{$\blacksquare$}:DPO. \textcolor{red}{$\blacksquare$}:Curriculum learning. \textcolor{ForestGreen}{$\blacksquare$}:Reject sampling finetuning. $X+Y=N$}
    \label{fig:post-training}
\end{figure}

  \end{minipage}
  \hfill
  \begin{minipage}[t]{0.6\textwidth}
  \vspace{0.5cm}
    \begin{table}[H]
      \centering
      \caption{Results of different post-training algorithms which further optimized a pre-trained model that had reached saturation. To our surprise, we found that for purely perceptual tasks, DPO could still enhance the results of the pre-trained model. However, this conclusion primarily stems from data-level adaptability and robustness, which is fundamentally different from the mechanisms of LLM reasoning.}
        \begin{tabular}{lccccc}
        \toprule[0.1em]
        \multicolumn{1}{c}{\multirow{2}[4]{*}{\textbf{Algrithm}}} & \multicolumn{3}{c}{\textbf{General}} & \multicolumn{2}{c}{\textbf{Domain}} \\
    \cmidrule(lr){2-4} \cmidrule(lr){5-6}          & \multicolumn{1}{c}{\textbf{SSv2}} & \multicolumn{1}{c}{\textbf{SSpro}} & \multicolumn{1}{c}{\textbf{Gold-S}} & \multicolumn{1}{c}{\textbf{Gold-PS}} & \multicolumn{1}{c}{\textbf{SSpro-PS}} \\
        \midrule[0.1em]
        Base Model &  86.2 & 32.7 & 88.5 & 71.8 & 41.2 \\
        \midrule
        SFT   &    85.8   &   34.2    &   89.1    &      74.8  &  43.1  \\
        Curriculum-L &    86.1   &   33.7    &    89.4   &  75.1   &   45.1  \\
        \midrule
        Reject Sampling &   \underline{\textit{86.2}}    &   34.5    &   88.6    &    75.5   &  45.1 \\
        DPO-sigmoid &   \underline{\textit{86.2}}    &   \textbf{35.2}    &   \underline{\textit{90.5}}    &   \textbf{76.8}    &  \textbf{49.0}  \\
        DPO-IPO &    \textbf{86.4}   &   \underline{\textit{34.8}}    &    90.3   &   75.5    &  43.1 \\
        DPO-NCA &   85.2    &   33.9    &    88.7   &     75.7  &  47.1  \\
        DPO-DiscoPoP &    85.4   &   \underline{\textit{34.8}}    &   \textbf{91.0}    &     76.3  & \textbf{49.0}  \\
        \bottomrule[0.1em]
        \end{tabular}%
      \label{tab:post-algo}%
    \end{table}%
\end{minipage}
\end{figure}

We employed a unified framework, as illustrated in Figure \ref{fig:post-training}, to integrate SFT, DPO, curriculum learning, and reject sampling finetuning. We ensured that the total number of samples used was consistent with the 200K samples used in SFT. Taking DPO as an example, the effective training samples only include "Case 1" in Figure \ref{fig:post-training}, resulting in less than 100K samples being used for training. We implemented DPO using the trl library \cite{trl}, which also includes several variants of DPO, such as sigmoid  \cite{dpo}(original DPO implementation), hinge \cite{hinge}, IPO \cite{IPO}, exo \cite{EXO}, nca \cite{NCA}, robust-dpo \cite{robustdpo}, SPPO \cite{SPPO}, AOT \cite{AOT}, DiscoPoP \cite{discopop}, and APO \cite{APO}. For each variant of DPO, as well as for SFT, curriculum learning, and reject sampling, we conducted hyperparameter tuning. This included a grid search for learning rates ranging from 3e-6 to 3e-4, as well as an exploration of the hyperparameters most influential in each algorithm, such as $\beta$ in DPO.

In Table \ref{tab:post-algo}, we present the results of various algorithms, including the top-performing variants of DPO. The results indicate that RL can consistently outperform SFT in purely perceptual tasks, even when starting from a highly optimized pre-trained checkpoint, which is a non-trivial conclusion. As previously mentioned, purely perceptual tasks lack text-level reasoning, exploration, and reflection. While existing studies \cite{ui-r1, gui-g1, gui-r1} have shown that reinforcement learning can enhance models that are either not pre-trained or insufficiently pre-trained, this improvement likely stems from mechanisms similar to those used by SFT. It remains uncertain whether reinforcement learning is beneficial for a model that has been pre-trained to its maximum potential.

In conjunction with the results of curriculum learning, we summarize the principles underlying the advantages of RL in post-training as follows, which are fundamentally different from those in RL work related to LLM reasoning:
\begin{itemize}
    \item Due to the targeted nature of rollouts selection by these algorithms, they exhibit greater robustness in the data distribution during post-training. The advantages of RL in post-training may stem from the adaptive data distribution selection, akin to curriculum learning.
    \item Similarly, because DPO selects samples that include both correct and incorrect outputs, it may mitigate the impact of simplistic data and, more importantly, erroneous ground truth.
    \item Compared to SFT and curriculum learning, which use ground truth coordinates for training, RL relies solely on the model it-self's outputs. This results in a more gradual and stable training process.
\end{itemize}

\subsection{Scaling}

\begin{figure}[h!]
  \begin{minipage}[t]{0.6\textwidth}
\begin{table}[H]
  \centering
  \caption{Detailed training data proportion for Phi-Ground-4B-16C with 40M training volume. $\dagger$: The number of samples here does not refer to the quantity of images or elements. In fact, because each element has three types of references—positional, appearance, and functional—we randomly combine them during training as model inputs. This combination could involve one, two, or all three types. Consequently, a single element can be paired with various references, resulting in multiple samples. This explains why the numbers here differ from those described earlier.}
   \resizebox{0.7\textwidth}{!}{
    \begin{tabular}{lccc}
    \toprule
    \textbf{Dataset} & \textbf{Samples} $\dagger$ & \textbf{Epoch} & \textbf{Weight} \\
    \midrule 
      BingSearch        &   158 K   &  7.6   & 3.0\% \\[0.3em]
       Linux   &    149 K     &   3.6    &  1.3\% \\[0.3em]
       MacOS   &      69 K    &   3.9    &  0.7\% \\[0.3em]
       Windows   &    3.5 M    &    1.1   & 10.0\% \\[0.3em]
       GUIAct   &   155 K   &   3.4    &  1.3\% \\[0.3em]
       E2ISynth   &   180 K    &   2.9    &   1.3\%  \\[0.3em]
       Fineweb   &  9.0 M   &   0.9    &  21.0 \%  \\[0.3em]
       CommonCrawl  &    25.3 M   &   1.0    &  60.0\% \\[0.3em]
       Human   &   160 K   &     3.5    &  1.4\%  \\
    \bottomrule
    \end{tabular}%
    }
  \label{tab:scaling-dataset}%
\end{table}%

  \end{minipage}
  \hfill
  \begin{minipage}[t]{0.38\textwidth}
\begin{figure}[H]
    \centering
    \includegraphics[width=\linewidth, trim=0 0 0 0, clip]{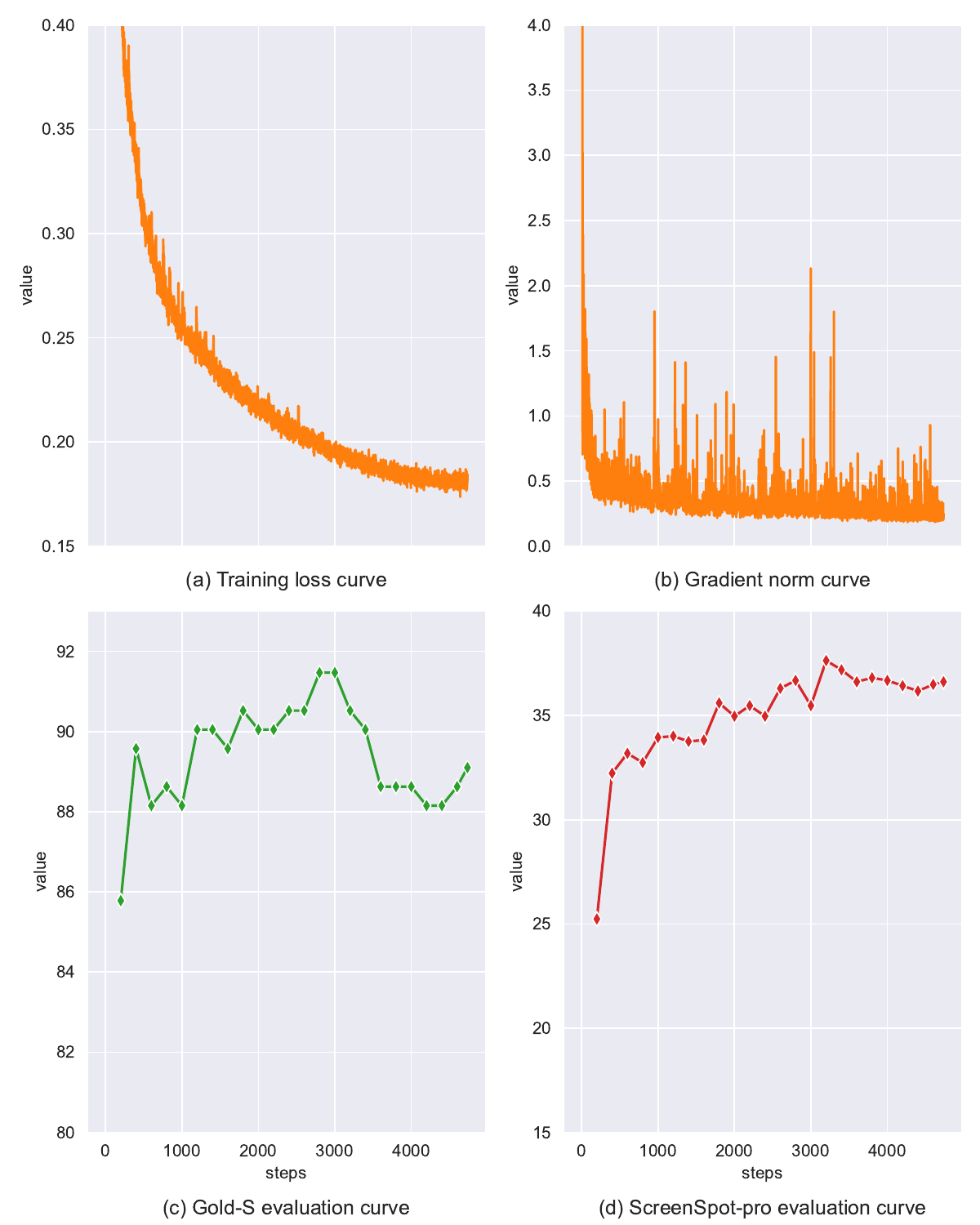}
    \vspace{-20px}
    \caption{Training and evaluation curves for Phi-Ground-4B-16C.}
    \label{fig:scaling-curves}
\end{figure}

  \end{minipage}
\end{figure}

\paragraph{Scaling settings.} We have set up the training settings for the scaling experiments by integrating all the conclusions drawn from previous sections. Specifically, we combined all available datasets at a certain ratio. Note that the Seeclick dataset was excluded after experimentation due to the excessive number of elements on average in each screenshot. For instance, in the case of 40M training, Table \ref{tab:scaling-dataset} shows the proportion of each dataset used. For other training volumes, such as 20M, the proportions are calculated accordingly based on the weights. During the training process, we further increased the batch size to 8K. As illustrated in Figure \ref{fig:scaling-curves}, the training loss curve is exceptionally smooth and stable. We also observed some fluctuations in the performance across different benchmarks during training, which led us to conduct tests at regular intervals and ultimately select the checkpoint that performed best across all benchmarks. Table \ref{tab:model-card} presents the model cards for all the models trained.

\paragraph{Scaling effect of parameters-computation trade-off.} Scaling laws \cite{sl1, sl2, sl3} typically examine the relationship between model parameters and training data given a fixed training budget. In our scenario, we further consider the computational cost during testing, which is not only related to the number of parameters but is also directly influenced by the number of image tokens. In perception tasks, the number of image tokens is highly correlated with the model's capability (see Table \ref{tab:main-input-format}). In this section, we aim to investigate the relationship among model parameters, the number of image tokens, and training data volume. This is particularly relevant in real-world applications where developers are concerned with both training performance under limited resources and application latency (rather than just parameter count). Conducting such experiments can offer more economical strategies.

\begin{table}[H]
  \centering
  \caption{Models card of Phi-Ground family.}
  \resizebox{\textwidth}{!}{
    \begin{tabular}{lllc}
    \toprule
    Name  & Model details & Training details & Image tokens \\
    \midrule
    Phi-Ground-4B-7C & \multicolumn{1}{p{26.225em}}{Pretrained with Phi-3.5-Vision-Instruct as base model, which is a 4.1B VLM with CLIP as image encoder. During training we set the num\_crops of the model to 6 and make sure the image's resolution is $2\times 3$. The model will devide the image into $2\times 3+1=7$ crops, in which there is a global crop. } & \multicolumn{1}{p{19.82em}}{Trained with 40M data; BS=8192; LR=8e-5. The training cost 200 A100 GPU days in total.} & 1045 \\
        \midrule
    Phi-Ground-4B-16C &  \multicolumn{1}{p{26.225em}}{Same with the above but change the image resolution to $3\times 5$ and num\_crops to 15. }  &  \multicolumn{1}{p{19.82em}}{Trained with 40M data; BS=8192; LR=8e-5. The training cost 440 A100 GPU days in total.} & 2353 \\    \midrule
    Phi-Ground-4B-16C-DPO &  \multicolumn{1}{p{26.225em}}{Same with  Phi-Ground-4B-16C}    & \multicolumn{1}{p{19.82em}}{Three rounds of DPO finetuning with more desktop data from Phi-Ground-4B-16C. See Sec. \ref{sec:post-training} and Table \ref{tab:exp-setting} for more details.} & 2353 \\    \midrule
    Phi-Ground-4B-29C &   \multicolumn{1}{p{26.225em}}{Same with  Phi-Ground-4B-16C except for changing the image resolution to $4\times 7$ and num\_crops to 28.}    &  \multicolumn{1}{p{19.82em}}{Trained with 20M data; BS=8192; LR=8e-5. The training cost 415 A100 GPU days in total.} & 4237\\    \midrule
    Phi-Ground-7B-7C &   \multicolumn{1}{p{26.225em}}{A variation of Phi-4-MM \cite{phi4mm} with 7B parameters.
    The model use SigLip as the image encoder.  The input image's resolution is $2\times 3$ and num\_crops of the model is 6 .}     &  \multicolumn{1}{p{19.82em}}{Trained with 30M data; BS=8192; LR=1e-5. The training cost 350 A100 GPU days in total.} & 1841 \\    \midrule
    Phi-Ground-7B-16C &  \multicolumn{1}{p{26.225em}}{Same with  Phi-Ground-7B-7C except for changing the image resolution to $3\times 5$ and num\_crops to 15.}      & \multicolumn{1}{p{19.82em}}{Trained with 15M data; BS=8192; LR=1e-5. The training cost 450 A100 GPU days in total.} & 4161 \\    \midrule
    Phi-Ground-7B-16C-DPO &  \multicolumn{1}{p{26.225em}}{Same with  Phi-Ground-7B-16C.}     &  \multicolumn{1}{p{19.82em}}{Three rounds of DPO finetuning with more desktop data from Phi-Ground-7B-16C. See Sec. \ref{sec:post-training} and Table \ref{tab:exp-setting} for more details.}  & 4161 \\ \midrule
    Phi-Ground-7B-29C &  \multicolumn{1}{p{26.225em}}{Same with  Phi-Ground-7B-7C except for changing the image resolution to $4\times 7$ and num\_crops to 28.}     & \multicolumn{1}{p{19.82em}}{Trained with 7.5M data; BS=8192; LR=1e-5. The training cost 390 A100 GPU days in total.} & 7505 \\
    \bottomrule
    \end{tabular}%
  \label{tab:model-card}%
  }
\end{table}%

Specifically, we used the 4.1B Phi-3.5-Vision-Instruct model and the 7B Phi-4-MM model as the base models for training. For each model, we configured three different image settings corresponding to varying numbers of image tokens. 
Specifically, the phi model family scales an image and crops it into a grid-shaped square, such as $336\times 336$ for Phi-3.5-V. By configuring the model's num\_crops parameter, we can adjust the maximum number of patches into which an image can be cropped. In our three image settings, we pad the images to shapes of $3\times 2$, $5\times 3$, and $7\times 4$ using white padding, and set num\_crops to 6, 15, and 28, respectively. Under these three configurations, the final images are divided into 7, 16, and 29 patches. This is because the model also includes a default global image patch.

During training, we fixed the total training budget for all models to 450 ($\pm 50$) NVIDIA A100-80G GPU days. Table \ref{tab:model-card} provides more detailed information on the image tokens and actual training durations for each model. In this manner, we trained six models under a fixed training budget. Our aim is to understand which model design can most effectively achieve optimal performance given the available computational resources.

\begin{figure}[H]
    \centering
    \vspace{-10px}
    \includegraphics[width=0.9\linewidth, trim=0 0 0 0, clip]{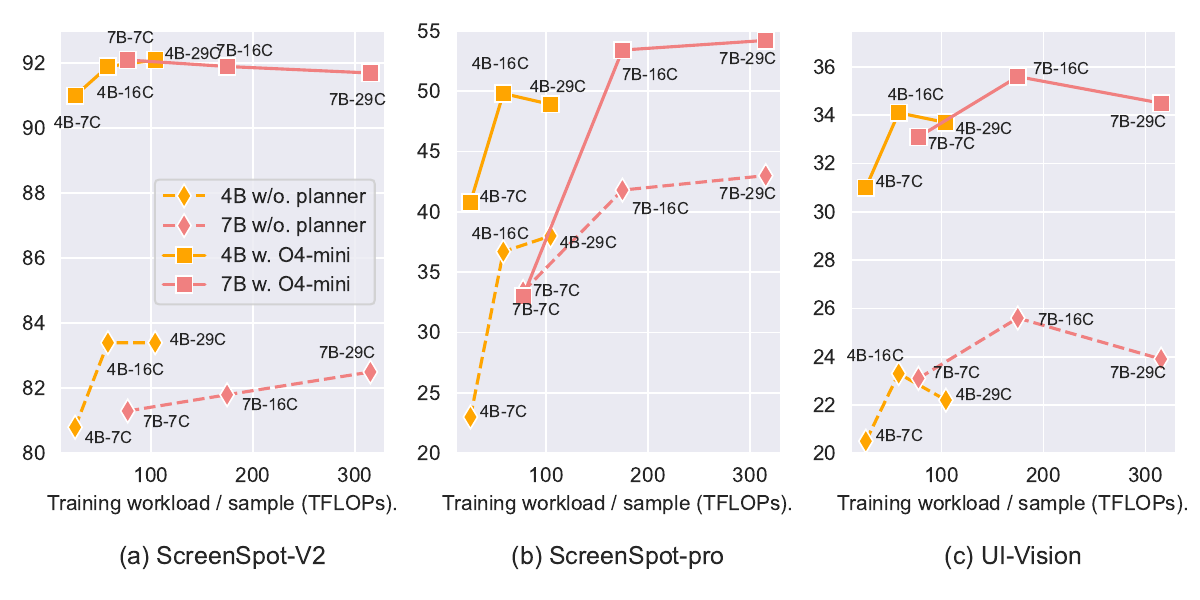}
    \caption{Illustration of the evaluation results in relation to the training computation load. The Y-axis represents the benchmark scores in click accuracy, while the X-axis denotes the training computation per sample in TFLOPs. This training computation is estimated using the formula $FLOPs = 6ND$, where $N$ is the number of image tokens and $D$ is the number of model parameters.}
    \label{fig:scaling-eval}
    \vspace{-10px}
\end{figure}

Figure \ref{fig:scaling-eval} presents the training results of these six models. Based on our evaluations, the inference time generally aligns with the relationship depicted on the x-axis of the graph. Many current studies typically report only the number of parameters when discussing model performance, without emphasizing computational aspects, such as the number of image tokens. In our experiments, where the model architecture is fixed, we observe that for more advanced and challenging benchmarks like ScreenSpot-pro and UI-Vision, the impact of image tokens is significant. Specifically, when the number of image tokens is low, it may become a bottleneck, resulting in the inability to perceive small objects and thus reducing the score. When the number of image tokens exceeds 2000, their impact gradually diminishes, meaning further increases in image tokens do not yield substantial marginal benefits akin to scaling laws.

In some test datasets that do not require high resolution, such as ScreenSpot-V2, neither the model size nor the number of image tokens significantly affects performance. Furthermore, when the number of image tokens exceeds the benchmark requirements (as previously mentioned bottleneck), the impact on perception is very limited, as illustrated by the results in the Figure \ref{fig:scaling-eval} with o4-mini as the planner. However, the difference between using a planner and not using one is quite significant.



\begin{figure}[H]
    \centering
\includegraphics[width=\linewidth, trim=0 0 0 0, clip]{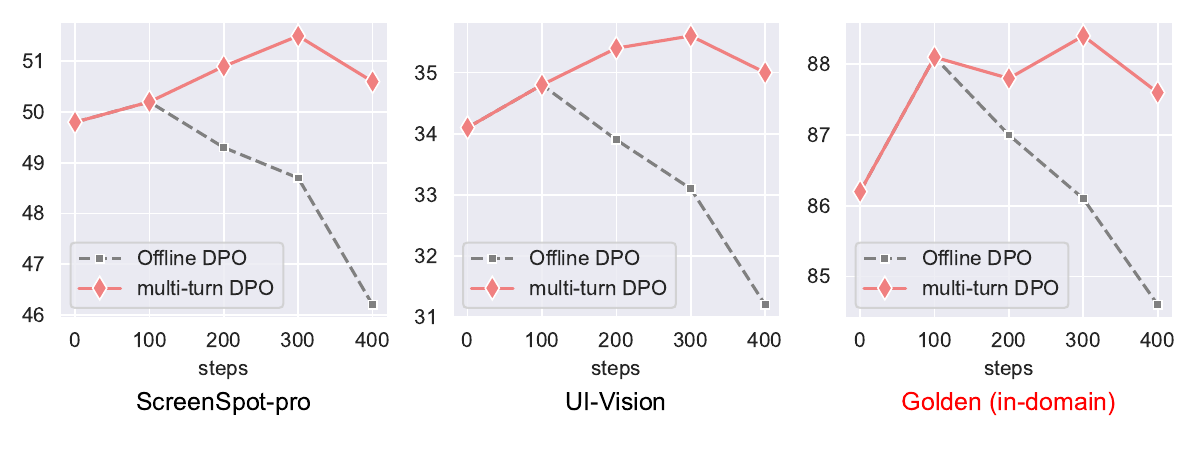}
\vspace{-20px}
    \caption{Multi-turns DPO vs. Offline DPO for in-domain post-training.}
    \label{fig:dpo-scaling-eval}
\end{figure}

\paragraph{Scaling post-training} We attempted to extend the duration of DPO post-training to observe its effects. Our findings indicate that conducting long-epoch Offline DPO training directly can lead to an initial increase in model performance followed by a decline, as illustrated by the gray line in Figure \ref{fig:dpo-scaling-eval}. We believe this phenomenon may be related to distribution shifting \cite{onlinedpo, disshift}, a concept frequently discussed in previous research. Consequently, we plan to update the rollouts more frequently. Given our stringent rollout selection criteria, we opted to conduct multiple rounds of DPO to approximate the effects of Online DPO \cite{onlinedpo} efficiently. Specifically, we performed a new rollout every 100 steps and initiated a new round of training. Ultimately, after three rounds of DPO, we achieved the best results, as shown in Figure \ref{fig:dpo-scaling-eval}.

In the post-training phase, as previously mentioned, we increased the proportion of in-domain data, which primarily consists of Web Search and human-labeled data. After training, the results for in-domain data (our Gold dataset) improved significantly. Surprisingly, several general benchmarks, such as ScreenSpot-Pro and UI-Vision, also showed noticeable improvements. This could be partly because these benchmarks have some overlap with our target software (e.g., PPT), and partly because the distribution of desktop applications is similar to that of these benchmarks. Other benchmarks not shown here maintained their original scores or experienced slight improvements after DPO, with detailed results available in the experimental tables in Appendix \ref{sec:detailed-evaluation-results}.


\section{Evaluation and Case Study}

\subsection{UI Grounding Benchmark Results}

Table \ref{tab:main-result} presents the test results of several open-source models with fewer than $10B$ parameters on our selected five GUI grounding test sets. The results in the upper block were obtained using the benchmark's built-in reference expressions, typically an instruction or short REs. In contrast, the lower block shows the results when we used o4-mini to generate Long REs, which were then tested by the grounding models.

\begin{table}[htbp]
  \centering
  \caption{The comparison of results across five GUI grounding test sets, which were tested by us, is presented. More detailed tables of results for additional open-source and closed-source models can be found in Appendix \ref{sec:detailed-evaluation-results}.}
  \resizebox{\textwidth}{!}{
    \begin{tabular}{lcccccccccccc}
    \toprule
    \multirow{2}[4]{*}{model} & \multicolumn{4}{c}{ScreenSpot-V2} & ScreenSpot-pro & \multicolumn{4}{c}{UI-Vision} & ShowDown   & \multicolumn{2}{c}{Gold} \\
\cmidrule(lr){2-5} \cmidrule(lr){6-6} \cmidrule(lr){7-10} \cmidrule(lr){11-11} \cmidrule(lr){12-13}         & Desktop & Web   & Mobile & AVG   & AVG   & basic & functional & spatial & AVG   & AVG & Gold-S & ALL \\
    \midrule
    \multicolumn{13}{c}{\textbf{\textit{End-to-end model setting (Use short REs)}}} \\
    SeeClick-9.6B \cite{seeclick} &  64.6   &   49.7    &    43.9   &    55.1     &   1.1    &    9.4   &   4.7    &   2.1    & 5.4 &   24.6 & 51.7 & 20.4 \\
    UGround-7B \cite{Uground} &  73.2   &   78.3    &   72.7    &   76.1    &   16.5    &     11.5  &   12.2    &   2.8    &  8.8     &  46.5  & 74.9 & 54.9 \\
    UGround-v1-7B \cite{Uground} &  87.1   &    86.1   &    89.4   &   87.7    &   31.1    &   15.4    &   17.1    &   6.25    &    12.9   &  57.8  &  84.4 & 66.4 \\
    OS-Atlas-4B \cite{osatlas} &  73.5   &   59.6    &    74.5   &   71.9    &   3.7    & -      &   -    &    -   &    -   & 15.8 & 47.9 & 22.0 \\
    OS-Atlas-7B \cite{osatlas} &   85.5  &    77.2   &    84.0   &    84.1   &   18.9    &  12.2     &   11.2    &    3.67   &  9.0     &  41.1& 66.4 & 48.8 \\
    UI-TARS-2B \cite{uitars} &   87.2    &   79.7    &    82.8   &   84.7    &    27.7   &    -   &     -  &   -    &   -    &  59.8 & 79.2 & 60.1 \\
    UI-TARS-7B \cite{uitars} &   \textbf{93.0}    &   \textbf{90.2}    &   89.4    &   \textbf{91.6}    &   35.7    &    20.1   &   24.3    &    8.4   &  17.6     &  66.1 & \textbf{87.2} & 76.8 \\
    UI-TARS-1.5-7B \cite{uitars15} &   86.9    &   87.6    &    \textbf{90.0}   &    89.0   &   42.6    &   28.8    &   27.5    &    \textbf{10.7}   &    22.3   &  \textbf{67.2} & 86.7 & 77.2 \\
    \textbf{Phi-Ground-4B-16C-DPO} &   84.4    &   86.4    &   78.1    &   84.1    &   38.0    &   33.4    &   34.3    &  5.8     &   24.5    &  58.2  & \textbf{87.2} & 78.2 \\
    \textbf{Phi-Ground-7B-16C-DPO} &   83.3    &    84.8   &   79.3    &   83.8    &   \textbf{43.2}    &   \textbf{36.8}    &    \textbf{37.1}   &  7.6     &   \textbf{27.2}    &  62.5  & 84.4 & \textbf{79.6} \\
    \midrule
    \multicolumn{13}{c}{\textbf{\textit{Agent setting (Use long REs) with O4-mini as planner}}} \\
    SeeClick-9.6B &  43.5   &   64.4    &    43.3   &   53.5    &   1.2    &    5.2   &   5.0    &   3.2    &   4.5    &  19.6 & 39.4 & 15.6 \\
    UGround-7B &   84.6  &   90.0    &    88.5   &   89.0    &   23.7    &     25.1  &   22.8    &   11.3    &   19.7    &  62.4 & 83.2 & 59.5 \\
    UGround-v1-7B &   89.8  &   92.2    &   91.9    &    92.1   &   32.5    &   30.5    &    29.3   &    13.9   &    24.6   &  66.7 & 88.0 & 73.8 \\
    OS-Atlas-4B &  41.9   &    64.6   &    57.7   &   57.4    &   2.0    &     -  &   -    &   -    &   -    &  18.0 & 37.0 & 18.2 \\
    OS-Atlas-7B &  80.6   &   86.2    &   88.9    &   84.7    &   21.1    &    17.2   &   16.6    &   7.5    &  13.8     &  45.5 & 68.3 & 52.2  \\
    UI-TARS-2B &    85.5   &   92.2    &   90.0    &   90.4    &   35.9    &    -   &    -   &    -   &   -    &  66.4 & 85.6 & 77.1 \\
    UI-TARS-7B &    91.5   &    93.1   &   92.9    &   93.0    &   40.6    &    33.2   &    33.4   &    16.7   &  27.8     &  69.8  & 90.9 & 81.5 \\
    UI-TARS-1.5-7B &   89.9    &    \textbf{92.7 }  &   92.0    &   92.2    &   48.8    &    35.1   &   35.1    &   17.9    &    29.4   &  71.6 & 90.4 & 80.3 \\
    \textbf{Phi-Ground-4B-16C-DPO} &    \textbf{92.8}   &   \textbf{92.7}    &    92.4   &   92.3    &    51.5   &    43.8   &   42.1    &   \textbf{21.0}    &   35.6    &  73.5 & \textbf{95.2} & \textbf{88.4} \\
    \textbf{Phi-Ground-7B-16C-DPO} &   92.6    &    92.6   &   \textbf{93.0}    &   \textbf{93.4}    &    \textbf{55.0}   &    \textbf{44.2}   &    \textbf{43.8}   &   20.5    &    \textbf{36.2}   &  \textbf{73.9} & 93.8 & 88.2 \\
    \bottomrule
    \end{tabular}%
    }
  \label{tab:main-result}%
\end{table}%

Our grounding model is trained specifically for the agent setting, meaning that the training dataset primarily consists of various combinations of REs. As a result, our model achieves significant advantages and SOTA results across all benchmarks in the agent setting. Specifically, ScreenSpot-pro achieved an accuracy of 55.0. UI-Vision also attained a result of 36.2, which is the highest for this benchmark. Furthermore, our results on the Showdown benchmark surpass those of commercial models like OpenAI Operator and Claude Computer Use (see Table \ref{tab:detailshowdown} in the Appendix).

In the end-to-end model setting, our model consistently outperforms others on the ScreenSpot-Pro, UI-Vision, and our Gold dataset, although its performance on ScreenSpot-V2 is relatively average. ScreenSpot V1 and V2 (with V2 sharing most of its data with V1) have long been crucial benchmarks for GUI grounding testing. We observed that some models perform well on ScreenSpot-V2 but do not demonstrate the same significant advantages on newly emerging benchmarks. This could be a result of developers optimizing their models based on a single benchmark over time. In our approach, we did not include any mobile data in the training set (as this is not our focus scenario), nor did we balance the training set with icon and text-based buttons (since the scenarios faced by product users are not balanced either). However, these techniques might significantly impact the accuracy on ScreenSpot-V1 and V2. Throughout our development process, the selection and ablation of each technique were carefully considered across multiple benchmarks. As a result, our model exhibits more balanced performance and better generalization.

\subsection{Error Analysis}
\label{sec:error-analyze}

\begin{figure}[H]
    \centering
\includegraphics[width=\linewidth, trim=0 30 0 0, clip]{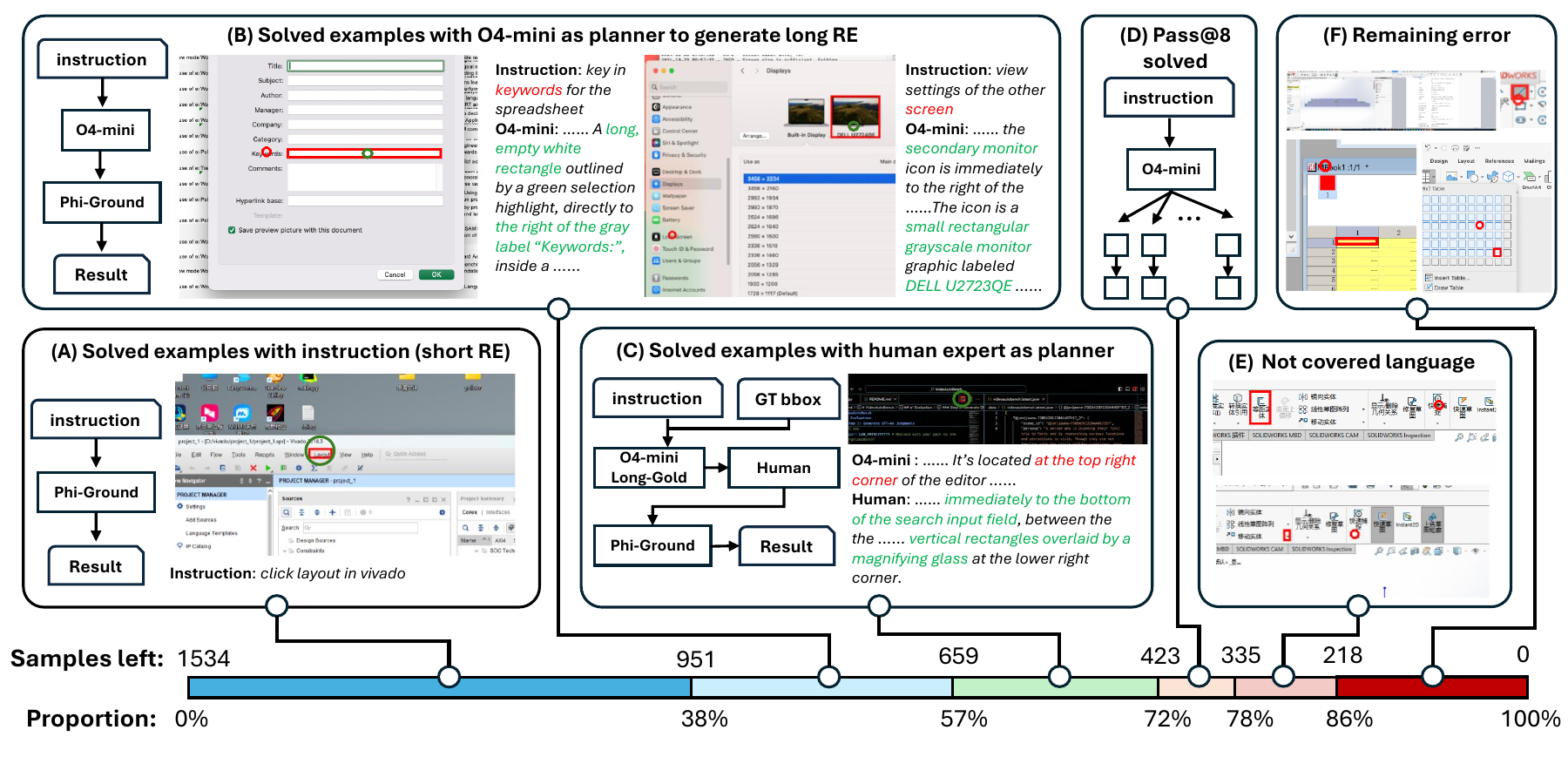}
    \caption{Types and Proportions of Errors on the ScreenSpot-pro Benchmark. In each image, the red rectangles represent the regions corresponding to the ground truth. Red circles indicate erroneous outputs from the previous stage, while green circles denote correct outputs from the current stage. The centers of the green circles fall within the ground truth boundaries. To avoid obstructing the image content, we have enlarged the green circles in some of the images.}
    \label{fig:error-proportion}
\end{figure}

To analyze and illustrate the errors made by current grounding models as a case study, we selected a challenging benchmark, ScreenSpot-Pro, as an example. We employed the Phi-Ground-4B-DPO as the grounding model and designed a cascading approach to sequentially process the benchmark data.

Specifically, as shown in Figure \ref{fig:error-proportion}, \textbf{Stage A} involves using the benchmark-provided instructions as reference expressions (also known as short REs) for the total of 1,534 test samples from ScreenSpot-Pro. As previously mentioned, our model successfully resolved 38\% of the test cases, leaving 951 incorrect samples. In \textbf{Stage B}, we used O4-mini as a planner for the remaining 951 samples to generate long REs as input for the model. This approach further resolved 292 samples, leaving 659 samples unresolved. Across both stages, we successfully addressed 57\% of the benchmark samples. In \textbf{Stage C}, for the remaining 659 samples, we intended to use human experts as planners to generate REs to observe the potential error rates in the planner section. However, this approach was deemed too costly. Therefore, we first had O4-mini generate REs using the Long-Gold method (see Sec. \ref{sec:eval-proto}), which involves disclosing the GT bbox during RE generation. Human experts then reviewed the samples, GT, and REs generated by O4-mini to correct any errors and produce the final REs for model input. This stage resolved an additional 236 samples. In \textbf{Stage D}, we recognized that for REs of similar quality, different styles and emphases might lead to varying results. Thus, we repeated the RE and grounding process of Stage C seven times for the remaining 423 samples, resolving an additional 88 samples. Ultimately, 335 samples remained with errors largely unrelated to RE quality. We will analyze the errors and their implications in each stage in detail below.

We first observe that end-to-end grounding models lack spatial reasoning capabilities, as illustrated in the example from Stage B of Figure \ref{fig:error-proportion}. When certain keywords appear in the instruction, such as "keyword" in Example 1 or "screen" in Example 2, the grounding model tends to directly highlight the locations of these words in the image. However, in Example 1, the interactive region is actually the white rectangular input box, and clicking on the text of the label might result in a failed interaction. Such spatial reasoning requires a degree of common sense, rather than being purely a grounding task. The introduction of a planner addresses this type of task effectively by directly describing "white rectangle" in the RE. We refer to such errors as "\textbf{planning omissions}," which account for 19\% of the total sample and 30.7\% of the total errors.

However, the planning of O4-mini may also encounter errors, particularly in scenarios where the target area contains multiple similar regions or when specialized application knowledge is required. In such cases, the planner's hallucinations can lead to mistakes. For instance, as illustrated in Figure \ref{fig:error-proportion} Stage C, the markdown display button is typically located in the upper right corner. However, in the example shown, two work pages are open, causing the button to be centered. This resulted in an incorrect RE by O4-mini, leading to erroneous grounding. After manually correcting the erroneous RE, our model was able to produce the correct result. We refer to this type of error as a "\textbf{planning error}," which accounts for 15.4\% of the total samples and 24.8\% of the total errors.

For the remaining samples, we observed that the grounding model might exhibit a preference for a certain style of RE, even when the quality is consistent, as evidenced by the pass@8 metric indicating new correct samples. However, this influence is minor, affecting only 5.7\% of the samples and accounting for 9.3\% of the errors. Additionally, we found that 117 samples (7.6\% of the samples and 12.3\% of the errors) were impacted because the target area or its vicinity contained languages not covered by our model, such as Chinese. Due to the strict selection of training data, primarily from the CommonCrawl dataset, which exclusively includes only English data, the model failed to correctly recognize many straightforward and easy situations due to language issues. Ultimately, 218 samples remained as particularly challenging data, making up another 14.2\% of the samples and 22.9\% of the errors.

For the remaining errors, we provide a more detailed case study in Appendix \ref{sec:more-case-study}. We categorize these errors into several identifiable types:
\begin{itemize}
    \item Accuracy issues arise due to excessively extreme screen sizes and shapes. For instance, screens with an ultra-wide aspect ratio may result in output coordinates that deviate from the intended target area.
    \item Language descriptions fail to adequately constrain areas of spatial planning, such as when generating tables and instructing to click on a blank cell in the 13th row and 8th column. Such specific spatial tasks present significant challenges for both the planner and the grounding model.
    \item Regions that are difficult to describe using natural language.
\end{itemize}

\section{Social Impacts and Open Questions}

With the development of CUA, we have both expectations and concerns regarding this direction. Primarily, there is the issue of user privacy. During our training process, we have verified the legality of the licenses for the open-source datasets used, and ensured that licenses are valid in Bing search filtering and web filtering. However, when CUA is successfully deployed in user environments in the future, the need for grounding and planning may require screenshots of users' screens to be uploaded to the cloud, potentially leading to privacy breaches. Throughout the entire research and product deployment process, we may need to establish relevant protocols, legal frameworks, or algorithms to ensure the protection of user privacy.

Secondly, there is the issue of accountability for erroneous actions performed by CUAs. There are instances where CUAs might execute irreversible and harmful operations, such as closing software without saving files or even deleting important documents. At the system level, we need to explore human-computer collaboration methods that allow CUAs to efficiently replace human labor while ensuring human oversight. From the perspective of GUI grounding, we have observed that errors due to incorrect grounding can have more severe consequences. This is because a trained grounding model, when making mistakes, still outputs interactively meaningful regions rather than blank areas, thus increasing the likelihood of irreversible impacts. For instance, the multiplication symbol on a calculator might be mistakenly interpreted as a command to close software due to similar symbols, leading to unintended software closure. Some recent studies \cite{verify1, verify2} have attempted to use MLLM to verify actions post-grounding, but these have shown limited effectiveness and increased time costs. Developing a benchmark to evaluate the potential harmfulness of GUI grounding models could also be highly beneficial.

\section{Conclusion}
In conclusion, we have developed the Phi-Ground model family, which significantly enhances GUI grounding capabilities by improving the perception of interactive elements in digital interfaces. Our comprehensive empirical study identified critical factors such as data distribution, input/output formats, and computational efficiency that influence model performance. Using a two-stage approach, we combined advanced MLLMs for generating detailed REs with a specialized grounding model for precise coordinate output, achieving state-of-the-art results across various benchmarks, including challenging ones like ScreenSpot-pro and UI-Vision. While our models demonstrate promising results, we acknowledge the societal implications of deploying CUAs, especially regarding user privacy and error accountability. Our research not only advances GUI grounding but also offers insights applicable to other multimodal perception tasks, contributing to the development of more reliable and efficient CUAs.


\begin{thebibliography}{10}

\bibitem{agent1}
Lei Wang, Chen Ma, Xueyang Feng, Zeyu Zhang, Hao Yang, Jingsen Zhang, Zhiyuan Chen, Jiakai Tang, Xu~Chen, Yankai Lin, et~al.
\newblock A survey on large language model based autonomous agents.
\newblock {\em Frontiers of Computer Science}, 2024.

\bibitem{agent3}
Zhiheng Xi, Wenxiang Chen, Xin Guo, Wei He, Yiwen Ding, Boyang Hong, Ming Zhang, Junzhe Wang, Senjie Jin, Enyu Zhou, et~al.
\newblock The rise and potential of large language model based agents: A survey.
\newblock {\em Science China Information Sciences}, 2025.

\bibitem{agent2}
Yuheng Cheng, Ceyao Zhang, Zhengwen Zhang, Xiangrui Meng, Sirui Hong, Wenhao Li, Zihao Wang, Zekai Wang, Feng Yin, Junhua Zhao, et~al.
\newblock Exploring large language model based intelligent agents: Definitions, methods, and prospects.
\newblock {\em arXiv preprint arXiv:2401.03428}, 2024.

\bibitem{cua1}
Chaoyun Zhang, Shilin He, Jiaxu Qian, Bowen Li, Liqun Li, Si~Qin, Yu~Kang, Minghua Ma, Guyue Liu, Qingwei Lin, et~al.
\newblock Large language model-brained gui agents: A survey.
\newblock {\em arXiv preprint arXiv:2411.18279}, 2024.

\bibitem{cua3}
Pascal~J Sager, Benjamin Meyer, Peng Yan, Rebekka von Wartburg-Kottler, Layan Etaiwi, Aref Enayati, Gabriel Nobel, Ahmed Abdulkadir, Benjamin~F Grewe, and Thilo Stadelmann.
\newblock A comprehensive survey of agents for computer use: Foundations, challenges, and future directions.
\newblock {\em arXiv preprint arXiv:2501.16150}, 2025.

\bibitem{robot1}
Mohsen Soori, Behrooz Arezoo, and Roza Dastres.
\newblock Artificial intelligence, machine learning and deep learning in advanced robotics, a review.
\newblock {\em Cognitive Robotics}, 2023.

\bibitem{robot2}
Demetris Vrontis, Michael Christofi, Vijay Pereira, Shlomo Tarba, Anna Makrides, and Eleni Trichina.
\newblock Artificial intelligence, robotics, advanced technologies and human resource management: a systematic review.
\newblock {\em Artificial intelligence and international HRM}, 2023.

\bibitem{o3o4}
OpenAI.
\newblock Introducing openai o3 and o4-mini, 2025.

\bibitem{claude4}
Anthropic.
\newblock Claude sonnet 4, 2025.

\bibitem{deepseekr1}
Daya Guo, Dejian Yang, Haowei Zhang, Junxiao Song, Ruoyu Zhang, Runxin Xu, Qihao Zhu, Shirong Ma, Peiyi Wang, Xiao Bi, et~al.
\newblock Deepseek-r1: Incentivizing reasoning capability in llms via reinforcement learning.
\newblock {\em arXiv preprint arXiv:2501.12948}, 2025.

\bibitem{qwen3}
An~Yang, Anfeng Li, Baosong Yang, Beichen Zhang, Binyuan Hui, Bo~Zheng, Bowen Yu, Chang Gao, Chengen Huang, Chenxu Lv, et~al.
\newblock Qwen3 technical report.
\newblock {\em arXiv preprint arXiv:2505.09388}, 2025.

\bibitem{operator}
OpenAI.
\newblock Operator system card, 2025.

\bibitem{claudeCUA}
Anthropic.
\newblock Introducing computer use, a new claude 3.5 sonnet, and claude 3.5 haiku, 2024.

\bibitem{cua2}
Michael~S Greenberg, Jennifer~C Byington, and David~G Harper.
\newblock Mobile agents and security.
\newblock {\em IEEE Communications magazine}, 1998.

\bibitem{privacy1}
Jose~M Such, Agust{\'\i}n Espinosa, and Ana Garc{\'\i}a-Fornes.
\newblock A survey of privacy in multi-agent systems.
\newblock {\em The Knowledge Engineering Review}, 2014.

\bibitem{privacy2}
Sohye Lim and Hongjin Shim.
\newblock No secrets between the two of us: Privacy concerns over using ai agents.
\newblock {\em Cyberpsychology: Journal of Psychosocial Research on Cyberspace}, 2022.

\bibitem{privacy3}
K~Cartrysse and JCA Van Der~Lubbe.
\newblock Privacy in mobile agents.
\newblock In {\em IEEE First Symposium onMulti-Agent Security and Survivability, 2004}. IEEE, 2004.

\bibitem{html1}
Evan~Zheran Liu, Kelvin Guu, Panupong Pasupat, Tianlin Shi, and Percy Liang.
\newblock Reinforcement learning on web interfaces using workflow-guided exploration.
\newblock {\em ICLR}, 2018.

\bibitem{html2}
Xiang Deng, Yu~Gu, Boyuan Zheng, Shijie Chen, Sam Stevens, Boshi Wang, Huan Sun, and Yu~Su.
\newblock Mind2web: Towards a generalist agent for the web.
\newblock {\em NIPS}, 2023.

\bibitem{aguvis}
Yiheng Xu, Zekun Wang, Junli Wang, Dunjie Lu, Tianbao Xie, Amrita Saha, Doyen Sahoo, Tao Yu, and Caiming Xiong.
\newblock Aguvis: Unified pure vision agents for autonomous gui interaction.
\newblock {\em ICML}, 2025.

\bibitem{guiact}
Wentong Chen, Junbo Cui, Jinyi Hu, Yujia Qin, Junjie Fang, Yue Zhao, Chongyi Wang, Jun Liu, Guirong Chen, Yupeng Huo, et~al.
\newblock Guicourse: From general vision language models to versatile gui agents.
\newblock {\em arXiv preprint arXiv:2406.11317}, 2024.

\bibitem{pg}
Segev Shlomov, Aviad Sela, Ido Levy, Liane Galanti, Roy Abitbol, et~al.
\newblock From grounding to planning: Benchmarking bottlenecks in web agents.
\newblock {\em arXiv preprint arXiv:2409.01927}, 2024.

\bibitem{tpsp}
Suyu Ye, Haojun Shi, Darren Shih, Hyokun Yun, Tanya Roosta, and Tianmin Shu.
\newblock Realwebassist: A benchmark for long-horizon web assistance with real-world users.
\newblock {\em arXiv preprint arXiv:2504.10445}, 2025.

\bibitem{uivision}
Shravan Nayak, Xiangru Jian, Kevin~Qinghong Lin, Juan~A Rodriguez, Montek Kalsi, Rabiul Awal, Nicolas Chapados, M~Tamer {\"O}zsu, Aishwarya Agrawal, David Vazquez, et~al.
\newblock Ui-vision: A desktop-centric gui benchmark for visual perception and interaction.
\newblock {\em arXiv preprint arXiv:2503.15661}, 2025.

\bibitem{sspro}
Kaixin Li, Ziyang Meng, Hongzhan Lin, Ziyang Luo, Yuchen Tian, Jing Ma, Zhiyong Huang, and Tat-Seng Chua.
\newblock Screenspot-pro: Gui grounding for professional high-resolution computer use.
\newblock {\em arXiv preprint arXiv:2504.07981}, 2025.

\bibitem{seeclick}
Kanzhi Cheng, Qiushi Sun, Yougang Chu, Fangzhi Xu, Li~YanTao, Jianbing Zhang, and Zhiyong Wu.
\newblock Seeclick: Harnessing gui grounding for advanced visual gui agents.
\newblock In {\em Proceedings of the 62nd Annual Meeting of the Association for Computational Linguistics (Volume 1: Long Papers)}, 2024.

\bibitem{sp}
Yuhang Liu, Pengxiang Li, Congkai Xie, Xavier Hu, Xiaotian Han, Shengyu Zhang, Hongxia Yang, and Fei Wu.
\newblock Infigui-r1: Advancing multimodal gui agents from reactive actors to deliberative reasoners.
\newblock {\em arXiv preprint arXiv:2504.14239}, 2025.

\bibitem{uitars}
Yujia Qin, Yining Ye, Junjie Fang, Haoming Wang, Shihao Liang, Shizuo Tian, Junda Zhang, Jiahao Li, Yunxin Li, Shijue Huang, et~al.
\newblock Ui-tars: Pioneering automated gui interaction with native agents.
\newblock {\em arXiv preprint arXiv:2501.12326}, 2025.

\bibitem{uitars15}
ByteDance Seed.
\newblock Ui-tars-1.5.
\newblock \url{https://seed-tars.com/1.5}, 2025.

\bibitem{Uground}
Boyu Gou, Ruohan Wang, Boyuan Zheng, Yanan Xie, Cheng Chang, Yiheng Shu, Huan Sun, and Yu~Su.
\newblock Navigating the digital world as humans do: Universal visual grounding for gui agents.
\newblock {\em ICLR}, 2025.

\bibitem{osatlas}
Zhiyong Wu, Zhenyu Wu, Fangzhi Xu, Yian Wang, Qiushi Sun, Chengyou Jia, Kanzhi Cheng, Zichen Ding, Liheng Chen, Paul~Pu Liang, et~al.
\newblock Os-atlas: A foundation action model for generalist gui agents.
\newblock {\em ICLR}, 2025.

\bibitem{tokenized1}
Zhengyuan Yang, Zhe Gan, Jianfeng Wang, Xiaowei Hu, Faisal Ahmed, Zicheng Liu, Yumao Lu, and Lijuan Wang.
\newblock Unitab: Unifying text and box outputs for grounded vision-language modeling.
\newblock In {\em ECCV}. Springer, 2022.

\bibitem{tokenized2}
Wenhai Wang, Zhe Chen, Xiaokang Chen, Jiannan Wu, Xizhou Zhu, Gang Zeng, Ping Luo, Tong Lu, Jie Zhou, Yu~Qiao, et~al.
\newblock Visionllm: Large language model is also an open-ended decoder for vision-centric tasks.
\newblock {\em NIPS}, 2023.

\bibitem{tokenized3}
Haoxuan You, Haotian Zhang, Zhe Gan, Xianzhi Du, Bowen Zhang, Zirui Wang, Liangliang Cao, Shih-Fu Chang, and Yinfei Yang.
\newblock Ferret: Refer and ground anything anywhere at any granularity.
\newblock {\em ICLR}, 2024.

\bibitem{smoothing1}
Jiasen Lu, Christopher Clark, Rowan Zellers, Roozbeh Mottaghi, and Aniruddha Kembhavi.
\newblock Unified-io: A unified model for vision, language, and multi-modal tasks.
\newblock {\em ICLR}, 2023.

\bibitem{smoothing2}
Jianfeng Wang, Zhengyuan Yang, Xiaowei Hu, Linjie Li, Kevin Lin, Zhe Gan, Zicheng Liu, Ce~Liu, and Lijuan Wang.
\newblock Git: A generative image-to-text transformer for vision and language.
\newblock {\em arXiv preprint arXiv:2205.14100}, 2022.

\bibitem{dataaug}
Parvinder Kaur, Baljit~Singh Khehra, and Er~Bhupinder~Singh Mavi.
\newblock Data augmentation for object detection: A review.
\newblock In {\em 2021 IEEE International Midwest Symposium on Circuits and Systems (MWSCAS)}. IEEE, 2021.

\bibitem{sl1}
Joel Hestness, Sharan Narang, Newsha Ardalani, Gregory Diamos, Heewoo Jun, Hassan Kianinejad, Md~Mostofa~Ali Patwary, Yang Yang, and Yanqi Zhou.
\newblock Deep learning scaling is predictable, empirically.
\newblock {\em arXiv preprint arXiv:1712.00409}, 2017.

\bibitem{sl2}
Jared Kaplan, Sam McCandlish, Tom Henighan, Tom~B Brown, Benjamin Chess, Rewon Child, Scott Gray, Alec Radford, Jeffrey Wu, and Dario Amodei.
\newblock Scaling laws for neural language models.
\newblock {\em arXiv preprint arXiv:2001.08361}, 2020.

\bibitem{sl3}
Jordan Hoffmann, Sebastian Borgeaud, Arthur Mensch, Elena Buchatskaya, Trevor Cai, Eliza Rutherford, Diego de~Las~Casas, Lisa~Anne Hendricks, Johannes Welbl, Aidan Clark, et~al.
\newblock Training compute-optimal large language models.
\newblock In {\em NIPS}, 2022.

\bibitem{e2isynth}
Xinyi Liu, Xiaoyi Zhang, Ziyun Zhang, and Yan Lu.
\newblock Ui-e2i-synth: Advancing gui grounding with large-scale instruction synthesis.
\newblock {\em arXiv preprint arXiv:2504.11257}, 2025.

\bibitem{systemoverfit1}
Hugh Zhang, Jeff Da, Dean Lee, Vaughn Robinson, Catherine Wu, William Song, Tiffany Zhao, Pranav Raja, Charlotte Zhuang, Dylan Slack, et~al.
\newblock A careful examination of large language model performance on grade school arithmetic.
\newblock {\em NIPS}, 2024.

\bibitem{systemoverfit2}
Tianwen Wei, Liang Zhao, Lichang Zhang, Bo~Zhu, Lijie Wang, Haihua Yang, Biye Li, Cheng Cheng, Weiwei L{\"u}, Rui Hu, et~al.
\newblock Skywork: A more open bilingual foundation model.
\newblock {\em arXiv preprint arXiv:2310.19341}, 2023.

\bibitem{showdown}
General~Agents Team.
\newblock The showdown computer control evaluation suite, 2025.

\bibitem{lmmpercept1}
Zhipeng Huang, Zhizheng Zhang, Yiting Lu, Zheng-Jun Zha, Zhibo Chen, and Baining Guo.
\newblock Visualcritic: Making lmms perceive visual quality like humans.
\newblock {\em arXiv preprint arXiv:2403.12806}, 2024.

\bibitem{lmmpercept2}
Rizhao Cai, Zirui Song, Dayan Guan, Zhenhao Chen, Yaohang Li, Xing Luo, Chenyu Yi, and Alex Kot.
\newblock Benchlmm: Benchmarking cross-style visual capability of large multimodal models.
\newblock In {\em ECCV}. Springer, 2024.

\bibitem{lmmpercept3}
Xingyu Fu, Yushi Hu, Bangzheng Li, Yu~Feng, Haoyu Wang, Xudong Lin, Dan Roth, Noah~A Smith, Wei-Chiu Ma, and Ranjay Krishna.
\newblock Blink: Multimodal large language models can see but not perceive.
\newblock In {\em ECCV}. Springer, 2024.

\bibitem{tokenizetext1}
Keqin Chen, Zhao Zhang, Weili Zeng, Richong Zhang, Feng Zhu, and Rui Zhao.
\newblock Shikra: Unleashing multimodal llm's referential dialogue magic.
\newblock {\em arXiv preprint arXiv:2306.15195}, 2023.

\bibitem{tokenizetext2}
Peng Wang, An~Yang, Rui Men, Junyang Lin, Shuai Bai, Zhikang Li, Jianxin Ma, Chang Zhou, Jingren Zhou, and Hongxia Yang.
\newblock Ofa: Unifying architectures, tasks, and modalities through a simple sequence-to-sequence learning framework.
\newblock In {\em ICML}. PMLR, 2022.

\bibitem{commoncrawl}
Common crawl - open repository of web crawl data, 2025.

\bibitem{omniparser}
Yadong Lu, Jianwei Yang, Yelong Shen, and Ahmed Awadallah.
\newblock Omniparser for pure vision based gui agent.
\newblock {\em arXiv preprint arXiv:2408.00203}, 2024.

\bibitem{winclick}
Zheng Hui, Yinheng Li, Tianyi Chen, Colby Banbury, Kazuhito Koishida, et~al.
\newblock Winclick: Gui grounding with multimodal large language models.
\newblock {\em arXiv preprint arXiv:2503.04730}, 2025.

\bibitem{gpt4o}
OpenAI.
\newblock Hello gpt-4o, 2024.

\bibitem{af1}
Samuel Lavoie, Polina Kirichenko, Mark Ibrahim, Mahmoud Assran, Andrew~Gordon Wilson, Aaron Courville, and Nicolas Ballas.
\newblock Modeling caption diversity in contrastive vision-language pretraining.
\newblock {\em arXiv preprint arXiv:2405.00740}, 2024.

\bibitem{af2}
Ziqiang Xu, Qi~Dai, Tian Xie, Yifan Yang, Kai Qiu, DongDong Chen, Zuxuan Wu, and Chong Luo.
\newblock Viarl: Adaptive temporal grounding via visual iterated amplification reinforcement learning.
\newblock {\em arXiv preprint arXiv:2505.15447}, 2025.

\bibitem{af3}
Sara Ghazanfari, Alexandre Araujo, Prashanth Krishnamurthy, Siddharth Garg, and Farshad Khorrami.
\newblock Emma: Efficient visual alignment in multi-modal llms.
\newblock {\em arXiv preprint arXiv:2410.02080}, 2024.

\bibitem{rs1}
Hugo Touvron, Louis Martin, Kevin Stone, Peter Albert, Amjad Almahairi, Yasmine Babaei, Nikolay Bashlykov, Soumya Batra, Prajjwal Bhargava, Shruti Bhosale, et~al.
\newblock Llama 2: Open foundation and fine-tuned chat models.
\newblock {\em arXiv preprint arXiv:2307.09288}, 2023.

\bibitem{rs2}
Walter~R Gilks and Pascal Wild.
\newblock Adaptive rejection sampling for gibbs sampling.
\newblock {\em Journal of the Royal Statistical Society: Series C (Applied Statistics)}, 41(2), 1992.

\bibitem{rs3}
Reiichiro Nakano, Jacob Hilton, Suchir Balaji, Jeff Wu, Long Ouyang, Christina Kim, Christopher Hesse, Shantanu Jain, Vineet Kosaraju, William Saunders, et~al.
\newblock Webgpt: Browser-assisted question-answering with human feedback.
\newblock {\em arXiv preprint arXiv:2112.09332}, 2021.

\bibitem{rs4}
Yuntao Bai, Andy Jones, Kamal Ndousse, Amanda Askell, Anna Chen, Nova DasSarma, Dawn Drain, Stanislav Fort, Deep Ganguli, Tom Henighan, et~al.
\newblock Training a helpful and harmless assistant with reinforcement learning from human feedback.
\newblock {\em arXiv preprint arXiv:2204.05862}, 2022.

\bibitem{dpo}
Rafael Rafailov, Archit Sharma, Eric Mitchell, Christopher~D Manning, Stefano Ermon, and Chelsea Finn.
\newblock Direct preference optimization: Your language model is secretly a reward model.
\newblock {\em NIPS}, 2023.

\bibitem{reinforce}
Ronald~J Williams.
\newblock Simple statistical gradient-following algorithms for connectionist reinforcement learning.
\newblock {\em Machine learning}, 8(3), 1992.

\bibitem{PPO}
John Schulman, Filip Wolski, Prafulla Dhariwal, Alec Radford, and Oleg Klimov.
\newblock Proximal policy optimization algorithms.
\newblock {\em arXiv preprint arXiv:1707.06347}, 2017.

\bibitem{reinforce++}
Jian Hu.
\newblock Reinforce++: A simple and efficient approach for aligning large language models.
\newblock {\em arXiv preprint arXiv:2501.03262}, 2025.

\bibitem{grpo}
Zhihong Shao, Peiyi Wang, Qihao Zhu, Runxin Xu, Junxiao Song, Xiao Bi, Haowei Zhang, Mingchuan Zhang, YK~Li, Yang Wu, et~al.
\newblock Deepseekmath: Pushing the limits of mathematical reasoning in open language models.
\newblock {\em arXiv preprint arXiv:2402.03300}, 2024.

\bibitem{ui-r1}
Zhengxi Lu, Yuxiang Chai, Yaxuan Guo, Xi~Yin, Liang Liu, Hao Wang, Han Xiao, Shuai Ren, Guanjing Xiong, and Hongsheng Li.
\newblock Ui-r1: Enhancing efficient action prediction of gui agents by reinforcement learning.
\newblock {\em arXiv preprint arXiv:2503.21620}, 2025.

\bibitem{gui-g1}
Yuqi Zhou, Sunhao Dai, Shuai Wang, Kaiwen Zhou, Qinglin Jia, and Jun Xu.
\newblock Gui-g1: Understanding r1-zero-like training for visual grounding in gui agents.
\newblock {\em arXiv preprint arXiv:2505.15810}, 2025.

\bibitem{gui-r1}
Run Luo, Lu~Wang, Wanwei He, and Xiaobo Xia.
\newblock Gui-r1: A generalist r1-style vision-language action model for gui agents.
\newblock {\em arXiv preprint arXiv:2504.10458}, 2025.

\bibitem{trl}
Leandro von Werra, Younes Belkada, Lewis Tunstall, Edward Beeching, Tristan Thrush, Nathan Lambert, Shengyi Huang, Kashif Rasul, and Quentin Gallouédec.
\newblock Trl: Transformer reinforcement learning.
\newblock \url{https://github.com/huggingface/trl}, 2020.

\bibitem{hinge}
Yao Zhao, Rishabh Joshi, Tianqi Liu, Misha Khalman, Mohammad Saleh, and Peter~J Liu.
\newblock Slic-hf: Sequence likelihood calibration with human feedback.
\newblock {\em arXiv preprint arXiv:2305.10425}, 2023.

\bibitem{IPO}
Mohammad~Gheshlaghi Azar, Zhaohan~Daniel Guo, Bilal Piot, Remi Munos, Mark Rowland, Michal Valko, and Daniele Calandriello.
\newblock A general theoretical paradigm to understand learning from human preferences.
\newblock In {\em International Conference on Artificial Intelligence and Statistics}. PMLR, 2024.

\bibitem{EXO}
Haozhe Ji, Cheng Lu, Yilin Niu, Pei Ke, Hongning Wang, Jun Zhu, Jie Tang, and Minlie Huang.
\newblock Towards efficient exact optimization of language model alignment.
\newblock In {\em ICML}, 2024.

\bibitem{NCA}
Huayu Chen, Guande He, Lifan Yuan, Ganqu Cui, Hang Su, and Jun Zhu.
\newblock Noise contrastive alignment of language models with explicit rewards.
\newblock {\em Advances in Neural Information Processing Systems}, 37:117784--117812, 2024.

\bibitem{robustdpo}
Sayak~Ray Chowdhury, Anush Kini, and Nagarajan Natarajan.
\newblock Provably robust dpo: Aligning language models with noisy feedback.
\newblock In {\em International Conference on Machine Learning}, pages 42258--42274. PMLR, 2024.

\bibitem{SPPO}
Yue Wu, Zhiqing Sun, Huizhuo Yuan, Kaixuan Ji, Yiming Yang, and Quanquan Gu.
\newblock Self-play preference optimization for language model alignment.
\newblock {\em arXiv preprint arXiv:2405.00675}, 2024.

\bibitem{AOT}
Igor Melnyk, Youssef Mroueh, Brian Belgodere, Mattia Rigotti, Apoorva Nitsure, Mikhail Yurochkin, Kristjan Greenewald, Jiri Navratil, and Jarret Ross.
\newblock Distributional preference alignment of llms via optimal transport.
\newblock {\em NIPS}, 2024.

\bibitem{discopop}
Chris Lu, Samuel Holt, Claudio Fanconi, Alex Chan, Jakob Foerster, Mihaela van~der Schaar, and Robert Lange.
\newblock Discovering preference optimization algorithms with and for large language models.
\newblock {\em NIPS}, 2024.

\bibitem{APO}
Karel D'Oosterlinck, Winnie Xu, Chris Develder, Thomas Demeester, Amanpreet Singh, Christopher Potts, Douwe Kiela, and Shikib Mehri.
\newblock Anchored preference optimization and contrastive revisions: Addressing underspecification in alignment.
\newblock {\em ACL}, 2025.

\bibitem{phi4mm}
Abdelrahman Abouelenin, Atabak Ashfaq, Adam Atkinson, Hany Awadalla, Nguyen Bach, Jianmin Bao, Alon Benhaim, Martin Cai, Vishrav Chaudhary, Congcong Chen, et~al.
\newblock Phi-4-mini technical report: Compact yet powerful multimodal language models via mixture-of-loras.
\newblock {\em arXiv preprint arXiv:2503.01743}, 2025.

\bibitem{onlinedpo}
Biqing Qi, Pengfei Li, Fangyuan Li, Junqi Gao, Kaiyan Zhang, and Bowen Zhou.
\newblock Online dpo: Online direct preference optimization with fast-slow chasing.
\newblock {\em arXiv preprint arXiv:2406.05534}, 2024.

\bibitem{disshift}
Yi~Ren and Danica~J Sutherland.
\newblock Learning dynamics of llm finetuning.
\newblock {\em ICLR}, 2024.

\bibitem{verify1}
Tiange Luo, Lajanugen Logeswaran, Justin Johnson, and Honglak Lee.
\newblock Visual test-time scaling for gui agent grounding.
\newblock {\em arXiv preprint arXiv:2505.00684}, 2025.

\bibitem{verify2}
Jungjae Lee, Dongjae Lee, Chihun Choi, Youngmin Im, Jaeyoung Wi, Kihong Heo, Sangeun Oh, Sunjae Lee, and Insik Shin.
\newblock Safeguarding mobile gui agent via logic-based action verification.
\newblock {\em arXiv preprint arXiv:2503.18492}, 2025.

\end{thebibliography}

\appendix

\section{Experiment Settings}

Due to resource constraints and the evolution of the development process, different ablation experiments may have utilized varying hyper-parameters and data configurations. We have documented the detailed setup for each ablation experiment in the table below.

\begin{table}[thbp]
  \centering
  \caption{Detailed training data configuration and hyper-parameters.}
\begin{tabular}{lll}
    \toprule
    EXP   & \multicolumn{1}{l}{Data configure} & \multicolumn{1}{l}{Hyper parameters} \\
    \midrule
    Tab. \ref{tab:main-input-format}, Tab. \ref{tab:main-output-format}  & \makecell[l]{Training samples: 2M \\ Dataset: [(SeeClick, 0.075 ep), (Fineweb, 0.062 ep), \\ (Windows, 0.088 ep), (MaxOS, 0.77 ep), \\ (Linux, 0.45 ep), (GUIAct, 0.44 ep)]} & \makecell[l]{lr=8e-5; batch size=2048; \\ weight decay=0.01; \\ random-resize=0.5\\max-grad-norm=0.1;\\
    warmup-steps=50} \\
    \midrule
    Tab. \ref{tab:data-aug}, Tab. \ref{tab:special-token}   & \makecell[l]{Training samples: 5M \\ Dataset: [(SeeClick, 0.19 ep), (Fineweb, 0.16 ep), \\ (Windows, 0.22 ep), (MaxOS, 0.77 ep), \\ (Linux, 0.65 ep), (GUIAct, 0.63 ep)] \\ (BingSearch, 0.95 ep)} & \makecell[l]{lr=8e-5; batch size=2048; \\ weight decay=0.01; \\max-grad-norm=0.1;\\
    warmup-steps=100} \\
    \midrule
    Tab. \ref{tab:data-abl}   & \makecell[l]{Training samples: 5M \\ Base: [(SeeClick, 0.19 ep), (Fineweb, 0.16 ep), \\ (Windows, 0.23 ep), (MaxOS, 0.77 ep), \\ (Linux, 0.65 ep), (GUIAct, 0.63 ep)]} & \makecell[l]{lr=8e-5; batch size=2048; \\ weight decay=0.01; \\max-grad-norm=0.1;\\
    warmup-steps=100} \\
    \midrule
    Tab. \ref{tab:finetune-strategy}   & \makecell[l]{Training samples: 5M \\ A-pretrain: [(SeeClick, 0.13 ep), (Fineweb, 0.11 ep), \\ (Windows, 0.15 ep), (MaxOS, 0.51 ep), \\ (Linux, 0.43 ep), (GUIAct, 0.52 ep),\\ (BingSearch, 0.95 ep), (CC, 0.06 ep)]} & \makecell[l]{lr=8e-5; batch size=2048; \\ weight decay=0.01; \\max-grad-norm=0.1;\\
    warmup-steps=100} \\
    \midrule
    Phi-Ground pre-train    &  In Table \ref{tab:scaling-dataset} and \ref{tab:model-card} & \makecell[l]{weight decay=0.01; \\max-grad-norm=0.1;\\
    warmup-ratio=6\%; \\ Others in Table \ref{tab:model-card}} \\
\bottomrule
    \end{tabular}%
  \label{tab:exp-setting}%
\end{table}%

\section{Detailed Evaluation Results}
\label{sec:detailed-evaluation-results}

\begin{table}[H]
  \centering
  \caption{Detailed ScreenSpot-V2 results.}
  \resizebox{0.9\textwidth}{!}{
    \begin{tabular}{lccccccc}
    \toprule
   \multirow{2}[4]{*}{\textbf{Model}} & \multicolumn{2}{c}{\textbf{Desktop}} & \multicolumn{2}{c}{\textbf{Web}} & \multicolumn{2}{c}{\textbf{Mobile}} & \multirow{2}[4]{*}{\textbf{AVG.}} \\
\cmidrule(lr){2-3} \cmidrule(lr){4-5} \cmidrule(lr){6-7}         & \textbf{Text}  & \textbf{Icon/Widget} & \textbf{Text}  & \textbf{Icon/Widget} & \textbf{Text}  & \textbf{Icon/Widget} &  \\
    \midrule
     \multicolumn{8}{c}{\textbf{\textit{End-to-end model setting (Use short REs)}}} \\
    SeeClick    &   78.4    &    50.7   &    70.1    &  29.3      &   55.2    &  32.5     &   \cellcolor[rgb]{ .867,  .922,  .969} 55.1  \\
    UGround-7B\textsuperscript{*}   &   85.1 & 61.2 & 84.6 & 71.9 & 84.3 & 61.1 &  \cellcolor[rgb]{ .867,  .922,  .969} 76.1  \\
    UGround-v1-7B\textsuperscript{*}    &   83.6 & \textbf{90.5} & 85.8 & 86.3 & 95.5 & 83.2 &  \cellcolor[rgb]{ .867,  .922,  .969} 87.7  \\
    OS-Atlas-4B   &    87.2  &  59.7  & 72.7  & 46.4 & 85.9 & 63.1 &  \cellcolor[rgb]{ .867,  .922,  .969}  71.9\\
    OS-Atlas-7B  &   95.2 & 75.8 & 90.7 & 63.6 & 90.6 &  77.3 &  \cellcolor[rgb]{ .867,  .922,  .969}  84.1 \\
\bottomrule
    \end{tabular}}
  \label{tab:detailssv2}%
\end{table}%

\begin{table}[H]
  \centering
  \caption{ScreenSpot-Pro results.}
  \resizebox{\textwidth}{!}{
    \begin{tabular}{lccccccccccccccccccccc}
    \toprule
    \multirow{2}[4]{*}{\textbf{Model}} & \multicolumn{3}{c}{\textbf{Development}} & \multicolumn{3}{c}{\textbf{Creative}} & \multicolumn{3}{c}{\textbf{CAD}} & \multicolumn{3}{c}{\textbf{Scientific}} & \multicolumn{3}{c}{\textbf{Office}} & \multicolumn{3}{c}{\textbf{OS}} & \multicolumn{3}{c}{\textbf{AVG.}} \\
\cmidrule(lr){2-4} \cmidrule(lr){5-7} \cmidrule(lr){8-10} \cmidrule(lr){11-13} \cmidrule(lr){14-16} \cmidrule(lr){17-19} \cmidrule(lr){20-22}         & \textbf{Text} & \textbf{Icon}  & \textbf{AVG.}  & \textbf{Text} & \textbf{Icon}  &\textbf{AVG.}  & \textbf{Text} & \textbf{Icon}  &\textbf{AVG.}  & \textbf{Text} & \textbf{Icon}  &\textbf{AVG.}  & \textbf{Text} & \textbf{Icon}  &\textbf{AVG.}  & \textbf{Text} & \textbf{Icon}  &\textbf{AVG.}  & \textbf{Text} & \textbf{Icon} & \textbf{AVG.} \\
    \midrule
    \multicolumn{22}{c}{\textbf{\textit{End-to-end model setting (Use short REs)}}} \\
       SeeClick \cite{seeclick}  & 0.6 & 0.0 &  0.3 & 1.0 & 0.0 & 0.6 &   2.5    &   0.0    &   1.9    &   3.5    &  0.0     &   2.0    &    1.1   &    0.0   &   0.9    &   2.8    &   0.0    &    1.5   &   1.8    &    0.0   & \cellcolor[rgb]{ .867,  .922,  .969} 1.1 \\
       OS-Atlas-4B \cite{osatlas} & 7.1 & 0.0 & 3.7 & 3.0 & 1.4 &  2.3 &  2.0     &   0.0    &    1.5   &   9.0    &   5.5    &    7.5   &   5.1    &    3.8   &   4.8    &   5.6    &   0.0    &    3.1   &   5.0    &   1.7    & \cellcolor[rgb]{ .867,  .922,  .969} 3.7 \\
        Show-UI-2B & 16.9 & 1.4 & 9.4 &  9.1  &  0.0 &  5.3  &    2.5   &    0.0   &   1.9    &   13.2    &   7.3    &  10.6     &   15.3    &    7.5   &   13.5    &   10.3    &   2.2    &   6.6    &   10.8    &   2.6    & \cellcolor[rgb]{ .867,  .922,  .969} 7.7 \\
\bottomrule
    \end{tabular}}
  \label{tab:detailsspro}%
\end{table}%

\begin{table}[H]
    \centering
    \resizebox{\textwidth}{!}{
    \begin{tabular}{l *{22}{c}}
    \toprule
         & \multicolumn{7}{c}{\textbf{Basic}} 
         & \multicolumn{7}{c}{\textbf{Functional}} 
         & \multicolumn{7}{c}{\textbf{Spatial}} 
         &  \\[1ex]
    \cmidrule(lr){2-8} \cmidrule(lr){9-15} \cmidrule(lr){16-22}
    \textbf{Model} 
        & Ed & Br & De & Pr & Cr & En & Overall 
         & Ed & Br & De & Pr & Cr & En & Overall  
         & Ed & Br & De & Pr & Cr & En & Overall 
         &  \textbf{Final Avg} \\[0.5ex] 
         & (215) & (56) & (376) & (605) & (438) & (82) & (1772)  
         & (215) & (56) & (376) & (605) & (438) & (82) & (1772)  
         & (212) & (31) & (338) & (740) & (586) & (28) & (1935) 
         &  \\[1ex] 
    \midrule

\multicolumn{23}{c}{\textit{\textbf{Closed-Source VLMs}}} \\

{GPT-4o}  
     & 2.23 
     & 0.00 
     & 1.86 
     & 1.16 
     & 1.14 
     & \underline{4.88} 
     & {1.58}
     & \underline{1.40} 
     & 0.00 
     & 3.19 
     & 0.83 
     & 0.91 
     & 3.66
     & {1.52}
     & 0.94 
     & 0.00 
     & 1.48 
     & 1.22 
     & 0.51 
     & 3.57 
     & {1.03} 
     & \cellcolor[rgb]{ .867,  .922,  .969}{1.38} \\[1ex]

{Gemini-1.5-pro} 
     & 0.47 
     & 0.00 
     & 1.60 
     & 0.83 
     & 0.46 
     & 0.00  
     & {0.79}
     & 0.00 
     & 1.79 
     & 0.27 
     & 0.17 
     & 0.46 
     & 0.00 
     & {0.28}
     & 0.94 
     & 0.00 
     & 0.89 
     & 0.54 
     & 0.34 
     & 0.00 
     & {0.57}
     & \cellcolor[rgb]{ .867,  .922,  .969}{0.55} \\[1ex]

{Gemini-Flash-2.0}  
     & 0.00 
     & 0.00 
     & 0.27 
     & 0.66 
     & 0.68 
     & 0.00 
     & {0.45}
     & 0.47 
     & 1.79 
     & 0.00 
     & 0.66 
     & 0.23 
     & 0.00 
     & {0.40}
     & 0.00 
     & 0.00 
     & 0.00 
     & 0.14 
     & 0.00 
     & 0.00 
     & {0.05}
     & \cellcolor[rgb]{ .867,  .922,  .969}{0.30} \\
\bottomrule

    \end{tabular}
    }
        \caption{Results of UI-Vision \cite{uivision}. The final column shows the overall average. Abbreviated category labels: Ed (Education), Br (Browsers), De (Development), Pr (Productivity), Cr (Creativity), En (Entertainment). The best model within each size category is highlighted in \textbf{bold}, and the runner-up is \underline{underlined}. We tested Agent setting results of UI-Vision of several open-source GUI models ($\leq 10B$ parameters). }
    \label{tab:UI-vision-long}
\end{table}

\begin{table}[H]
  \centering
  \caption{Showdown-click-dev results. $\dagger$: For the latency of the models we tested, we report the inference speed of the models accelerated using the vllm Python library if supported, otherwise we report the latency using huggingface transformers, marked with 'hf'. This may be faster than the results provided by the benchmark itself, but the comparison between the models we tested remains fair. For the settings with GPT-4O and O4-mini as planners, we directly added 2.5 seconds (aligned with the original benchmark) and 8 seconds (our tested average level, which may be highly dependent on the endpoint) to the original model latency, respectively. *: Results from the original GitHub repository.}
    \resizebox{\textwidth}{!}{\begin{tabular}{lcc|lcc}
    \toprule
    \textbf{Model} & \textbf{Accuracy(\%) $\uparrow$} & \textbf{Latency$^{\dagger}$(ms) $\downarrow$} & \textbf{Model} & \textbf{Accuracy(\%) $\uparrow$} & \textbf{Latency$^{\dagger}$(ms) $\downarrow$} \\
    \midrule
    \multicolumn{6}{c}{\textit{\textbf{End-to-end model setting (Use short REs)}}} \\
    GPT-4O\textsuperscript{*} &    \cellcolor[rgb]{ .867,  .922,  .969} 5.21  &   2500    & UI-TARS-1.5-7B &   \cellcolor[rgb]{ .867,  .922,  .969} \textbf{67.15} & 445  \\
    Qwen2.5-VL-73B-Instruct\textsuperscript{*} &   \cellcolor[rgb]{ .867,  .922,  .969} 24.78   &   3790    & UGround-7B &  \cellcolor[rgb]{ .867,  .922,  .969} 46.50 & 1871 (hf) \\
    Gemini 2.0 Flash\textsuperscript{*} &    \cellcolor[rgb]{ .867,  .922,  .969}  33.39 &   3069    & UGround-v1-7B &    \cellcolor[rgb]{ .867,  .922,  .969} 57.81 & 209  \\
    UI-TARS-72B-SFT\textsuperscript{*} &   \cellcolor[rgb]{ .867,  .922,  .969}  54.40  &   1977    & \textbf{Phi-Ground-4B-7C} &    \cellcolor[rgb]{ .867,  .922,  .969} 54.40  &  122 \\
    Claude 3.7 Sonnet (Computer Use)\textsuperscript{*} &    \cellcolor[rgb]{ .867,  .922,  .969} 53.68  &  9656     & \textbf{Phi-Ground-4B-16C} &    \cellcolor[rgb]{ .867,  .922,  .969} 59.96  &  212 \\
    Molmo-72B-0924\textsuperscript{*} &    \cellcolor[rgb]{ .867,  .922,  .969} 54.76  &    6599   & \textbf{Phi-Ground-4B-16C-DPO} &   \cellcolor[rgb]{ .867,  .922,  .969} 58.17   &  212 \\
    Operator (OpenAI CUA)\textsuperscript{*} &    \cellcolor[rgb]{ .867,  .922,  .969} 64.27  &    6385   & \textbf{Phi-Ground-4B-29C} &   \cellcolor[rgb]{ .867,  .922,  .969}  55.11  &  401 \\
    seeclick &     \cellcolor[rgb]{ .867,  .922,  .969} 24.60 & 847 (hf)    & \textbf{Phi-Ground-7B-7C} &   \cellcolor[rgb]{ .867,  .922,  .969} 57.45   &  168  \\
    OS-ATLAS-4B &    \cellcolor[rgb]{ .867,  .922,  .969} 15.80 & 1288  (hf)  & \textbf{Phi-Ground-7B-16C} &     \cellcolor[rgb]{ .867,  .922,  .969} 61.04 &  313 \\
    OS-ATLAS-7B &   \cellcolor[rgb]{ .867,  .922,  .969} 41.11 & 1788  (hf)  & \textbf{Phi-Ground-7B-16C-DPO} &    \cellcolor[rgb]{ .867,  .922,  .969} 62.48  & 313 \\
    UI-TARS-2B &   \cellcolor[rgb]{ .867,  .922,  .969} 59.78 & 186     & \textbf{Phi-Ground-7B-29C} &  \cellcolor[rgb]{ .867,  .922,  .969}  61.22   & 603 \\
    UI-TARS-7B &   \cellcolor[rgb]{ .867,  .922,  .969} \textit{\underline{66.07}} & 237    &       &       &  \\
    
    \midrule
    \multicolumn{6}{c}{\textit{\textbf{Agent setting (Use long REs) with GPT-4O as planner}}} \\
    
    seeclick &   \cellcolor[rgb]{ .867,  .922,  .969} 15.62 &  3347 (hf)  & \textbf{Phi-Ground-4B-7C} &    \cellcolor[rgb]{ .867,  .922,  .969} 60.68  & 2622 \\
    OS-ATLAS-4B &  \cellcolor[rgb]{ .867,  .922,  .969} 13.46 & 3788 (hf)    & \textbf{Phi-Ground-4B-16C} &   \cellcolor[rgb]{ .867,  .922,  .969}  62.84  & 2712 \\
    OS-ATLAS-7B &   \cellcolor[rgb]{ .867,  .922,  .969} 40.22 & 4288 (hf)    & \textbf{Phi-Ground-4B-16C-DPO} &    \cellcolor[rgb]{ .867,  .922,  .969} 62.59  & 2712 \\
    UI-TARS-2B &   \cellcolor[rgb]{ .867,  .922,  .969} 58.89 & 2687    & \textbf{Phi-Ground-4B-29C} &    \cellcolor[rgb]{ .867,  .922,  .969} 61.04  & 2901 \\
    UI-TARS-7B &  \cellcolor[rgb]{ .867,  .922,  .969} 61.58 & 2745    & \textbf{Phi-Ground-7B-7C} &   \cellcolor[rgb]{ .867,  .922,  .969} 59.25   & 2668 \\
    UI-TARS-1.5-7B &   \cellcolor[rgb]{ .867,  .922,  .969} 61.40 & 2950     & \textbf{Phi-Ground-7B-16C} &     \cellcolor[rgb]{ .867,  .922,  .969} \textit{\underline{63.02}} & 2813 \\
    UGround-7B &   \cellcolor[rgb]{ .867,  .922,  .969} 52.96 &  4371 (hf)     & \textbf{Phi-Ground-7B-16C-DPO} &    \cellcolor[rgb]{ .867,  .922,  .969} \textbf{64.39}  &  2813 \\
    UGround-v1-7B &  \cellcolor[rgb]{ .867,  .922,  .969} 57.99 & 2717    & \textbf{Phi-Ground-7B-29C} &   \cellcolor[rgb]{ .867,  .922,  .969}  61.93  & 3103 \\
    
    \midrule
    \multicolumn{6}{c}{\textit{\textbf{Agent setting (Use long REs) with O4-mini as planner}}} \\
    
    seeclick &   \cellcolor[rgb]{ .867,  .922,  .969} 19.60 & 8847 (hf)    & \textbf{Phi-Ground-4B-7C} &    \cellcolor[rgb]{ .867,  .922,  .969} 69.60  & 8122 \\
    OS-ATLAS-4B &   \cellcolor[rgb]{ .867,  .922,  .969} 17.99 & 9288 (hf)    & \textbf{Phi-Ground-4B-16C} &   \cellcolor[rgb]{ .867,  .922,  .969}  72.12  &  8212 \\
    OS-ATLAS-7B &   \cellcolor[rgb]{ .867,  .922,  .969} 45.50 & 9788 (hf)     & \textbf{Phi-Ground-4B-16C-DPO} &    \cellcolor[rgb]{ .867,  .922,  .969}  \textit{\underline{73.51}} & 8212  \\
    UI-TARS-2B &    \cellcolor[rgb]{ .867,  .922,  .969} 66.37 & 8188     & \textbf{Phi-Ground-4B-29C} &    \cellcolor[rgb]{ .867,  .922,  .969}  69.96 & 8401 \\
    UI-TARS-7B &   \cellcolor[rgb]{ .867,  .922,  .969} 69.78 & 8243    & \textbf{Phi-Ground-7B-7C} &    \cellcolor[rgb]{ .867,  .922,  .969} 71.94  & 8168 \\
    UI-TARS-1.5-7B &  \cellcolor[rgb]{ .867,  .922,  .969} 71.58 & 8454   & \textbf{Phi-Ground-7B-16C} &    \cellcolor[rgb]{ .867,  .922,  .969}  72.12 & 8313 \\
    UGround-7B &   \cellcolor[rgb]{ .867,  .922,  .969} 62.41 & 9871 (hf)     & \textbf{Phi-Ground-7B-16C-DPO} &  \cellcolor[rgb]{ .867,  .922,  .969}  \textbf{73.87}   &  8313 \\
    UGround-v1-7B &   \cellcolor[rgb]{ .867,  .922,  .969} 66.73 & 8214     & \textbf{Phi-Ground-7B-29C} &    \cellcolor[rgb]{ .867,  .922,  .969} 71.40  & 8603 \\
    \bottomrule
    \end{tabular}}
  \label{tab:detailshowdown}%
\end{table}%

\begin{table}[H]
  \centering
  \caption{Gold dataset evaluation results.}
  \resizebox{\textwidth}{!}{
    \begin{tabular}{lcccccccc}
    \toprule
    \multirow{2}[4]{*}{model} & \multicolumn{7}{c}{Gold dataset}           & \multirow{2}[4]{*}{Gold-S} \\
\cmidrule(lr){2-8}          & PhotoShop & ClipChamp & Excel & PowerPoint & Word  & Windows Setting & AVG. & \\
    \midrule
    \multicolumn{9}{c}{\textbf{\textit{End-to-end model setting (Use short REs)}}} \\
    SeeClick-10B \cite{showdown} & 1.41 & 14.53 & 21.50 & 13.41 & 11.63 & 59.76 &  \cellcolor[rgb]{ .867,  .922,  .969} 20.37   &  51.66 \\
    UGround-7B \cite{Uground} & 29.11 & 65.36 & 48.60 & 58.54 & 41.86 & 86.01 &  \cellcolor[rgb]{ .867,  .922,  .969} 54.91   & 74.88 \\
    UGround-v1-7B \cite{Uground} & 49.30 & \textbf{79.33} & 57.01 & 69.51 & 47.67 & 95.73 &  \cellcolor[rgb]{ .867,  .922,  .969} 66.42   & 84.36 \\
    OS-ATLAS-4B \cite{osatlas} & 5.63 & 21.23 & 14.95 & 13.41 & 16.28 & 60.55 &  \cellcolor[rgb]{ .867,  .922,  .969} 22.01   & 47.87 \\
    OS-ATLAS-7B \cite{osatlas} & 32.86 & 70.39 & 27.10 & 31.71 & 39.53 & 90.91 &  \cellcolor[rgb]{ .867,  .922,  .969} 48.75   & 66.35 \\
    UI-TARS-2B \cite{uitars} & 46.95 & 67.60 & 55.14 & 57.32 & 48.84 & 84.90 &  \cellcolor[rgb]{ .867,  .922,  .969} 60.12  & 79.15 \\
    UI-TARS-7B \cite{uitars} & 56.34 & \textit{\underline{76.54}} & 72.90 & 85.37 & 75.58 & 93.75 &  \cellcolor[rgb]{ .867,  .922,  .969} 76.75  & \textbf{87.20} \\
    UI-TARS-1.5-7B \cite{uitars15} & 63.85 & 65.92 & \textbf{83.18} & 80.49 & 81.40 & 88.62 &  \cellcolor[rgb]{ .867,  .922,  .969} 77.24  & 86.73 \\
    \textbf{Phi-Ground-4B-7C} &   46.01 & 62.57 & 52.34 & 65.85 & 58.14 & 91.86   &  \cellcolor[rgb]{ .867,  .922,  .969} 62.8 &  76.78   \\
    \textbf{Phi-Ground-4B-16C} &   58.22 & 73.18 & 76.64 & 81.71 & 76.74 & 93.28  &  \cellcolor[rgb]{ .867,  .922,  .969} 76.63 &  85.78   \\
    \textbf{Phi-Ground-4B-16C-DPO} &    64.79 & 68.72 & 78.50 & 87.80 & 75.58 & 93.99    &  \cellcolor[rgb]{ .867,  .922,  .969} 78.23  &  \textbf{87.20}   \\
    \textbf{Phi-Ground-4B-29C} &    56.81 & 68.72 & \textit{\underline{80.37}} & 87.80 & 73.26 & 94.15   &  \cellcolor[rgb]{ .867,  .922,  .969} 76.85 &   83.41  \\
    \textbf{Phi-Ground-7B-7C} &    57.28 & 69.83 & 66.36 & 81.71 & 68.60 & 91.94   &  \cellcolor[rgb]{ .867,  .922,  .969} 72.62 &  85.31   \\
    \textbf{Phi-Ground-7B-16C} &  \textit{\underline{69.95}} & 69.83 & 77.57 & 87.8 & \textbf{77.91} & 93.28    &  \cellcolor[rgb]{ .867,  .922,  .969} \textit{\underline{79.39}} &  84.83   \\
    \textbf{Phi-Ground-7B-16C-DPO} &   \textbf{72.30} & 69.27 & 72.90 & \textbf{90.24} & \textbf{77.91} & \textit{\underline{95.18}}  &  \cellcolor[rgb]{ .867,  .922,  .969} \textbf{79.63} &  84.36   \\
    \textbf{Phi-Ground-7B-29C} &    67.61 & 67.60 & 79.44 & \textbf{90.24} & 72.09 & \textbf{96.28}   &  \cellcolor[rgb]{ .867,  .922,  .969} 78.88 &  84.83   \\
    
    \midrule
    \multicolumn{9}{c}{\textbf{\textit{Agent setting (Use long REs) with GPT-4O as planner}}} \\
    
    SeeClick-10B & 1.41 & 9.50 & 11.21 & 6.10 & 2.33 & 48.46 &  \cellcolor[rgb]{ .867,  .922,  .969} 13.17   & 32.23 \\
    UGround-7B & 38.97 & 64.25 & 57.94 & 56.10 & 53.49 & 86.88 &  \cellcolor[rgb]{ .867,  .922,  .969} 59.60  & 81.52 \\
    UGround-v1-7B &  52.11 & 73.18 & 67.29 & 74.39 & 62.79 & 95.57 &  \cellcolor[rgb]{ .867,  .922,  .969} 70.89   & 84.83 \\
    OS-ATLAS-4B &  7.98 & 18.44 & 15.89 & 15.85 & 10.47 & 61.34 &  \cellcolor[rgb]{ .867,  .922,  .969} 21.66   &  38.39 \\
    OS-ATLAS-7B & 31.92 & 62.57 & 26.17 & 41.46 & 50.00 & 87.67 &  \cellcolor[rgb]{ .867,  .922,  .969} 49.97   &  67.30 \\
    UI-TARS-2B & 55.40 & 76.54 & 71.96 & 80.49 & 69.77 & 93.91 &  \cellcolor[rgb]{ .867,  .922,  .969} 74.68  & 87.20 \\
    UI-TARS-7B &  62.91 & 74.86 & 78.50 & 82.93 & 81.40 & 94.39 &  \cellcolor[rgb]{ .867,  .922,  .969} 79.17 &  85.31 \\
    UI-TARS-1.5-7B & 66.67 & 62.57 & 85.05 & 84.15 & \textbf{87.21} & 86.88 &  \cellcolor[rgb]{ .867,  .922,  .969} 78.76  &  88.63 \\
    \textbf{Phi-Ground-4B-7C} &    65.26 & 71.51 & 68.22 & 78.05 & 75.58 & 96.76   &  \cellcolor[rgb]{ .867,  .922,  .969} 75.90 &   88.63  \\
    \textbf{Phi-Ground-4B-16C} &   70.89   &   \textbf{78.77}    &  84.11     &   86.59    &  83.72     &  96.92    &  \cellcolor[rgb]{ .867,  .922,  .969} 83.50 &  \textbf{91.47}   \\
    \textbf{Phi-Ground-4B-16C-DPO} &  \textbf{76.06}    &   \textit{\underline{77.65}}    &   \textit{\underline{87.85}}    &   89.02    &   84.88    &   \textit{\underline{97.71}}   &  \cellcolor[rgb]{ .867,  .922,  .969} \textbf{85.52} &   \textit{\underline{91.00}}  \\
    \textbf{Phi-Ground-4B-29C} &    70.89 & 70.95 & 78.50 & \textbf{91.46} & 80.23 & 96.36     &  \cellcolor[rgb]{ .867,  .922,  .969} 81.40  &  89.57   \\
    \textbf{Phi-Ground-7B-7C} &   69.48 & 75.42 & 75.70 & 87.80 & 74.42 & 97.15  &  \cellcolor[rgb]{ .867,  .922,  .969} 80.00 &  88.63   \\
    \textbf{Phi-Ground-7B-16C} &  73.24    &   75.42    &   \textbf{89.72}    &   \textbf{91.46}    &   81.40    &   96.36   &  \cellcolor[rgb]{ .867,  .922,  .969} 84.60  &   90.05  \\
    \textbf{Phi-Ground-7B-16C-DPO} &  \textbf{76.06} & 75.98 & 86.92 & 89.02 & \textit{\underline{86.05}} & \textbf{97.94}  &  \cellcolor[rgb]{ .867,  .922,  .969} \textit{\underline{85.33}}  &  \textit{\underline{91.00}}   \\
    \textbf{Phi-Ground-7B-29C} &    71.83 & 73.18 & 84.11 & 89.02 & 77.91 & 97.00    &  \cellcolor[rgb]{ .867,  .922,  .969} 82.17 &   88.15  \\

    \midrule
    \multicolumn{9}{c}{\textbf{\textit{Agent setting (Use long REs) with O4-mini as planner}}} \\

    SeeClick-10B & 2.35 & 20.11 & 9.43 & 4.94 & 6.90 & 49.72 &  \cellcolor[rgb]{ .867,  .922,  .969} 15.57   & 39.42 \\
    UGround-7B & 39.44 & 70.39 & 56.60 & 54.32 & 48.28 & 87.75 &  \cellcolor[rgb]{ .867,  .922,  .969} 59.46   & 83.17 \\
    UGround-v1-7B & 52.11 & 77.09 & 67.92 & 80.25 & 68.60 & 96.76 &  \cellcolor[rgb]{ .867,  .922,  .969} 73.79   &  87.98 \\
    OS-ATLAS-4B &  6.57 & 17.88 & 6.60 & 16.05 & 9.20 & 53.04 &  \cellcolor[rgb]{ .867,  .922,  .969} 18.22   & 37.02 \\
    OS-ATLAS-7B & 35.68 & 65.92 & 33.96 & 39.51 & 49.43 & 88.62 &  \cellcolor[rgb]{ .867,  .922,  .969} 52.19    &  68.27 \\
    UI-TARS-2B &  57.28 & 82.12 & 73.58 & 81.48 & 73.26 & 94.86 &  \cellcolor[rgb]{ .867,  .922,  .969} 77.10  & 85.58 \\
    UI-TARS-7B & 67.14 & 82.12 & 77.36 & 88.89 & 77.91 & 95.26 &  \cellcolor[rgb]{ .867,  .922,  .969} 81.45  & 90.87 \\
    UI-TARS-1.5-7B & 63.38 & 74.86 & 85.85 & 81.48 & \textbf{87.21} & 89.09 &  \cellcolor[rgb]{ .867,  .922,  .969} 80.31  & 90.38 \\
    \textbf{Phi-Ground-4B-7C} &   70.42 & 82.12 & 73.58 & 81.48 & 73.26 & 99.13   &  \cellcolor[rgb]{ .867,  .922,  .969} 80.00  &  92.79   \\
    \textbf{Phi-Ground-4B-16C} &   75.59 & 88.27 & 83.96 & 87.65 & 82.56 & 98.97    &  \cellcolor[rgb]{ .867,  .922,  .969}  86.17 &  \textbf{95.19}   \\
    \textbf{Phi-Ground-4B-16C-DPO} &    \textit{\underline{79.81}} & 88.83 & \textbf{88.68} & 88.89 & \textit{\underline{84.88}} & \textit{\underline{99.21}}   &  \cellcolor[rgb]{ .867,  .922,  .969} \textbf{88.38} &   \textbf{95.19}  \\
    \textbf{Phi-Ground-4B-29C} &   73.71 & 87.71 & 83.96 & 90.12 & 82.56 & 98.97  &  \cellcolor[rgb]{ .867,  .922,  .969} 86.17 &  90.87   \\
    \textbf{Phi-Ground-7B-7C} &71.83 & 85.47 & 78.30 & 88.89 & 80.23 & \textbf{99.29}&  \cellcolor[rgb]{ .867,  .922,  .969}84.00  &  92.79   \\
    \textbf{Phi-Ground-7B-16C} &   76.53 & \textit{\underline{89.39}} & \textit{\underline{86.79}} & \textbf{91.36} & 79.07 & 98.81  &  \cellcolor[rgb]{ .867,  .922,  .969} 86.99 &  92.31   \\
    \textbf{Phi-Ground-7B-16C-DPO} &   \textbf{81.22} & \textbf{89.94} & 85.85 & \textbf{91.36} & 81.40 & \textit{\underline{99.21}}  &  \cellcolor[rgb]{ .867,  .922,  .969} \textit{\underline{88.16}} &  93.75   \\
    \textbf{Phi-Ground-7B-29C} &    76.53 & 86.03 & 84.91 & 90.12 & 80.23 & 98.10  &  \cellcolor[rgb]{ .867,  .922,  .969} 85.99  &  92.31   \\
    \bottomrule
    \end{tabular}%
    }
  \label{tab:gold-detail}%
\end{table}%

\newpage
\section{Coordinates Representations and Loss}
As discussed in the main text, we experimented with various coordinate representations and loss function designs. We found that these techniques can accelerate training when dealing with small datasets. However, when the training dataset exceeds 1 million samples, these methods do not exhibit significant improvements. Consequently, the content presented in this section highlights approaches that failed to scale. We disclose these findings to help future researchers avoid similar pitfalls.

Overall, the development of all the techniques discussed in this section stems from the following considerations: Unlike regression loss, modeling with natural language treats the difference between "19" and "20" as a gap of two tokens, which should theoretically be equivalent in distance to that between "18" and "19". Furthermore, differences in the units, tens, and hundreds places should have varying impacts—we might tolerate errors in the units place, but errors in the hundreds place are entirely unacceptable. These aspects highlight a disparity to regression loss and our expectations. However, experimental results indicate that using the most straightforward next token prediction and expressing coordinates in natural language is well enough, and there is no significant difference in outcomes among these techniques when the batch size is extremely large and the training volume is very high.

\subsection{Tokenized Coordinates}
\label{sec:special-token}
It has been observed that in most LLM, numbers are tokenized by digits, meaning the number 123 would be tokenized into three separate tokens: "1", "2", and "3". This form of representation offers limited interpretability in the context of images. In previous work, many researchers have modeled regions within an image using newly introduced special tokens. We also attempted this approach; however, when processing screenshots, the images often have extremely high resolutions, and buttons are relatively small. If we divide both the height and width into 1,000 discrete intervals and assign a new token to each square region, this would add 1 million tokens to the model, which is entirely impractical. Therefore, we opted to model the coordinate values from 1 to 1,000 as new tokens, for example $<p~123>$ for value $123$, using two tokens to represent a single position: $<p~123><p~456>$. In this way, we added only 1000 special tokens to the model. We discussed various strategies for initializing these 1000 tokens. To better illustrate our point, we first introduce the following definition: let the function $emb: V\rightarrow R^n$ represent the retrieval of the embedding for a specific token $v\in V$ from a pre-trained model. Then we define the following variables:
\[C_{rand}=random(mean=MEAN(\{emb(v), v \in V\}), std=STD(\{emb(v), v \in V\}))\]
\[C_{digit}=random(mean=MEAN(\{emb(v), v \in digit\}), std=STD(\{emb(v), v \in digit\}))\]
\[M_{digit}=\frac{1}{|digit|}\sum_{v\in dight} emb(v) \qquad\qquad E_{p} = emb(\text{"point"})\]

We then consider the following five initialization methods:
\begin{itemize}
    \item The implementation in the Hugging Face Transformers library (hf) calculates the mean and variance of all pre-trained embeddings to generate a random vector, which is then used to initialize all newly added tokens.
    \item We adapted the hf method by restricting the embeddings used for calculating the mean and variance to only digit tokens, naming this approach R-digit.
    \item The digit-mean method directly uses the mean of the embeddings of digit tokens from the pre-trained model to initialize larger number tokens.
    \item The term main-digit refers to using the embedding of the digit in the hundredths place as the initialization. For example, for the number 234, the initialization would use the embedding of the digit "2."
    \item Prefix-learning method first freeze all parameters except the embedding of newly added tokens, and then used the learned embedding to initialize in the later training.
\end{itemize}

Additionally, we handle <point>, </point> and digits 0 to 9 differently. The specific assignments can be found in Table \ref{tab:special-token}, where the training results are also presented. 

\begin{table}[htbp]
  \centering
  \scriptsize
  \caption{The details and result of different initialization methods for newly added special tokens.}
    \begin{tabular}{l|cccccccccc|c}
    \toprule
    \multirow{2}[2]{*}{method name} & \multicolumn{10}{c|}{Initial value for special tokens}                        &  \\
          & \multicolumn{1}{c}{<point>} & \multicolumn{1}{c}{</point>} & \multicolumn{1}{c}{<p 0>} & \multicolumn{1}{c}{<p 1>} & \multicolumn{1}{c}{…} & \multicolumn{1}{c}{<p 9>} & \multicolumn{1}{c}{<p 10>} & \multicolumn{1}{c}{<p 11>} & \multicolumn{1}{c}{…} & \multicolumn{1}{c|}{<p 999>} & \multicolumn{1}{c}{Gold-S} \\
    \midrule
    hf    &   $C_{rand}$    &   $C_{rand}$    &   $C_{rand}$    &    $C_{rand}$   &       &   $C_{rand}$    &   $C_{rand}$    &     $C_{rand}$  &     &   $C_{rand}$    &  81.7\\
    R-digit &  $E_p$    &    $E_p$    &   $C_{digit}$    &   $C_{digit}$    &       &    $C_{digit}$   &   $C_{digit}$    &    $C_{digit}$   &       &   $C_{digit}$    &  83.2 \\
    digit-mean &    $E_p$    &     $E_p$   &    $emb(\text{"0"})$   &   $emb(\text{"0"})$    &       &    $emb(\text{"9"})$   &    $M_{digit}$   &   $M_{digit}$    &       &   $M_{digit}$    &  74.6 \\
    main-digit &    $E_p$    &    $E_p$    &    $emb(\text{"0"})$   &   $emb(\text{"0"})$    &       &  $emb(\text{"0"})$     &    $emb(\text{"0"})$   &   $emb(\text{"0"})$    &       &  $emb(\text{"9"})$     & 12.6 \\
    prefix-learning &    -   &    -   &  -     &    -   &       &   -    &     -  &  -     &       &   -    &  82.3 \\
    \midrule
    Natural language &    -   &    -   &  -     &    -   &       &   -    &     -  &  -     &       &   -    &  88.9 \\
    \bottomrule
    \end{tabular}%
  \label{tab:special-token}%
\end{table}%

We set the training volume for these experiments to 5 million samples and detailed setting can be found in Table \ref{tab:exp-setting}. Regrettably, the results in Table \ref{tab:special-token} indicate that all initialization methods significantly underperform compared to directly using natural language to express coordinates. Additionally, during training, we observed slow convergence and significant fluctuations in the gradient norm. These phenomena suggest that the introduction of too many special tokens, which have not undergone large-scale pre-training, can interact adversely with the pre-trained parameters, leading to training collapse. However, when high resolution is required, 1000 special tokens become necessary. Thus, in the field of UI grounding, this technique appears to be less practical.

\subsection{Label Smoothing}
\label{sec:label-smoothing}
It is fairly intuitive to consider that applying label smoothing to digit tokens might produce an effect similar to that of regression loss. Specifically, for instance, when the ground truth token is the digit 5, the standard Cross-Entropy loss assigns a label of 1 to digit 5 and 0 to all other tokens. However, if we apply label smoothing and assign a certain degree of smoothing to digits 4 and 6, we effectively inform the model that digits 5 and 4, 6 are numerically close. This approach results in an outcome akin to regression loss. In fact, it is possible to derive how to set label smoothing parameters to achieve equivalence with using regression loss.

We will next derive the formula for our label smoothing technique. First, we provide the following notations. Let $\textbf{x} = (x_0, x_1, ... x_n)$ be a tokenized sentence and $x_n$ is a digit token whose digit value(integer number) is $T$, $t$ is the token id of $x_n$. We usually model the probability given by language model as $p(\textbf{x})=p(x_0) p(x_1 | x_0) \ldots p(x_n | x_0, x_1, \ldots, x_{n-1})$. Define $V$ as the vocabulary set, $V_d$ as the set of digit token and $V = V_d \cup V_t$. Now let's recall the formulation of classical language modeling loss:
\[  
L_{lm} = -(\log p(x_0) + \log p(x_1 | x_0) + \ldots + \log p(x_n | x_0, \ldots, x_{n-1})) 
\]  
The last term:
\[  
L_n = -\log p(x_n | x_0, \ldots, x_{n-1}) := -\log p_n = -\sum_{i=1}^{|V|} y^{(i)} \log p^{(i)},  
\] 
where $p^{(i)}$ is the softmax result of last hidden activations and $y_i$ is the label. In general cross-entropy loss, the label is just a one-hot encoding:
\[  
y^{(i)} =   
\begin{cases}   
1 & \text{if } i=t \\  
0 & \text{else}  
\end{cases}  
\]  
In the following, we will derive how to approximate the regression loss with an appropriate designed label $y^{(i)}$. For $k \in V_d$ let the digit number of digit token id $k$ be $K$. We design a regularized MSE loss as:
\[L_{MSE} = \mathbb{E}_{k \in V_d}[(K - T)^2] + \psi \mathbb{E}_{k \in V_t}[1],\]
where the second term is a punishment for other none-digit tokens and $\psi$ is the punishment factor. We can calculate that:

\begin{align*} 
L_{MSE} &= \mathbb{E}_{k \in V_d}[(K - T)^2] + \psi \mathbb{E}_{k \in V_t}[1]  \\
&= \sum_{k \in V_d} (K - T)^2 p^{(k)} + \psi \sum_{k \in V_t} p^{(k)}  \\
 &= \sum_{k \in V_d/\{t\}} (K - T)^2 p^{(k)} - \psi\sum_{k\in V_d/\{t\}}p^{(k)} + \psi\sum_{k\in V_d/\{t\}}p^{(k)} + \psi \sum_{k \in V_t} p^{(k)}  \\ 
&= \sum_{k \in V_d/\{t\}} [(K - T)^2 - \psi] p^{(k)} + \psi (1 - p^{(t)})  \\
&= -\psi \left\{ p^{(t)} + \sum_{k \in V_d/\{t\}} [1-\frac{(K - T)^2}{\psi}] p^{(k)} - 1 \right\}
\end{align*} 

This will be equivalent to optimize loss: $\tilde{L}_{MSE}=- \sum_{i=1}^{V} \tilde{y}^{(i)} p^{(i)}$, where:
\begin{equation}
    \tilde{y}^{(i)} =   
\begin{cases}   
1 & \text{if } i=t \\  
1- \frac{d(K, T)}{\psi} & \text{if } i \in V_d \setminus \{t\} \\  
0 & i \in V_t  
\end{cases}  
\label{eq:label-smoothing}
\end{equation}
$d(K,T)$ is the distance function, for MSE, it should be $d(K,T)=(K-T)^2$. We directly use Equation \ref{eq:label-smoothing} for label smoothing, denoted as $\hat{L}_{MSE}=- \sum_{i=1}^{V} \tilde{y}^{(i)}\log p^{(i)}$. It is important to note that this differs from $\tilde{L}_{MSE}$. However, by leveraging the convexity of the logarithmic function, we can easily demonstrate that $\hat{L}_{MSE}$ serves as an upper confidence bound for $\tilde{L}_{MSE}$. From a practical standpoint, the information embedded in Equation \ref{eq:label-smoothing} suggests that the label values are larger for positions closer to the target, which is highly intuitive.

\begin{table}[htbp]
  \centering
  \caption{The scaling effect of label smoothing technique is very limited.}
    \begin{tabular}{lc|cc|cc}
    \toprule
    \multirow{2}[2]{*}{$d(K,T)$} & \multirow{2}[2]{*}{$\psi$} & \multicolumn{2}{c|}{$N=50K,\quad BS=64$} & \multicolumn{2}{c}{$N=1M \quad BS=2048$} \\
          &       & ScreenSpot-V2  & Gold-S & ScreenSpot-V2  & Gold-S \\
    \midrule
    $d(K,T)=(K-T)^2$     & 10    &   52.5    &  66.9     &   81.7    &   85.9\\
    $d(K,T)=(K-T)^2$     & 30    &   55.8    &  66.2     &   83.3    &  88.1 \\
    $d(K,T)=|K-T|$     & 10    &   54.6    &  67.6     &      \textbf{84.3} &  87.8 \\
    $d(K,T)=|K-T|$     & 30    &    \textbf{60.1}   &  \textbf{69.6}    &  84.1     &   87.6 \\
    \midrule
    no label smoothing    &      &   52.3    &   63.2    &   84.2    &  \textbf{88.8} \\
    \bottomrule
    \end{tabular}%
  \label{tab:label-smoothing}%
\end{table}%

The results, as shown in Table \ref{tab:label-smoothing}, indicate that this technique demonstrates a clear advantage when dealing with smaller datasets and smaller batch sizes. However, when we increased the training data size to 1 million and the batch size to 2048, we observed little to no improvement, with results falling within the margin of fluctuation. This suggests that while the approach may offer some acceleration benefits in resource-constrained scenarios, it holds limited significance for large-scale training. In such cases, with larger batch sizes, the model's optimization direction aligns with the regression loss, reducing the scaling advantage of this technique.

\subsection{Loss Re-weighting}
\label{sec:loss-reweright}
We attempt to assign different weights to the loss of different tokens. For instance, we assign higher weights to tokens in the tens place compared to those in the units place. This approach ensures that the model prioritizes the correctness of the tens place over the units place. If expressed formally using equations, it can be represented as follows:
\[  
L_{re} = -\sum_{t=0}^{n}w_t\log p(x_t|x_{t-1}...,x_{0}).
\] 
Initially, we set $w_t = 1.0$ when $x_t$ is not a digit token. Then we consider different settings of weights for digit tokens, as shown in Table \ref{tab:reweight}.

\begin{table}[htbp]
  \centering
  \caption{Experiment on selecting parameters for loss reweighting.}
    \begin{tabular}{ccc|cc|cc}
    \toprule
    \multicolumn{3}{c|}{weights for digit position} & \multicolumn{2}{c|}{$N=50K,\quad BS=64$} & \multicolumn{2}{c}{$N=1M \quad BS=2048$} \\
    hundrads & tens  & units & ScreenSpot-V2  & Gold-S & ScreenSpot-V2  & Gold-S \\
    \midrule
    1.0     & 1.0     & 1.0     &  52.3    &   63.2    &   84.2    &  \textbf{88.8} \\
    \midrule
    2.0     & 1.5   & 1.0     &    6.3   &   16.2    &    15.5   &  22.7 \\
    4.0     & 2.0     & 1.0     &    0.0   &   0.0    &   0.0    &  0.0 \\
    \midrule
    1.0     & $1/10$    & $1/100$    &    55.4   &  \textbf{66.8}     &   \textbf{84.5}    &  85.8 \\
    1.0     & $1/\sqrt{10}$    & $1/10$    &   54.7    &   63.6    &    83.3   &  86.7 \\
    1.0     & $1/\ln 10$    & $1/(\ln 10)^2$    &  \textbf{57.6}     &   64.8    &   84.1    &  86.7 \\
    \bottomrule
    \end{tabular}%
  \label{tab:reweight}%
\end{table}%

The first block in the table represents the control group without using reweighting techniques. We first confirm that the weight of digit tokens cannot exceed 1.0, even though these format-related tokens appear in almost all data. We found that if the weight of digit tokens is even slightly greater than 1.0, it causes the model to output in an unexpected format during testing. This results in parsing errors, causing the test results to be nearly zero. In contrast, in models trained under normal conditions, the proportion of parsing errors is nearly zero.

When the weight of digit tokens is proportionally adjusted to smaller than 1.0 and the model is trained accordingly, we obtain results similar to those described in Section \ref{sec:label-smoothing}. Specifically, when training resources are extremely limited, we observe that the reweighting technique accelerates model convergence and achieves consistently better results in low-sample scenarios. However, when the data size and batch size increase significantly, the benefits of this technique are minimal or unstable. Such results are insufficient to support us to widespread this technique. The reason for this phenomenon may be that when the training volume and batch size increase, learning based on higher numeric values exhibits greater certainty and stability (or lower perplexity), which facilitates more effective learning. In contrast, learning based on lower numeric values might fluctuate due to errors present in the dataset, leading to slower learning. This process is similar to our reweighting technique, and therefore, as the training volume and batch size increase, this technique is effectively replaced.

\section{Data Pre-processing Details}

\subsection{CommonCrawl Data Pre-processing}

\subsubsection{Rendering Resolutions and Filtering}
\label{sec:CC-render-rule}

\paragraph{Rendering resolutions.} When rendering each web page, we initially select a screen size with equal probability (1/3 chance) from the options of 1080p, 2.5k, and 4k. The corresponding screen area ($Space$) are $1920\times 1080$, $2560\times 1440$, and $3840\times 2160$, respectively. For a given screen area, we then randomly choose an aspect ratio $(R_w, R_h)$ from the following set of aspect ratios:
\[
(R_w, R_h) \in \left\{ (1+\frac{i}{N}, 2-\frac{i}{N}) | i = 0, 1, ..., N \right\}.
\]
The final screen size $(W, H)$ used for rendering can be calculated using the following formula:
\[
(W, H) = (R_w\times S, R_w\times S), \qquad S=\left\lceil \sqrt{\frac{Space}{R_wR_h}} \right\rceil
\]

\paragraph{Filtering.} During webpage rendering, we implemented certain filtering processes. Unlike the filtering described in Section \ref{sec:CC-filter-rule}, this stage of filtering requires the relevant code (such as JavaScript) used in browser rendering. In contrast, the subsequent filtering refers to offline filtering conducted after the necessary information has been stored. Therefore, even though both processes involve rule-based filtering, they are described in two separate sections. During webpage rendering, we can use JavaScript to obtain interactive information, and based on this information, we have established the following filtering rules for all HTML elements.

We first retain only those elements that meet any of the following conditions:
\begin{itemize}
    \item Interactive Tags: The HTML tag name of the element is one of \textit{'button'}, \textit{'input'}, \textit{'textarea'}, \textit{'select'}, \textit{'a'}, \textit{'form'} 
    \item Event Attribute: The elements have specific JavaScript methods (functions) attached, such as \textit{'onclick'}, \textit{'onmousedown'}, \textit{'onmouseup'}, \textit{'onmouseover'}, \textit{'onmouseout'}, \textit{'onkeydown'}
    \item Role attribute: Elements with the following role attributes are generally interactive: \textit{'button'}, \textit{'link'}, \textit{'textbox'}, \textit{'menuitem'}, \textit{'option'}, \textit{'checkbox'}, \textit{'radio'}, \textit{'tab'}, \textit{'switch'}.
    \item Interactive class: The class name of the element is a string type and the class name is one of \textit{'btn'}, \textit{'button'}, \textit{'input'}, \textit{'link'}, \textit{'nav'}, \textit{'menu'}, \textit{'item'}.
    \item Is icon: Tag name is one of \textit{'i'}, \textit{'span'}, \textit{'svg'} and the class name is one of \textit{'fa'}, \textit{'fas'}, \textit{'far'}, \textit{'fal'}, \textit{'fab'}, \textit{'material-icons'}
    \item Is image: The tag name is \textit{'img'}
\end{itemize}

Subsequently, we remove all elements that are not visible on the screen, such as those that require scrolling to be viewed. We then store all relevant information of the elements that meet the criteria, in JSON format, along with the rendered images. This includes details such as bounding box coordinates, HTML code, and various attributes. We also store a layout diagram, which matches the size of the rendered screenshot. However, different types of elements are represented by different colors occupying their respective areas; for example, interactive text is marked in red, and images in cyan. This type of layout diagram allows for the expression of button positions without focusing on the content itself and is used for data deduplication at the webpage level.

\subsubsection{Offline Rule-based Filtering}
\label{sec:CC-filter-rule}
Once the data has been stored, we consider the following filtering rules:

\paragraph{Boxes deduplication.} If a box completely encompasses multiple other boxes, we first remove the outer box. Such boxes are typically div containers of a module area, containing multiple related buttons. When one box contains another (determined by an IoU greater than a certain threshold), we remove the larger box. This situation is common in web design with nested containers and errors generated by OmniParser when using it to create boxes.

\paragraph{Remove empty boxes.} In both webpage rendering and OmniParser, there are instances where certain boxes appear in a blank, solid-colored area devoid of any content. For each candidate box, we crop the corresponding region from the screenshot and use the pixel standard deviation to directly determine if the area is solid-colored. If it is deemed to be solid-colored, we delete the box.

\paragraph{Text content recognition.} Sometimes, the text content within a screenshot, such as a long sentence, is recognized as an element. We wish to retain buttons with text but not these non-interactive content texts. To achieve this, we use the aspect ratio of the box as a filtering criterion. If the aspect ratio of a box exceeds a certain threshold, we discard that box.

\subsection{Re-sampling algorithm}
\label{sec:re-sample-alg}
\begin{algorithm}[H]
    \centering
  \caption{Re-sampling algorithm}
  \begin{algorithmic}[1]
    \Require The center point set of training dataset $C=\{(x_i, y_i)\}$ using relative coordinates. Segmentation granularity $N,M$, Sampling factor $\psi$.
    \Ensure Sampled center point set $\hat C$
    \State $\text{all\_box} \gets \{(i, j): \text{list()} \mid i \in \{0,1,...N-1\}, j \in \{0,1,...M-1\}\}$  

    \For{$(x,y)$ \textbf{in} $C$}
    \State $\text{locate} \gets (\text{int}(x // (1.0/N)), \text{int}(y // (1.0/M)))$  
    \State $\text{all\_box}[\text{locate}].\text{append}((x,y))$ 
    \EndFor
\State $\text{dist} \gets [\text{len}(v) \mid v \in \text{all\_box.values()}]$  
\State $\text{dist.sort()}$  
\State $\text{keep\_number} \gets \text{dist}[\text{int}(N\times M \times \psi)]$  
\State $\hat C \gets \text{list()}$

\For{v \textbf{in} \text{all\_box}}
    \State $\text{N\_Sample} \gets \textbf{min}(\text{len}(v), \text{keep\_number})$
    \State $\hat C.\text{extend}(\textbf{random\_sample}(v, \text{N\_Sample}))$
\EndFor

  \end{algorithmic}
\label{alg:data-sampling}

\end{algorithm}

\newpage

\section{Prompts}

\subsection{Reference expression generation prompt.}
\label{sec:prompt-long-gold}

\begin{tcolorbox}[title={Prompt for generating Long-Gold RE / Training data generation}]
System prompt:

------------------------------------------------------------------------------------------------------------------

User will provide you with a screenshot, in which a specific area will be 
highlighted with a red rectangular box.
We will also provide a cropped image of the corresponding area, and 
(optionally) additional information related to the area to help you 
understand it. 
Your task is to generate several references regarding the target area 
on the original image.

Specific task requirements are as follows: You will analyze and output a dictionary in JSON file format. The key-value pairs included are as follows:

\begin{itemize}
    \item area\_type: Choose one from 'icon', 'text'. These represent whether the target area is an indicative icon, text.
    \item interactive: bool, indicating whether this element in the screenshot scenario is an interactive element (e.g., clickable, inputable, etc.). If it is static text or an image, then it is not interactive.
    \item context: Generate a background context describing why the current screen and area would be used. For example, if the area is the close button of a image file, the context could be that the user is editing a file and has completed their task at this moment.
    \item functional\_reference: A reference about the target area, involving the function of the target area.
    \item positional\_reference: A reference about the target area by describing the position of the target area, such as layout and nearby elements. 
    \item appearance\_reference: A reference about the target area by describing the appearance of the target area.
\end{itemize}

\textbf{Ensure that anyone can uniquely identify this area in the screenshot through any one of the references.}

\textbf{Don't mention red rectangular box.}

\textbf{Your output references should only include the element description itself and follow the requirements. Do not start with "the target element" or "the element"}

Your output should follow this format strictly:
\begin{lstlisting}
# Analyze
A free form analyze of the screen shot and task.
# Output
```json
{
    "area_type": ...,
    "interactive": ...,
    "context": ...,
    "functional_reference": ..., 
    ..., 
    "appearance_reference": ...
}
```
\end{lstlisting}

\end{tcolorbox}

\newpage

\begin{tcolorbox}

Model input:

------------------------------------------------------------------------------------------------------------------

\# Screenshot with highlight

\includegraphics[width=0.6\columnwidth]{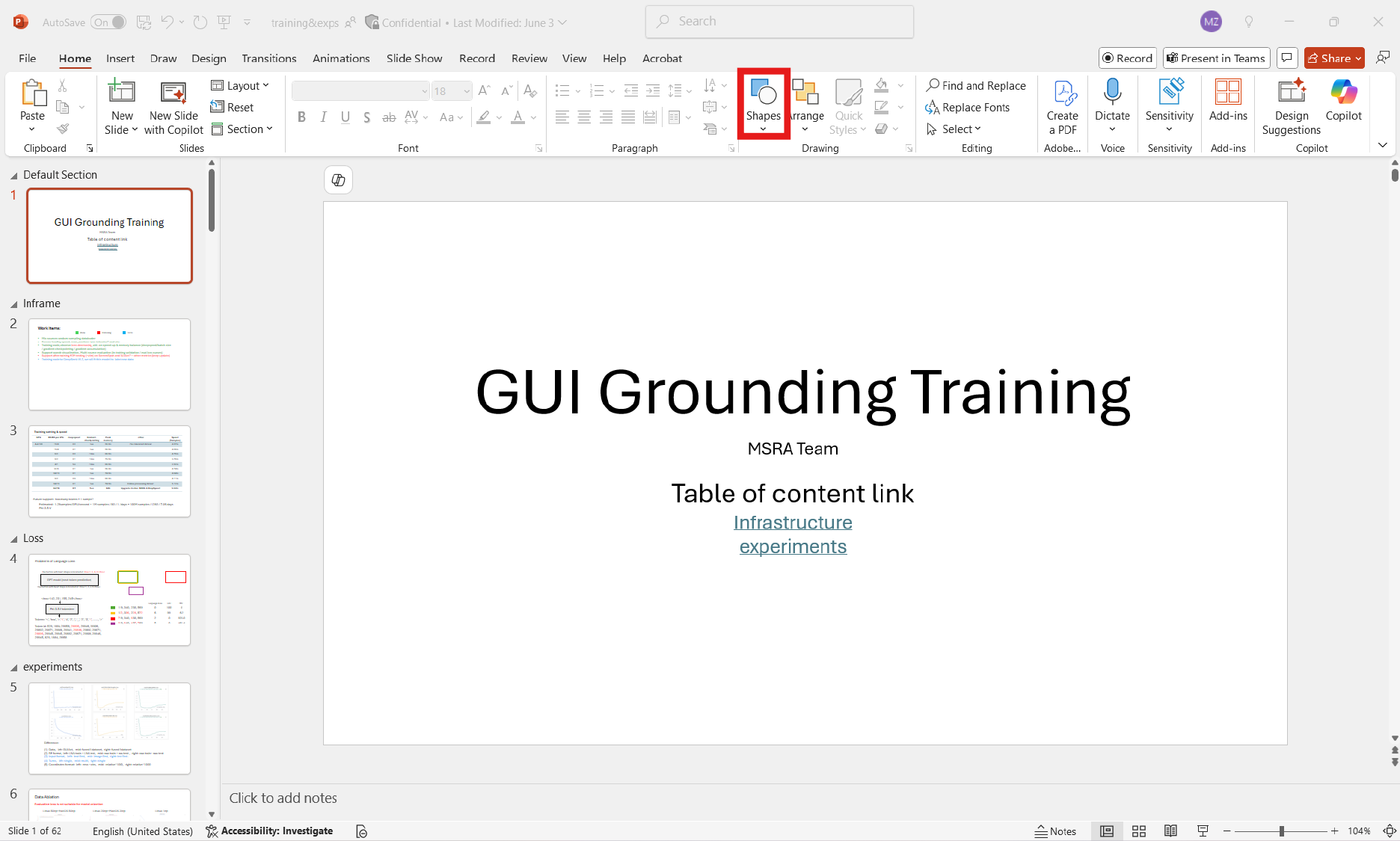}

\# Cropped target image

\includegraphics[width=0.2\columnwidth]{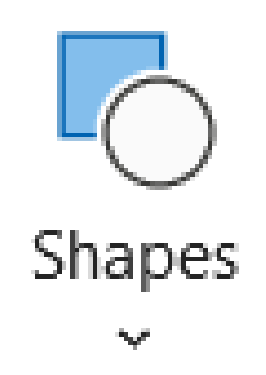}

Model output:

------------------------------------------------------------------------------------------------------------------

\# Analyze

The screenshot shows a PowerPoint presentation interface. The highlighted area is located on the toolbar of the PowerPoint application and is labeled as "Shapes." This area is used to insert various shapes into the presentation slides, an essential feature for users looking to enhance their slides with graphical elements. The toolbar section where the "Shapes" button is located is a common area for formatting and editing tools, making it easily accessible for users during the slide creation process.

\# Output
\begin{lstlisting}
{  
    "area_type": "icon",
    "interactive": true,
    "context": "While working on a PowerPoint presentation, the user is 
    likely looking to add graphical elements or illustrations to their 
    slides to enhance the visual appeal or to convey information more 
    effectively.",  
    "functional_reference": "Shapes button is used to add various 
    graphical shapes to a slide in a PowerPoint presentation.",  
    "positional_reference": "Located on the toolbar under the 'Home' tab, 
    to the right of the 'Layout' button and to the left of the 'Arrange'
    button.",  
    "appearance_reference": "Contains a white circle and a blue square, 
    with the label 'Shapes' underneath."  
}  
\end{lstlisting}
\end{tcolorbox}

\subsection{Planner's generation prompt for agent setting.}
\label{sec:prompt-long}

\begin{tcolorbox}[title={Prompt for generating Long RE / Agent settings for evaluation}]

System prompt:

------------------------------------------------------------------------------------------------------------------

User will provide you with a screenshot and a short instruction related to the area to help you understand it.Your task is to classify the target area and generate several references regarding the target area on the original image.

Specific task requirements are as follows: You will analyze and output a dictionary in JSON file format. The key-value pairs included are as follows:

\begin{itemize}
    \item context: Generate a background context describing why the current screen and area would be used. For example, if the area is the close button of a image file, the context could be that the user is editing a file and has completed their task at this moment.
    \item functional\_reference: A reference about the target area, involving the function of the target area.
    \item positional\_reference: A reference about the target area by describing the position of the target area, such as layout and nearby elements. 
    \item appearance\_reference: A reference about the target area by describing the appearance of the target area.
\end{itemize}

\textbf{Ensure that anyone can uniquely identify this area in the screenshot through anyone of the references.}

\textbf{Ensure that the reference is complete and independent.}

Your output should follow this format strictly:

\begin{lstlisting}
# Analyze
A free form analyze of the screen shot and task.
# Output
```json
{
    "context": ...,
    "functional_reference": ..., 
    ..., 
    "appearance_reference": ...
}
```
\end{lstlisting}

Model input:

------------------------------------------------------------------------------------------------------------------

\# Screenshot

\includegraphics[width=0.6\columnwidth]{images/appendix/prompt/long-ss.pdf}

\# Instruction

adjust the options of "show previous"

\end{tcolorbox}

\begin{tcolorbox}
Model output:

------------------------------------------------------------------------------------------------------------------

\# Analyze

The screenshot displays a macOS System Settings window focused on the "Notifications" section. The user intends to adjust the "Show previews" option, which is part of the Notification Center settings. This area allows users to control how notifications are displayed on their device, particularly when the screen is unlocked. The screen is divided into two main sections: the left pane with different settings categories and the right pane showing detailed options for the selected category, which is "Notifications" in this case.

\# Output

\begin{lstlisting}
{  
    "context": "The user is configuring notification settings to determine 
    how previews of notifications are displayed on their device when the 
    screen is unlocked.",  
    "functional_reference": "The 'Show previews' option allows the user to 
    control whether notification previews are shown on the screen when it 
    is unlocked.",  
    "positional_reference": "The 'Show previews' option is located in the 
    right pane of the System Settings window under 'Notification Center' 
    settings, just below the descriptive text about notifications.",  
    "appearance_reference": "The 'Show previews' option is a dropdown menu 
    located next to the text 'Show previews' with the current setting 
    'When Unlocked'."  
}  
\end{lstlisting}

\end{tcolorbox}

\section{More Cases Study}

\label{sec:more-case-study}

\begin{tcolorbox}[title={Error grounding case study with human level reference. Case-1: Similar Icons}]

Reference:

------------------------------------------------------------------------------------------------------------------

Toggles visibility of all annotations (dimensions, notes, symbols) in the graphics area.
Located in the heads-up view toolbar at the top center of the viewport, immediately to the right of the Temporary Axes button and left of the Measure button.
Blue uppercase letter A on a light gray square background.

------------------------------------------------------------------------------------------------------------------

Output: 

------------------------------------------------------------------------------------------------------------------

\includegraphics[width=0.8\columnwidth]{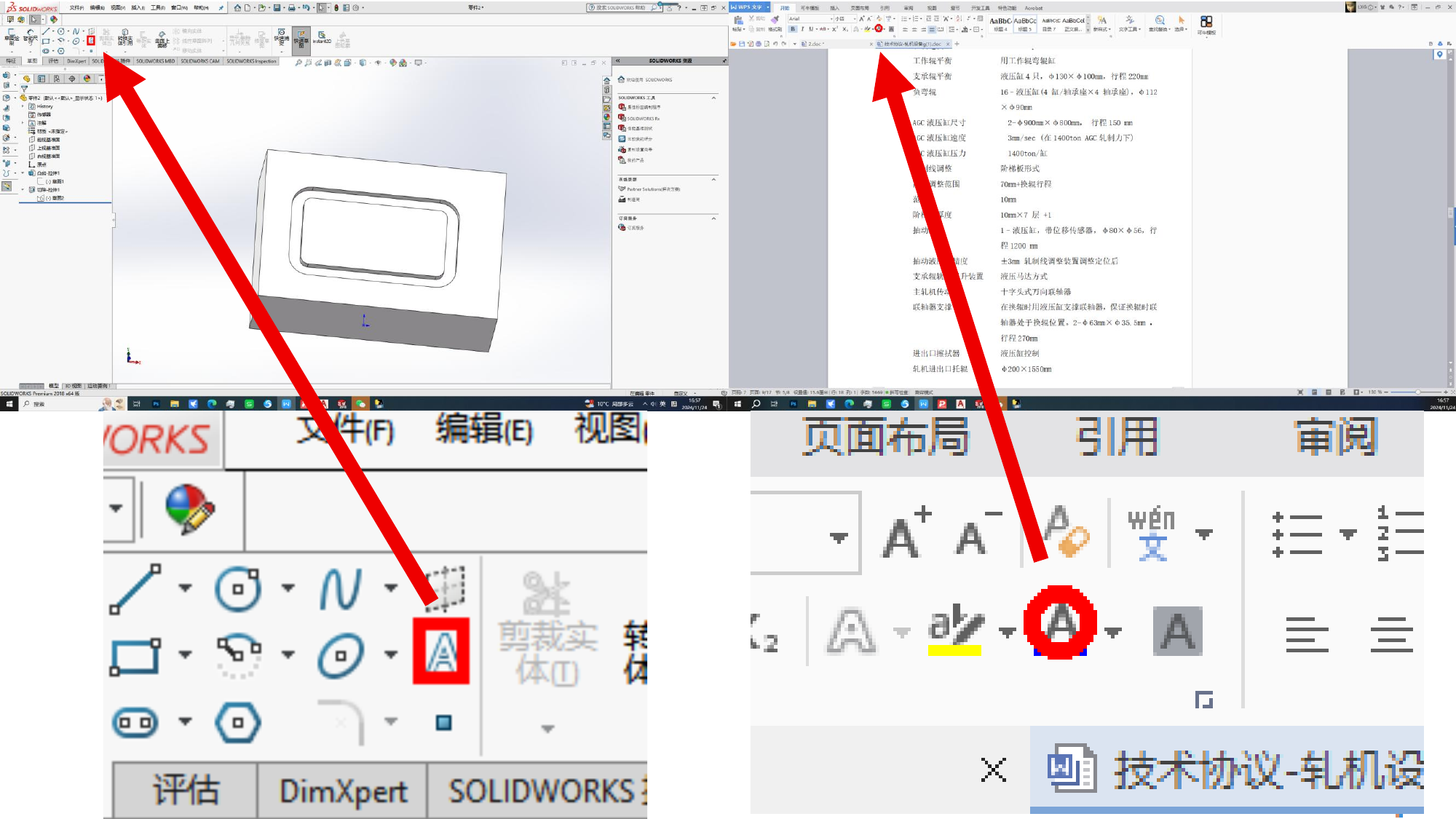}

\end{tcolorbox}

\begin{tcolorbox}[title={Error grounding case study with human level reference. Case-2: Precision Error}]

Reference:

------------------------------------------------------------------------------------------------------------------

Convert Entities command projects selected edges or curves from the model into the active sketch as sketch entities.Located on the Quick Access Toolbar at the very top of the SOLIDWORKS window, immediately to the right of the Redo icon and before the application menu bar.Grey square button showing a diagonal blue dashed line connecting two white square endpoints.

------------------------------------------------------------------------------------------------------------------

Output: 

------------------------------------------------------------------------------------------------------------------

\includegraphics[width=0.8\columnwidth]{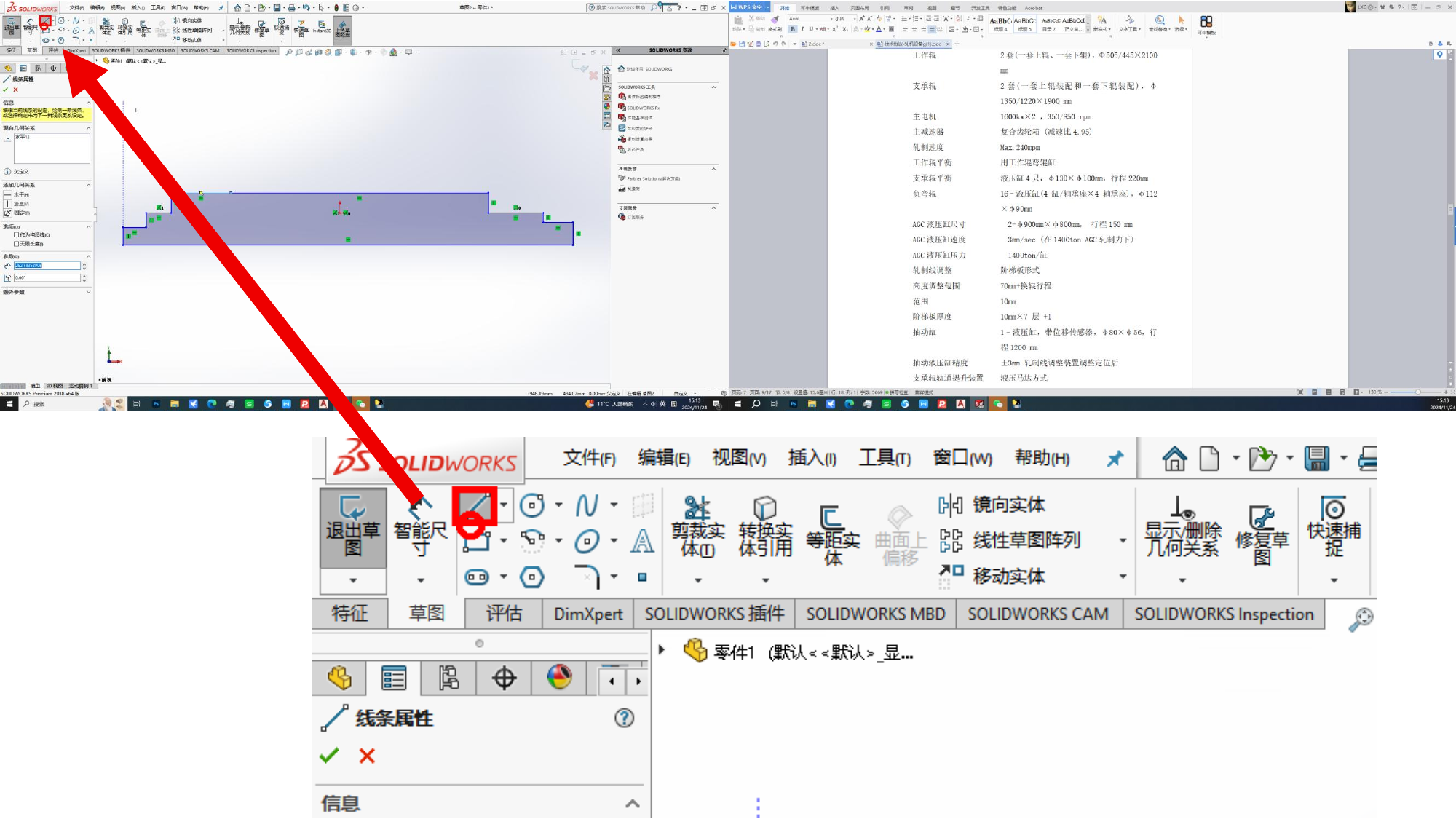}

\end{tcolorbox}

\begin{tcolorbox}[title={Error grounding case study with human level reference. Case-3: Lack spatial reasoning}]

Reference:

------------------------------------------------------------------------------------------------------------------

Grid cell for selecting a 9x7 table dimension when inserting a new table. Cell in the ninth column of the seventh row within the table size preview grid under the Table button in the Insert tab. Light-blue interior square outlined by a white inner border and a thicker blue outer border.

------------------------------------------------------------------------------------------------------------------

Output: 

------------------------------------------------------------------------------------------------------------------

\includegraphics[width=\columnwidth]{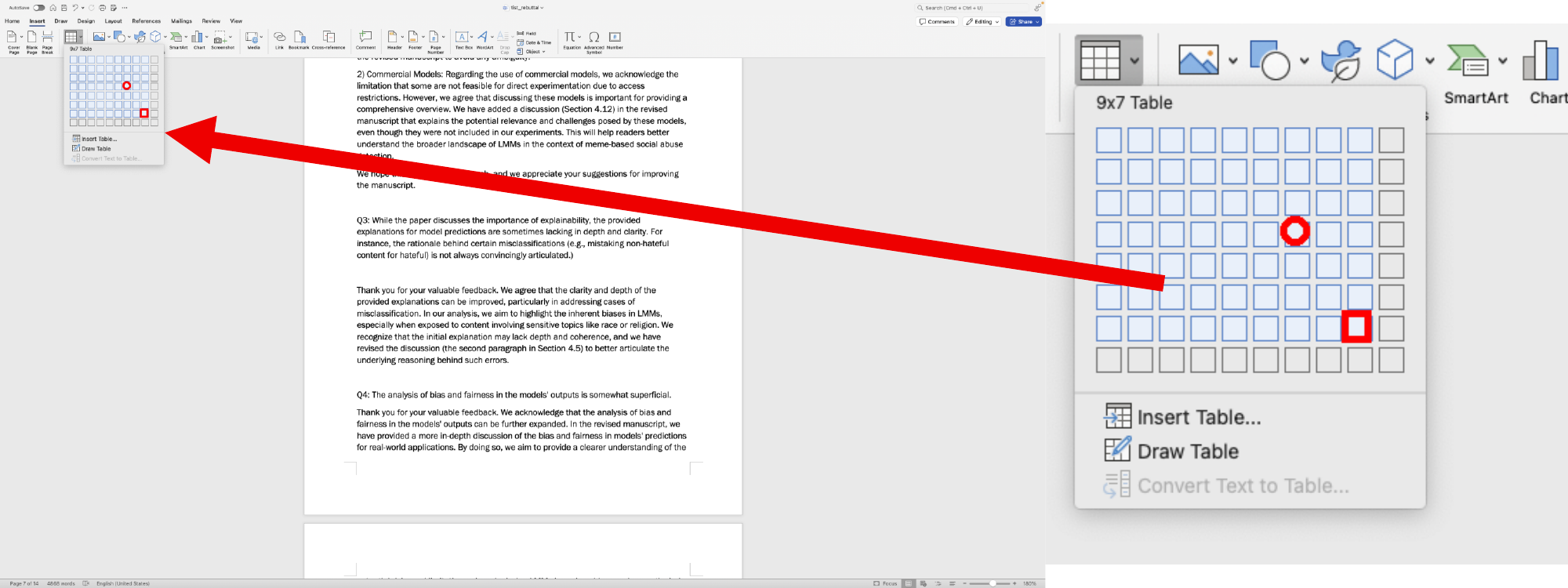}

\end{tcolorbox}

\begin{tcolorbox}[title={Error grounding case study with human level reference. Case-4: Two same area}]

Reference:

------------------------------------------------------------------------------------------------------------------

scatter plot template used to generate an X-Y scatter diagram from worksheet data. icon located in the top section of the 2D plot palette, in the row of scatter type graphs beneath the line and bar icons; it is the second icon from the left in that row. white rectangular button with several solid black dots arranged randomly inside and the Chinese label displayed underneath

------------------------------------------------------------------------------------------------------------------

Output: 

------------------------------------------------------------------------------------------------------------------

\includegraphics[width=\columnwidth]{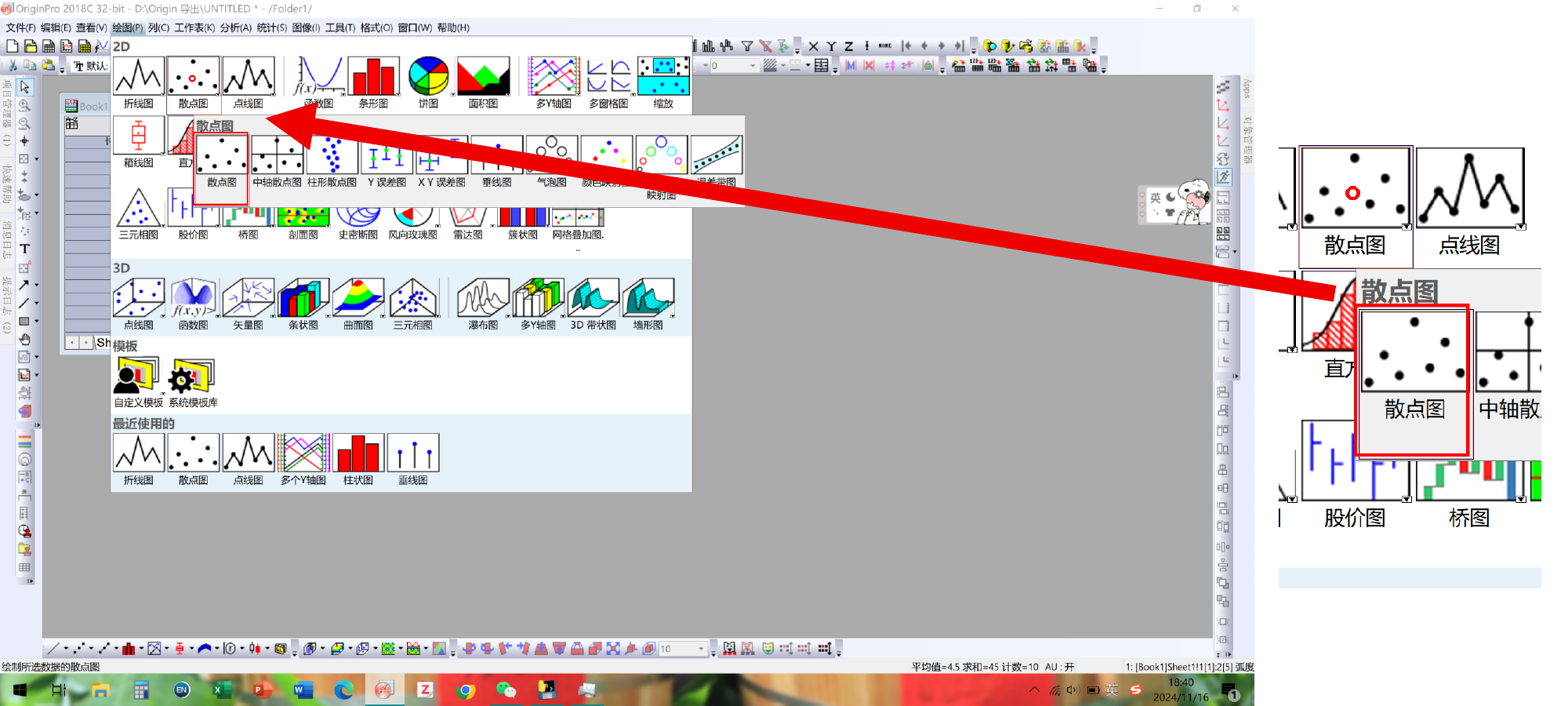}

\end{tcolorbox}

\begin{tcolorbox}[title={Error grounding case study with human level reference. Case-5: Interactive area}]

Reference:

------------------------------------------------------------------------------------------------------------------

Toggles the applied Digital Glitch effect on or off for live comparison in the Effect Controls panel. Found immediately to the left of the "Digital Glitch" effect name under the adjustment layer's effects list in the top-left panel. Small grey lowercase "fx" icon on a dark background matching the style of other effect-toggle buttons.

------------------------------------------------------------------------------------------------------------------

Output: 

------------------------------------------------------------------------------------------------------------------

\includegraphics[width=\columnwidth]{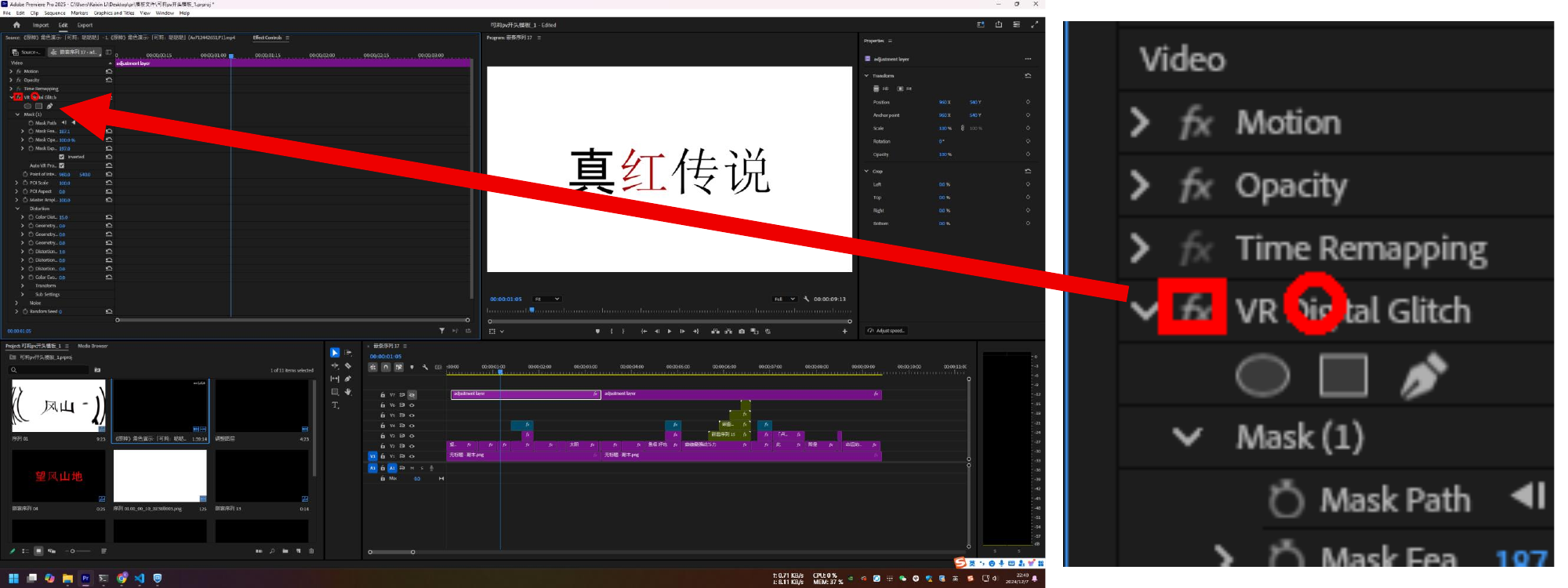}

\end{tcolorbox}

\end{document}